%% file: main.tex
\documentclass[10pt]{article} %
\usepackage[accepted]{tmlr}

\input{math_commands.tex}

\usepackage{hyperref}
\usepackage{url}

\input{config/packages}

\input{config/definitions}

\title{{ESLM: Risk-Averse Selective Language Modeling with \\ Hierarchical Batch Selection}}

\author{\name Melis Ilayda Bal \email mbal@tuebingen.mpg.de \\
      Max Planck Institute for Intelligent Systems, T\"{u}bingen, Germany \\
      LIONS - École Polytechnique Fédérale de Lausanne
      \\
      \\
      \name Volkan Cevher \email volkan.cevher@epfl.ch \\
      \addr LIONS - École Polytechnique Fédérale de Lausanne  
      \\
      \\
      \name  Michael Muehlebach \email michaelm@tuebingen.mpg.de\\
      \addr Max Planck Institute for Intelligent Systems, T\"{u}bingen, Germany
}

\def\month{08}  %
\def\year{2026} %
\def\openreview{\url{https://openreview.net/forum?id=vcynsg6rMN}} %

\begin{document}

\maketitle

\begin{abstract}
Large language model pretraining is compute-intensive, yet many tokens contribute marginally to learning, resulting in inefficiency.
We introduce Efficient Selective Language Modeling (\textsc{Eslm}), an online, risk-aware batch selection algorithm that improves training efficiency and distributional robustness.
\textsc{Eslm} operates in two phases: $(i)$ instance-level selection via a shallow early-exit model pass that computes proxy per-instance statistics (e.g., loss or entropy) and retains data points using value-at-risk thresholding; and $(ii)$ loss shaping with token-level selection via risk-aware thresholding on per-token scores.
This data-centric mechanism reshapes the training objective, prioritizing high-risk tokens and eliminating redundant gradient computation. 
Framed as a bilevel game, the model competes with a masking adversary selecting worst-case token subsets, recovering conditional value-at-risk minimization, and linking selective pretraining to distributionally robust optimization.
We extend our approach to \textsc{Ada-Eslm}, which adaptively tunes selection confidence during training. 
Experiments on GPT-2 pretraining show that \textsc{Eslm} consistently improves FLOPs-to-target efficiency, reducing effective training FLOPs while matching or improving validation perplexity and maintaining competitive downstream performance across multiple model scales, dataset mixtures, and knowledge distillation settings.
\end{abstract}

\section{Introduction}
\label{sec:introduction}
The growing scale of large language models (LLMs) has brought substantial improvements in downstream performance at the expense of significantly higher pretraining costs
\citep{brown2020language}.
Training LLMs is notoriously compute-intensive, requiring massive GPU resources and often processing billions of tokens uniformly.
Yet, many tokens, e.g., predictable or low-entropy ones, contribute little to model learning \citep{hullermeier2021aleatoric}.
Standard causal language modeling (CLM) treats all tokens equally in the loss, allocating compute uniformly to trivial tokens and more informative ones, leading to inefficient training and suboptimal use of resources \citep{lin2024rho}.
\looseness -1

Efforts to improve pretraining efficiency span architectural advances \citep{dao2023flashattention2}, token pruning \citep{hou-etal-2022-token}, and increasingly, data-centric strategies \citep{xia2024less, wang2024greats}. 
Among these, data-centric approaches show particular promise for sample efficiency through selective weighting or filtering of training examples \citep{katharopoulos2018not}. 
However, existing methods often rely on reference models or heuristics \citep{lin2024rho}, operate at the sequence level \citep{yu2024mates}, or require offline scoring \citep{xie2023dataselectionlanguagemodels, wettig2024qurating}.
These choices limit adaptability and scaling of the model to massive web-scale corpora, and the ability to exploit token-level heterogeneity, despite being crucial for optimizing training dynamics and resource usage in LLM pretraining.
\looseness -1

\begin{figure*}[t]
    \centering  
{\includegraphics[width=0.98\linewidth,trim=2 113 3 83,clip]{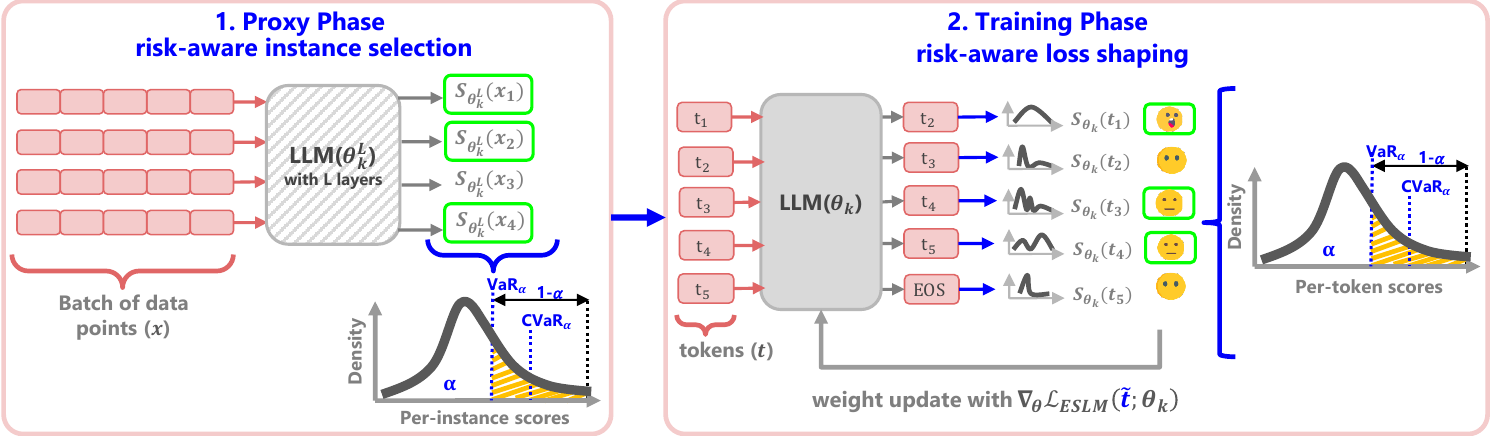}}
\setlength{\belowcaptionskip}{-1pt}
\caption{\textsc{Eslm} illustration:
\textsc{Eslm} uses VaR thresholding on proxy per-instance scores to select high-risk data points, then applies token-level VaR to perform loss shaping.
This reshapes the effective training distribution and loss by focusing computational resources on tokens with higher learning value.
\looseness -1
 }
\label{fig:eslm}
\vspace{-0.2cm}
\end{figure*}

We address this gap with \textsc{Eslm}: \textit{Efficient Selective Language Modeling}, a self-supervised data-centric framework that performs \textit{online token-level batch selection} for efficient and robust pretraining. 
\textsc{Eslm} proceeds in two phases (see~\cref{fig:eslm}): $(i)$ a \emph{proxy phase} that selects instances (sequences) via a shallow early-exit pass using only $L$ model layers based on proxy statistics (e.g., per-instance loss or predictive entropy \citep{shannon1948mathematical}), keeping instances by (conditional-)value-at-risk (CVaR/VaR) thresholding; $(ii)$ the training phase with risk-aware \emph{token-level loss shaping}, which retains only the highest-risk tokens in the training objective via VaR thresholding over per-token statistics.
This dynamic filtering shapes the training loss to emphasize uncertain or informative tokens, reducing redundant gradient updates and improving compute efficiency.
\looseness -1

Beyond its algorithmic simplicity, \textsc{Eslm} is grounded in a solid theoretical foundation.
Its risk-aware selection mechanism can be viewed as a bilevel game in which the model competes with a constrained adversary that restricts learning to the most challenging tokens, directly linked to distributionally robust optimization through targeted reshaping of the training distribution. 
Building on this, we extend our approach to \textsc{Ada-Eslm}, an adaptive variant that dynamically calibrates the confidence level in response to the training dynamics, enabling a principled control over the compute-efficiency and generalization trade-off.
Notably, \textsc{Eslm} is significantly \textit{lightweight}: it requires no auxiliary supervision, reference models, or computationally expensive offline scoring.
Our key contributions are as follows:
\looseness -1
\begin{itemize}[left=0.2cm]
    \item We propose \textsc{Eslm}, a risk-averse, two-phased 
    selective language modeling that prioritizes high-risk inputs (i.e., informative or uncertain) for efficient LLM pretraining.
    \textsc{Eslm} first performs instance-level filtering using a lightweight proxy, then applies token-level loss shaping on selected instances.
    Both phases use VaR thresholding over loss or entropy statistics.
    We provide its two variants: \myinlinecolorbox{electric-blue!15}{$\operatorname{VaR}$-entropy} and \myinlinecolorbox{salmon!30}{$\operatorname{CVaR}$-loss} based on the risk score choice.
    \looseness -1
    \item We frame \textsc{Eslm} as a bilevel adversarial game between the model and a masker that perturbs the effective training distribution by selecting worst-case token subsets under a constrained thresholding rule. We further show that for loss-based selection, \textsc{Eslm} admits a distributionally robust optimization interpretation by recovering the CVaR objective \citep{rockafellar2002conditional}, a well-established risk-sensitive formulation from robust statistics \citep{ben2009robust}.
    \item We propose \textsc{Ada-Eslm}, which adaptively adjusts the confidence level via a risk-aware CVaR-based controller to balance compute-efficiency and generalization. 
    \looseness -1
    \item We demonstrate \textsc{Eslm}'s utility for knowledge distillation by enabling a sparsified risk-aware teacher supervision provided for selected high-risk tokens. 
    \looseness -1
    \item Our experiments on GPT-2 (124M--1.5B) pretraining show that \textsc{Eslm} consistently improves FLOPs-to-target efficiency, reducing effective training FLOPs while matching or improving validation perplexity and maintaining competitive downstream performance across multiple model scales, dataset mixtures, and training settings.
    \looseness -1
\vspace{-0.2cm}
\end{itemize}
\vspace{-0.2cm}
\section{Background}
\label{sec:background}
This section introduces the key components underlying \textsc{Eslm}: CLM as the pretraining objective, token-level uncertainty estimation, and risk measures: VaR and CVaR.
\looseness -1 
\vspace{-0.2cm}
\paragraph{Causal Language Modeling (CLM).}
CLM trains a language model (LM) $\theta$ to predict each token in a sequence given the previous context. 
Given a corpus $\mathcal{C}$ of sequences $x = (x_1, \dots, x_M)$, each is of length $T$, drawn from a distribution $\mathcal{D}$ over a vocabulary $\mathcal{V}$,
the model factorizes the joint probability as:
$
P_\theta(x) = \prod_{j=1}^{T} P_\theta(x_j \mid x_{<j}).
\nonumber
$
CLM minimizes the average autoregressive loss:
\[
\vspace{-0.2cm}
\mathcal{L}_{\mathrm{CLM}}(x; \theta) = \mathbb{E}_{x \sim \mathcal{D}}[\ell_\theta(x)] = \frac{1}{T} \sum_{j=1}^{T} -\log P_\theta(x_j \mid x_{<j}),
\vspace{-0.2em}
\]
treating all tokens equally, despite many offering a limited learning signal \citep{lin2024rho}.
\vspace{-0.2cm}
\paragraph{Token-level risk.}
Let $S_\theta(x_j) \in \mathbb{R}$ denote the \textit{risk score} of token $x_j$, under LM $\theta$, computed using measures:
\begin{enumerate}[left=0.2cm,topsep=0pt,itemsep=0pt,label=(\roman*)]
    \item \label{entropy-defn}\textbf{Entropy} \citep{shannon1948mathematical}: $H_\theta(x_j) = -\sum_{v \in \mathcal{V}} P_\theta(v \mid x_{<j}) \log P_\theta(v \mid x_{<j})$.
    \item \label{loss-defn} \textbf{Loss}: $\ell_\theta(x_j) = -\log P_\theta(x_j \mid x_{<j})$.
\end{enumerate}
Both serve as proxies for token difficulty and informativeness—highlighting ambiguous, uncertain, or mispredicted tokens.
Entropy quantifies the model's predictive uncertainty based solely on its output distribution, prioritizing tokens for which the model remains uncertain. Loss instead measures the prediction error with respect to the observed target token, emphasizing examples that currently contribute the largest optimization error. Consequently, the entropy-based variant performs uncertainty-aware selective learning, whereas the loss-based variant focuses optimization on the high-loss tail of the empirical training distribution.
\textbf{Instance-level risk scores} are obtained by aggregating token-level risk scores, i.e., the mean score over non-padded tokens within each sequence (line 4, Algorithm~\ref{alg:eslm}).
\paragraph{Risk measures.} To prioritize high-impact tokens, we adopt risk-sensitive criteria from robust statistics \citep{gagne2021two}. Let $S_\theta(x_j)$ denote per-token risk score computed via LM $\theta$. The \textit{value-at-risk} (VaR) \citep{rockafellar2000optimization} at confidence level $\alpha \in (0,1)$ is the minimum threshold such that only the top $(1-\alpha)$ fraction of scores exceed the threshold; see~\cref{fig:eslm}: 
\begin{equation} 
\operatorname{VaR}_\alpha(S_\theta) := \inf \{ \tau \in \mathbb{R} \mid P(S_\theta \geq \tau) \leq 1 - \alpha \}. 
\label{eq:VaR-definition} 
\end{equation} 
The corresponding CVaR is a coherent risk measure \citep{artzner1999coherent} that computes the expected score among these highest-risk tokens \citep{rockafellar2002conditional}: 
\begin{equation} 
\vspace{-0.2em}
\operatorname{CVaR}_\alpha(S_\theta) := \min_\tau \mathbb{E}_{x \sim \mathcal{D}} [\tau + \frac{1}{1 - \alpha} \max(0, S_\theta(x) - \tau) ]. 
\label{eq:CVaR-definition} 
\vspace{-0.2em}
\end{equation} 
These tail-risk measures allow us to reshape the training objective to emphasize tokens that are difficult or uncertain, an idea we exploit in \textsc{Eslm} for efficient pretraining.
\section{\textsc{ESLM}: Risk-Averse Selective Language Modeling}
\label{sec:eslm}
\begin{algorithm}[t]
\caption{\textsc{Eslm}}
\label{alg:eslm}
\begin{algorithmic}[1]
\STATE \text{\bfseries Input:} LM $\theta$, dataset $\mathcal{D}$, learning rate $\eta$, confidence level $\alpha \in (0,1)$, batch size $M$, proxy $L$ layers.
\FOR{each training iteration $k=1,\dots,K$}
\STATE Sample a batch of instances $\mathcal{B} = \{x_1, \dots, x_M\} \sim \mathcal{D}$ of length $T$ $\{x_j^t\}_{t=1}^{T}, \forall j\in\{1,...,M\}$.
\STATE Compute per-instance scores $S_{\theta_k^L}(x)$ using early-exit $\theta_k^{L}$ proxy model. \hfill {\color{Periwinkle} /*Entropy or loss*/}
\STATE Compute threshold  $ \hspace*{0.2cm} S_{\theta_k^L, \alpha}^{\operatorname{VaR}} \gets \operatorname{VaR}_{\alpha} \left( \{ S_{\theta_k^L}(x_j) \}_{j=1}^{M} \right)$ using (\ref{eq:VaR-definition}).
\STATE $\Bar{\mathcal{B}} \gets \{ x_j \in \mathcal{B} \mid S_{\theta_k^L}(x_j) \geq S_{\theta_k^L, \alpha}^{\operatorname{VaR}} \}$.  \hfill {\color{Periwinkle} /* High-risk instance selection */}
    \STATE Compute per-token statistics via: \newline \hspace*{0.2cm}
    $ S_{\theta_k}(x_j^t) = 
    \begin{cases} 
      H_{\theta_k}(x_j^t) \text{ as in \ref{entropy-defn}}, & \text{(\myinlinecolorbox{electric-blue!15}{$\operatorname{VaR}$-entropy})} \\
      \ell_{\theta_k}(x_j^t) \text{ as in \ref{loss-defn}}, & \text{(\myinlinecolorbox{salmon!30}{$\operatorname{CVaR}$-loss})} 
    \end{cases}
    $
    \STATE Compute threshold \hspace*{0.2cm} $S_{\theta_k, \alpha}^{\operatorname{VaR}} \gets \operatorname{VaR}_{\alpha} \left( \{\{ S_{\theta_k}(x_j^t) \}_{j \in \Bar{\mathcal{B}}}\}_{t=1}^{T} \right)$ using (\ref{eq:VaR-definition}).
    \STATE $\tilde{\mathcal{B}} \gets \{ x_j^t \in \Bar{\mathcal{B}} \mid S_{\theta_k}(x_j^t) \geq S_{\theta_k, \alpha}^{\operatorname{VaR}} \}$.  \hfill {\color{Periwinkle} /* High-risk token selection */}
    \STATE Compute loss over selected tokens:\hfill {\color{Periwinkle} /*Shaped loss*/} \newline
    $ 
    \mathcal{L}_{\tilde{\mathcal{B}}}(x;\theta_k) =$ 
    $
    \begin{cases} 
      \mathbb{E}[\ell_{\theta_k}(x_j^t) \mid x_j^t \in \tilde{\mathcal{B}}], & \text{(\myinlinecolorbox{electric-blue!15}{$\operatorname{VaR}$-entropy}}) \\
      \operatorname{CVaR}_{\alpha}(\ell_{\theta_k}(x)) \text{ using } (\ref{eq:CVaR-definition}), & \text{(\myinlinecolorbox{salmon!30}{$\operatorname{CVaR}$-loss}}) \\
    \end{cases}
    $
    \STATE Update model parameters using optimizer $O$: \hspace*{0.2cm} $\theta_{k+1} \leftarrow O(\theta_{k}, \nabla_{\theta} \mathcal{L}_{\tilde{\mathcal{B}}}(x;\theta_k), \eta)$.
\ENDFOR
\RETURN $\theta_{K}$.
\end{algorithmic}
\end{algorithm}
We now introduce \textsc{Eslm}, a two-phase selective language modeling that improves pretraining efficiency by focusing optimization on high-risk inputs. 
We consider the causal language modeling setup presented in~\cref{sec:background},  where a language model with parameters $\theta$ is trained to minimize the expected token-level autoregressive loss.
While effective, the expectation-based CLM objective assumes uniform importance across all tokens, leading to two key inefficiencies: 
\begin{enumerate}[left=0.2cm]
\item It wastes computation on trivially predictable tokens that dominate the loss landscape but offer little signal.
\item It disregards token-level risk and overlooks rare, ambiguous, or out-of-distribution samples that are more informative for generalization and robustness.
\end{enumerate}
To address these inefficiencies, we adopt the Selective Language Modeling (SLM) paradigm \citep{lin2024rho}, which optimizes the model over a selected subset of tokens per training step. Formally, let $\pi_\phi(x)$ be a token selection policy that produces a binary mask $m = (m_1, \dots, m_T) \in \{0,1\}^T$ for an input sequence $x$, the SLM objective becomes:
\vspace{-0.1cm}
\begin{equation*}
\mathcal{L}_{\mathrm{SLM}}(\theta, \phi)= \mathbb{E}_{x \sim \mathcal{D},\, m \sim \pi_\phi(x)} \left[ \sum_{j=1}^T m_j \cdot \ell_\theta(x_j)\right],
\end{equation*}
where $\ell_\theta(x_j)$ is the per-token loss given in \ref{loss-defn}.
Existing approaches typically rely on learned or reference model ($\phi$)-based policies for $\pi_\phi$, that are expensive to train and may not generalize well \citep{lin2024rho}, or design offline selectors \citep{xie2023dataselectionlanguagemodels, wettig2024qurating}.
In contrast, \textsc{Eslm} is a self-supervised online SLM framework rooted in statistical risk that eliminates the need for an auxiliary external selector to improve computational efficiency.

\textsc{Eslm} reshapes the loss towards \textit{high-risk} inputs within each batch by dynamically filtering training signals both at the instance and token levels using risk-based thresholds derived from empirical batch distributions.  
Concretely, the risk is characterized by either $(i)$ high predictive uncertainty (\myinlinecolorbox{electric-blue!15}{$\operatorname{VaR}$-entropy} selection) or $(ii)$ high loss impact (\myinlinecolorbox{salmon!30}{$\operatorname{CVaR}$-loss} selection).
At each step, a candidate batch $\mathcal{B} = \{x_1, \dots, x_M\} \sim \mathcal{D}$ is sampled.  
A shallow early-exit proxy model $\theta^L$ (first $L$ layers of $\theta$) is used to compute per-token risk scores (entropy \ref{entropy-defn} or loss \ref{loss-defn}), which are aggregated into per-instance scores:
$
S_{\theta^L}(x_j) = \frac{1}{T} \sum_{t=1}^{T} S_{\theta^L}(x_j^t).
$
Given the empirical score distribution $\hat{\mathbb{P}}_{\mathcal{B}}$ over the batch, a VaR threshold at confidence level $\alpha$ is then applied to select high-risk instances:
\[
\vspace{-0.2em}
\begin{aligned}
\Bar{\mathcal{B}}
&= \{x_j \in \mathcal{B} \mid S_{\theta^L}(x_j) \geq S_{\theta^L,\alpha}^{\operatorname{VaR}} \}, \\
\text{where } 
S_{\theta, \alpha}^{\operatorname{VaR}}
&= \inf \left\{ \tau \in \mathbb{R} \;\middle|\;
\hat{\mathbb{P}}_{\mathcal{B}}\bigl(S_{\theta^L}(x_j) \geq \tau\bigr)
\leq 1 - \alpha
\right\}.
\end{aligned}
\vspace{-0.2em}
\]
On the reduced batch $\Bar{\mathcal{B}}$, the model $\theta$ computes per-token risk scores $S_\theta(x_j^t)$ using either entropy (\myinlinecolorbox{electric-blue!15}{$\operatorname{VaR}$-entropy}) or loss (\myinlinecolorbox{salmon!30}{$\operatorname{CVaR}$-loss}). Given the empirical score distribution $\hat{\mathbb{P}}_{\Bar{\mathcal{B}}}$ over the batch, a VaR threshold is applied at the token level, which defines a high-risk subset
$\tilde{\mathcal{B}} = \{x_j^t \in \Bar{\mathcal{B}} \mid S_\theta(x_j^t) \geq S_{\theta,\alpha}^{\operatorname{VaR}} \}$, and an associated normalized training distribution: $Q_\tau \in \mathcal{P}_\alpha(\Bar{\mathcal{B}}; \theta)$, $Q_\tau(x_j^t) \propto \mathbbm{1}[x_j^t \in \tilde{\mathcal{B}}]$, where $\tau$ corresponds to the minimizer defined in (\ref{eq:VaR-definition}).
The training proceeds by minimizing:
\[
\mathcal{L}_{\tilde{\mathcal{B}}}(\theta) = \mathbb{E}_{\Bar{\mathcal{B}} \sim \mathcal{D}} \left[
\mathbb{E}_{x_j^t \sim Q_\tau} \left[ \ell_\theta(x_j^t) \right] \right] = \mathbb{E}[\ell_\theta(x_j^t) \mid x_j^t \in \tilde{\mathcal{B}}],
\]
which corresponds to CVaR (in (\ref{eq:CVaR-definition})) when risk is based on token-level loss, and to an uncertainty-weighted loss when based on entropy.
Our approach is detailed in Algorithm~\ref{alg:eslm}. 
\paragraph{\textsc{ESLM} variations.}
While both \myinlinecolorbox{electric-blue!15}{$\operatorname{VaR}$-entropy} and 
\myinlinecolorbox{salmon!30}{$\operatorname{CVaR}$-loss} strategies select the upper tail of their respective score distributions; their inductive biases differ. The \myinlinecolorbox{salmon!30}{$\operatorname{CVaR}$-loss} selection emphasizes high-loss inputs, including both confidently incorrect and uncertain correct predictions. This helps the model correct overconfident mistakes and calibrate uncertainty. In contrast, the \myinlinecolorbox{electric-blue!15}{$\operatorname{VaR}$-entropy} selection focuses purely on predictive uncertainty, regardless of correctness, promoting learning in ambiguous or underexplored regions.
We illustrate these differences through qualitative examples in Appendix~\ref{app:token-selection-analysis}, showing that \textsc{Eslm} selects rare, or semantically rich tokens across domains.
\looseness -1

The following formulations apply to token-level selection; however, instance-level selection is similar.
\paragraph{Bilevel game formulation.} 
\textsc{Eslm} can be framed as a two-player adversarial game between the \textit{model} and a \textit{masker} (\textit{adversary}).
This provides a bilevel optimization perspective where the masker perturbs the effective training distribution by choosing a threshold $\tau$ that determines which tokens are selected for training, under the VaR constraint, and the model minimizes its loss over the induced sub-distribution.
Formally, the training process can be written as follows:
\begin{equation}
\vspace{-0.2em}
\label{eq:adversarial-game-formulation}
\begin{aligned}
& \min_\theta \;  
 \mathbb{E}_{\Bar{\mathcal{B}} \sim \mathcal{D}} \left[
\mathbb{E}_{x_j^t \sim Q_\tau} \left[ \ell_\theta(x_j^t) \right]
\right] \\
\text{s.t. } \;
\tau & \in \arg\min_{\tilde{\tau} \in \mathbb{R}} \left\{
\tilde{\tau} \;\middle|\;
\hat{\mathbb{P}}_{\Bar{\mathcal{B}}}
\bigl(S_\theta(x_j^t) \geq \tilde{\tau}\bigr)
\leq 1 - \alpha
\right\},
\end{aligned}
\vspace{-0.2em}
\end{equation}
where $\hat{\mathbb{P}}_{\Bar{\mathcal{B}}}$ is the empirical batch risk score distribution. 
This induces an adversarial dynamic in which the masker constrains optimization to the most challenging subset of tokens, pushing the model to improve performance on the tail distribution.
When the score function is $S_\theta(x_j^t) = \ell_\theta(x_j^t)$ (\myinlinecolorbox{salmon!30}{$\operatorname{CVaR}$-loss}), the selected tokens correspond to the upper tail of the empirical loss distribution, yielding a direct connection to Conditional Value-at-Risk (CVaR) optimization and risk-sensitive learning~\citep{curi2020adaptive, gagne2021two}. In contrast, when entropy is used as the scoring function (\myinlinecolorbox{electric-blue!15}{$\operatorname{VaR}$-entropy}), the framework instead prioritizes tokens according to predictive uncertainty. Thus, the CVaR interpretation applies directly to the loss-based variant, while the entropy-based variant should be viewed as a complementary uncertainty-aware selective learning strategy within the same risk-aware selection framework.

\paragraph{Distributionally Robust Optimization (DRO) interpretation.}
\textsc{Eslm} admits a distributionally robust optimization \citep{duchi-dro, kuhn2025distributionallyrobustoptimization} interpretation.
In classical DRO, the notion of a \emph{worst-case} distribution is defined with respect to the optimization objective: the adversarial distribution places greater probability mass on samples that incur the largest loss under the current model. Thus, optimization is driven by the high-loss regions of the data distribution rather than the average-case objective.
In ESLM, the VaR threshold restricts the training loss to a subset of tokens within the batch, those with scores in the top $(1 - \alpha)$ quantile.
This induces an adversarial sub-distribution $Q$ over the batch, supported only on the most challenging tokens.
Consequently, the retained tokens correspond to the highest-loss examples under the current model, making them the empirical analogue of the worst-case samples emphasized in DRO. Optimizing over this selected subset therefore resembles empirical CVaR optimization, where learning is driven by the tail of the loss distribution instead of its expectation. Formally, this can be written as:
\looseness -1
\begin{equation*}
\begin{aligned}
& \min_\theta \;
\sup_{Q \in \mathcal{P}_\alpha(\Bar{\mathcal{B}}; \theta)}
\mathbb{E}_{x_j^t \sim Q} \left[ \ell_\theta(x_j^t) \right]  \\ \text{where } \mathcal{P}_\alpha(\Bar{\mathcal{B}}; \theta) &:=
 \left\{
 Q \ll \hat{\mathbb{P}}_{\Bar{\mathcal{B}}} 
\mid
\operatorname{supp}(Q) \subseteq
\left\{
x_j^t \in \Bar{\mathcal{B}}
\mid S_\theta(x_j^t) \geq S_{\theta, \alpha}^{\operatorname{VaR}}
\right\}
\right\}.
\end{aligned}
\end{equation*}
This robust optimization perspective explains why \textsc{Eslm} improves generalization: by optimizing performance under adversarial distributions, 
the model develops robustness to distributional shifts. 
The empirical results in Section~\ref{sec:results} also confirm consistent generalization improvements by \textsc{Eslm}.
We emphasize that this DRO formulation is intended as an interpretive perspective rather than a formal robustness guarantee. Since the ambiguity set is constructed adaptively from the current empirical mini-batch using model-dependent risk scores, it differs from the uncertainty sets typically considered in classical DRO. Instead, this perspective provides intuition for understanding ESLM as allocating optimization effort toward high-risk regions of the empirical training distribution; establishing formal robustness or out-of-distribution guarantees remains an interesting direction.

\looseness -1

\paragraph{Risk-aware metric intuition.}
Rather than using fixed or heuristic thresholds, \textsc{Eslm} employs a quantile-based cutoff to batch score distributions, selecting instances and tokens from the high-risk tail. 
This links \textsc{Eslm} to distributionally robust optimization while offering clear advantages over ad hoc thresholds: $(i)$ the VaR/CVaR formulation provides a statistically principled mechanism to identify informative, difficult tokens with robustness to distribution shifts and coverage guarantees; $(ii)$ tail-risk formulation improves generalization (Section~\ref{sec:results}), beyond mere computational savings.
Further, adjusting the risk level $\alpha$ induces a token-level curriculum that balances efficiency and generalization, as demonstrated by \textsc{Eslm}'s adaptive extension (Section~\ref{sec:adaptive-eslm}).

\paragraph{Implementation and computational cost.} 
We implement \textsc{Eslm} at the mini-batch level, compatible with distributed training. Each step has two passes: we first run an early-exit proxy forward with the first $L$ transformer blocks of the same model in inference mode.
From this shallow pass, we compute per-token statistics and aggregate to per-instance scores (mean over non-padded tokens). We then apply a VaR threshold over the $M$ instance scores to select the subset.
In the training pass, we run the full model on the selected instances and apply token-level loss shaping as given in Algorithm~\ref{alg:eslm}.
The computational overhead from VaR thresholding is minimal, requiring $O(M \log M)$ time per batch (with batch size $M$). This cost is negligible compared to the dominant forward/backward FLOPs of the main model.
We discuss the runtime overhead in Appendix~\ref{app:hardware}.

\paragraph{FLOPs accounting.} Following \citet{kaplan2020scaling, chowdhery2023palm}, a full forward–backward pass costs $6N + 12LHQ T$ FLOPs per token, for an LM with $N$ parameters, $L$ layers, $H$ attention heads, head size $Q$, and sequence length $T$.
In the proxy phase, we run a forward-only early-exit of depth $L' \leq L$, costing $2N_{\text{proxy}} + 6L'HQT$ FLOPs per token, where $N_{\text{proxy}}$ is the parameters for embeddings, layer norms and the first $L'$ transformer blocks. Since the risk scores need logits, we add $2DV$ FLOPs per token for the head projection, with embedding dimension $D$ and vocabulary size $V$.
During token-level loss shaping, masked tokens skip part of the backward in the final layer components (head + FFN matmuls + final/pre-FFN layer norms), yielding $\approx 4DV + 4D + 32D^2 + 4D$ FLOPs saving \emph{per masked token} \citep{kaplan2020scaling}.
\textsc{Eslm}'s optimized FLOPs savings come primarily from $(i)$ dropping whole instances after the proxy pass, which avoids their full forward–backward in the training phase; and $(ii)$ skipping last-block gradient computation for loss-masked tokens.
\looseness -1

\paragraph{Downstream impact.}
Effective pretraining increasingly hinges on how data is selected \citep{tirumala2023d4,mayilvahanan2025llms}.
Improvements in training loss does not guarantee better downstream generalization, particularly under distribution shift \citep{ramanujan2023connection, isik2025scaling}.
\textsc{Eslm} addresses this by providing hierarchical control over which parts of the input receive focus, concentrating optimization to high-risk tokens.
In~\cref{sec:results}, we demonstrate that \textsc{Eslm} improves both loss–vs–compute efficiency and downstream performance than standard training.
\looseness -1
\paragraph{Token vs instance-level selection.}
Unlike methods that filter or reweight entire sequences \citep{wang2024greats, sow2025dynamic}, \textsc{Eslm} uses a \emph{hierarchical} scheme: a cheap proxy pass prunes low-value instances, then loss is shaped at the finer granularity of tokens.
This combines compute-awareness with information-awareness, retaining useful tokens within kept sequences.
\textsc{Eslm} is natively compatible with autoregressive training and avoids data pipeline changes.
Under same compute budget, this hierarchical selection generalizes better than coarse instance-only selection methods (\cref{sec:results}).
\looseness -1

\subsection{\textsc{Ada-ESLM}: Adaptive Confidence Thresholding}
\label{sec:adaptive-eslm}
While a fixed confidence level $\alpha$ in \textsc{Eslm} yields strong efficiency gains (see Section~\ref{sec:results}), the optimal $\alpha$ level may vary throughout training. 
Early in training, broad token coverage may improve generalization, whereas later stages benefit from focusing on harder or more informative tokens. 
To accommodate this, we introduce \textsc{Ada-Eslm}, a dynamic variant that adjusts $\alpha$ during training using a \textit{risk-sensitive controller} driven by CVaR feedback.
In each evaluation step $k$, we compute $\operatorname{CVaR}_{\alpha_k}$ of the per-token risk scores, using (\ref{eq:CVaR-definition}).
We then track the changes in CVaR to detect shifts in training difficulty, estimated from model training dynamics, and update $\alpha$ using a multiplicative rule:
\[
\begin{aligned}
\alpha_{k+1} 
&\leftarrow \alpha_k \cdot \exp\!\left(-\gamma \cdot \Delta_{\text{norm}}(\alpha_k)\right), \\
\text{where } \quad
&\Delta_{\text{norm}}(\alpha_k) 
:= \frac{\operatorname{CVaR}_{\alpha_k} - \operatorname{CVaR}_{\alpha_{k-1}}}
{\operatorname{CVaR}_{\alpha_{k-1}} + \varepsilon}.
\end{aligned}
\]
Here, $\Delta_{\text{norm}}(\alpha_k)$ is a dimension and scale-independent signal capturing the relative change in CVaR, $\gamma > 0$ controls adaptation rate, and $\varepsilon$ is a small constant for numerical stability. 
The core idea for this update rule is \textit{stabilizing} CVaR: if $\Delta_{\text{norm}} > 0$ (i.e., CVaR increases), the model is encountering harder tokens, $\alpha$ is then decreased to include more tokens and expand the training signal. 
Conversely, if $\Delta_{\text{norm}} < 0$, the model is improving on difficult tokens. We increase $\alpha$ to focus learning on high-risk tokens.
\textsc{Ada-Eslm} extends the adversarial game in (\ref{eq:adversarial-game-formulation}) by equipping the masker with a CVaR-driven controller that adapts token sparsity in response to training dynamics, offering a form of curriculum learning.
We provide the \textsc{Ada-Eslm} algorithm in Appendix~\ref{app:ada-eslm} (see~\cref{alg:adaptive-eslm}).
\looseness -1

\subsection{\textsc{ESLM-KD}: Risk-aware Knowledge Distillation}
\label{sec:kd-eslm}
Knowledge distillation transfers knowledge from a teacher model to a student by encouraging the student to match the teacher’s output distribution \citep{buciluǎ2006model, hinton2015distilling}.
In language modeling, \citet{rawat2024little} showed that a small LM supervision improves the training of a much more capable LLM. 
While the standard framework operates over all tokens—typically using sequence- or word-level KL divergence \citep{kim-rush-2016-sequence}—we can utilize \textsc{Eslm} for risk-aware distillation.
\looseness -1

To this end, we provide a use case, \textsc{Eslm-Kd}, with the implementation details presented in Algorithm~\ref{alg:eslm-kd} in Appendix~\ref{app:eslm-kd}.
Specifically, we apply $\operatorname{VaR}_\alpha$ thresholding using a student LM 
to select high-risk tokens, which are then used to compute the KL divergence between teacher and student logits.
The student is trained only on these selected tokens, focusing its capacity on uncertain or error-prone regions.
This strategy is teacher-agnostic, relying on the internal statistics of the student model for selection.
\textsc{Eslm-Kd} generates a sparse supervision signal based on selected tokens, resulting in improved compute and sample efficiency, as further empirically supported in Section~\ref{sec:results}. 
\looseness -1
\section{Related Work}
\label{sec:related-work}
\paragraph{Online data subset selection.}
Efficient data selection is essential for scaling LLM pretraining, where full-corpus training is often prohibitively expensive \citep{albalak2024survey}.
The premise that not all training examples carry equal learning value is long-standing and predates deep learning-scale pretraining: uncertainty sampling in classical active learning selects points the model is least confident about \citep{lewis1995sequential}, and Bayesian active learning frameworks select points expected to maximally reduce model uncertainty \citep{gal2017deep}. This intuition was carried into deep learning-scale training through online hard-example and high-loss selection methods \citep{loshchilov2015online, katharopoulos2018not, jiang2019accelerating}. \textsc{Eslm} builds on this same premise rather than introducing it; our contribution lies in \emph{how} selection is performed and \emph{where} it is applied, detailed below.

While early work on data selection for LLM pretraining focused on static or offline methods—filtering \citep{marion2023moreinvestigatingdatapruning} or scoring examples before training \citep{coleman2020selection, xie2023dataselectionlanguagemodels, wettig2024qurating} or during fine-tuning \citep{xia2024less}—these lack adaptability and struggle to scale in large-batch or continual pretraining.
Online data selection overcomes this by adapting to the model's evolving state, extending example-level high-loss selection \citep{loshchilov2015online, katharopoulos2018not, jiang2019accelerating} with gradient-based criteria \citep{killamsetty2021grad}. Recent works \citep{mindermann2022prioritized, wang2024greats} leverage gradient influence \citep{sachdeva2024train} to guide instance selection, or use reference models for token selection \citep{fan2023irreducible, lin2024rho}; however, they often incur high memory due to expensive gradient computations, or supervision costs from curated references and validation sets. 
Relative to this body of work, \textsc{Eslm} differs along four axes: \textbf{(i) Selection mechanism.} Prior high-loss selection typically uses a fixed top-$k$ or manually calibrated score threshold; \textsc{Eslm} instead applies VaR/CVaR thresholding relative to the \emph{empirical batch score distribution}, so the effective selection ratio adapts automatically as the score distribution shifts over training, without retuning. \textbf{(ii) Supervision.} Unlike methods relying on an external reference model \citep{lin2024rho} or curated validation data and per-sample gradients \citep{wang2024greats}, \textsc{Eslm}'s proxy scores are drawn from the model being trained itself, requiring no auxiliary model or held-out set. \textbf{(iii) Granularity.} Prior selection methods in this space operate at a single granularity (instance or token); \textsc{Eslm} combines both in a hierarchical scheme, first pruning low-value instances via a lightweight proxy, then shaping the loss at the token level within retained instances. \textbf{(iv) Theoretical interpretation.} We connect selection thresholds explicitly to VaR/CVaR risk measures and a DRO interpretation (Section~\ref{sec:eslm}), providing a risk-sensitive account of \emph{why} tail-focused selection should help, rather than motivating difficulty-based selection heuristically.
In contrast, \textsc{Eslm} introduces a novel, \textit{lightweight}, self-supervised, hierarchical batch selection mechanism, eliminating offline preprocessing, external supervision, or costly gradient tracing.
This yields an easily integrable approach that achieves a favorable trade-off between compute efficiency and robustness, while remaining agnostic to training configurations.

\looseness -1

\paragraph{Risk-aversion in language modeling.}
Risk-sensitive optimization offers a principled mechanism to enhance robustness by focusing training on high-risk examples \citep{rockafellar2000optimization}.
The CVaR objective has been previously studied in classification \citep{curi2020adaptive}, submodular optimization \citep{MAEHARA2015526}, and fair learning \citep{williamson2019fairness}, typically to mitigate the influence of tail-risk or worst-case samples.
However, in the context of language modeling, CVaR-based approaches remain underexplored. Notable exceptions include methods \citep{oren2019distributionally} that aggregate losses over 
topics to address distributional shift but these typically operate at the 
group level, or in fine-tuning LLMs with reinforcement learning \citep{chaudhary2024risk}.
On the contrary, \textsc{Eslm} brings risk-aware optimization to the token level for LLM pretraining.
Each batch is shaped into a high-risk sub-distribution by the fine-grained risk control of \textsc{Eslm}, incorporating a distributionally robust view of token-level optimization.
Unlike heuristic loss-based filtering, \textsc{Eslm} offers a theoretically grounded and practical hierarchical input selection method for large-scale pretraining under uncertainty.
\looseness -1

\section{Experiments}
\label{sec:experiments}
In this section, we evaluate two \textsc{Eslm} variants (\myinlinecolorbox{salmon!30}{$\operatorname{CVaR}$-loss} and \myinlinecolorbox{electric-blue!15}{$\operatorname{VaR}$-entropy}) across diverse pretraining settings, assessing the impact on efficiency and generalization with varying model scales, data mixtures, and training budgets.
\paragraph{Experimental setup.} In our experiments, we train on SlimPajama-6B \citep{cerebras2023slimpajama}: a 6B-token mixture across seven domains \{Arxiv, Book, CommonCrawl, C4, GitHub, StackExchange, Wikipedia\}, using both uniform and DoReMi \citep{xie2023doremi} domain weights (see Appendix~\ref{app:corpus} for details). We pretrain GPT-2 models with 124M, 350M, 774M, and 1.5B parameters using a BPE tokenizer \citep{sennrich-etal-2016-neural} with vocabulary size 50,304.
All models are trained with a sequence length of 1024, gradient accumulation over 40 steps, and mini-batch sizes of \{4,8,12\}.
We use AdamW with cosine learning rate decay; full hyperparameters are provided in Appendix~\ref{app:exp-setup}.
We apply \textsc{Eslm} with confidence level $\alpha$ tuned over the set $\{0.05, 0.1, 0.2\}$, with ablations presented in Section~\ref{sec:ablations}.

\paragraph{Baselines.} We compare \textsc{Eslm} variants against regular training and online batch selection methods: $\textbf{(1)}$ CLM, introduced in Section~\ref{sec:background}, $\textbf{(2)}$ Rho-1 \citep{lin2024rho}, an online SLM using a reference model to score token loss differentials, $\textbf{(3)}$ GREATS \citep{wang2024greats}, a state-of-the-art online sample selection method based on high-quality validation data and per-sample gradients.
GREATS’ high memory requirements, even with ghost inner product optimizations, limited our comparisons to the 124M setting.
For distillation experiments (Section~\ref{sec:results}), we further compare against the dense distillation and SALT \citep{rawat2024little} methods.
We provide the baseline details in Appendix~\ref{app:baselines}.
\looseness -1
 \begin{figure*}[t]
    \centering  
    \begin{subfigure}[t]{0.48\linewidth}\centering{\includegraphics[width=1\linewidth,trim=0 0 0 0,clip]{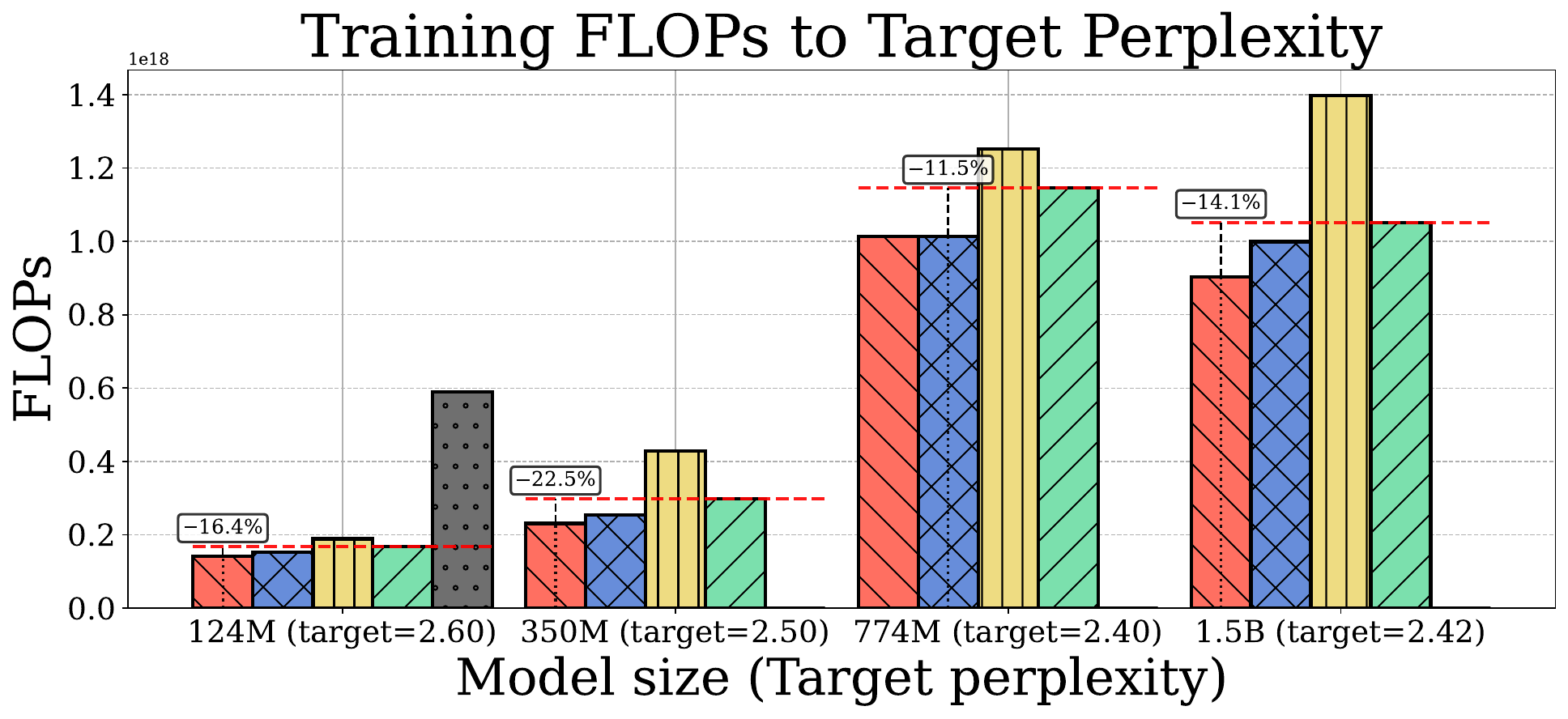}}
    \caption{SlimPajama-6B-Unif.}
    \end{subfigure}%
    \begin{subfigure}[t]{0.48\linewidth}\centering{\includegraphics[width=1\linewidth,trim=0 0 0 0,clip]{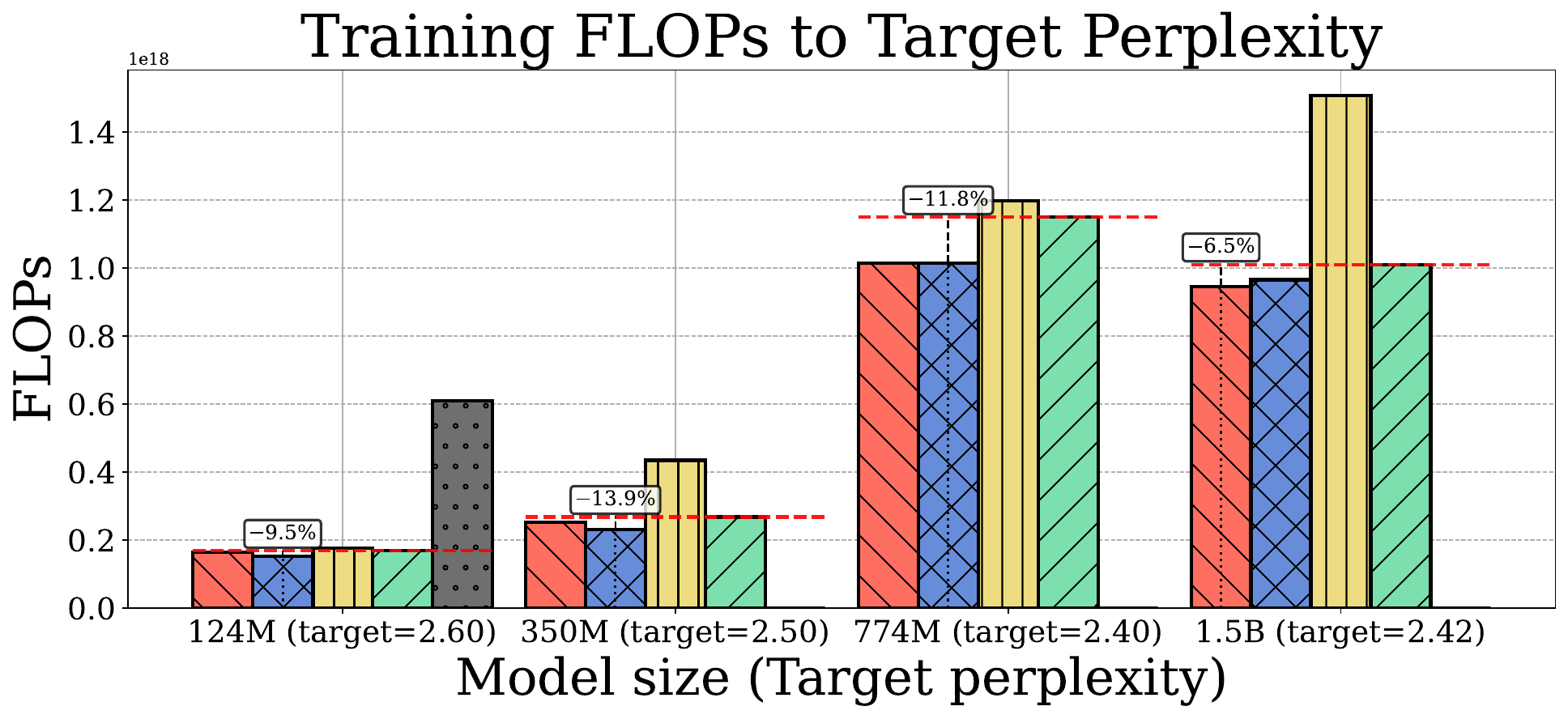}}
    \caption{SlimPajama-6B-DoReMi.}
    \end{subfigure}%
    \\
    \begin{subfigure}{0.8\linewidth}\centering{\includegraphics[width=1\linewidth,trim=30 10 10 10,clip]{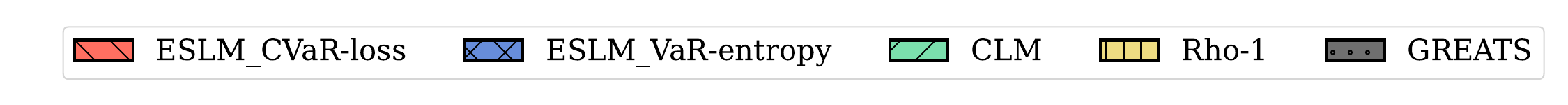}}
    \end{subfigure}%
    \setlength{\belowcaptionskip}{-1pt}
    \caption{\textbf{Training FLOPs ($\downarrow$) required for convergence to target validation (log) perplexity.} We report training FLOPs required by the \{124M, 350M, 774M, 1.5B\} parameter models to converge to a target validation loss threshold across datasets. The \% labels show the percentage FLOPs savings provided by \textsc{Eslm} relative to the best performing baseline. \textsc{Eslm} reduces training cost by focusing optimization on the high-risk instances and tokens and eliminating redundant gradient computation.
    }
\label{fig:flops-required-for-target-loss}
\end{figure*}
 \begin{figure*}[t]
    \centering  
    \begin{subfigure}[t]{0.25\linewidth}\centering{\includegraphics[width=1\linewidth,trim=0 0 0 0,clip]{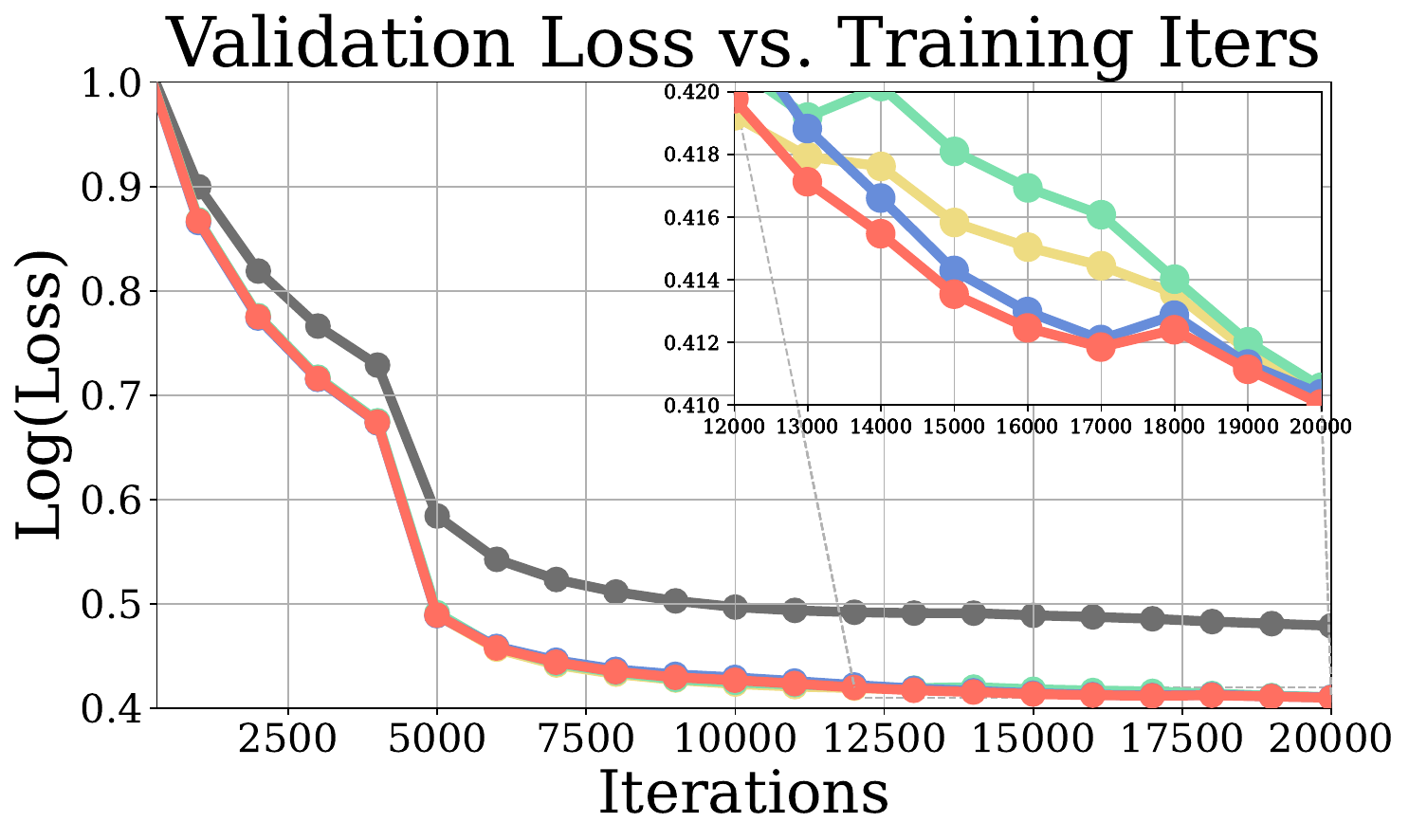}}
    \caption{124M (Iterations).}
    \end{subfigure}%
    \begin{subfigure}[t]{0.25\linewidth}\centering{\includegraphics[width=1\linewidth,trim=0 0 0 0,clip]{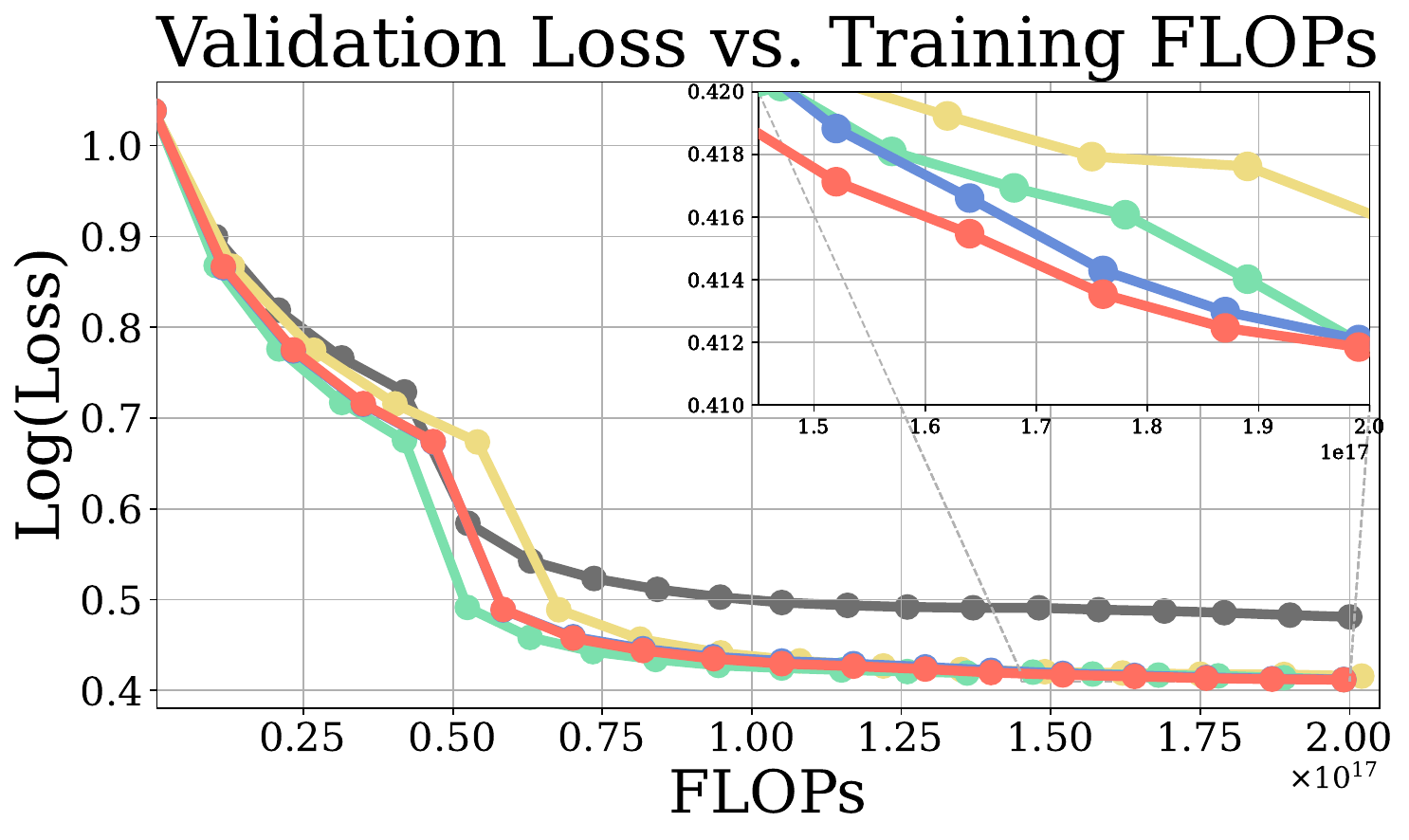}}
    \caption{124M (FLOPs).}
    \end{subfigure}%
    \begin{subfigure}[t]{0.25\linewidth}\centering{\includegraphics[width=1\linewidth,trim=0 0 0 0,clip]{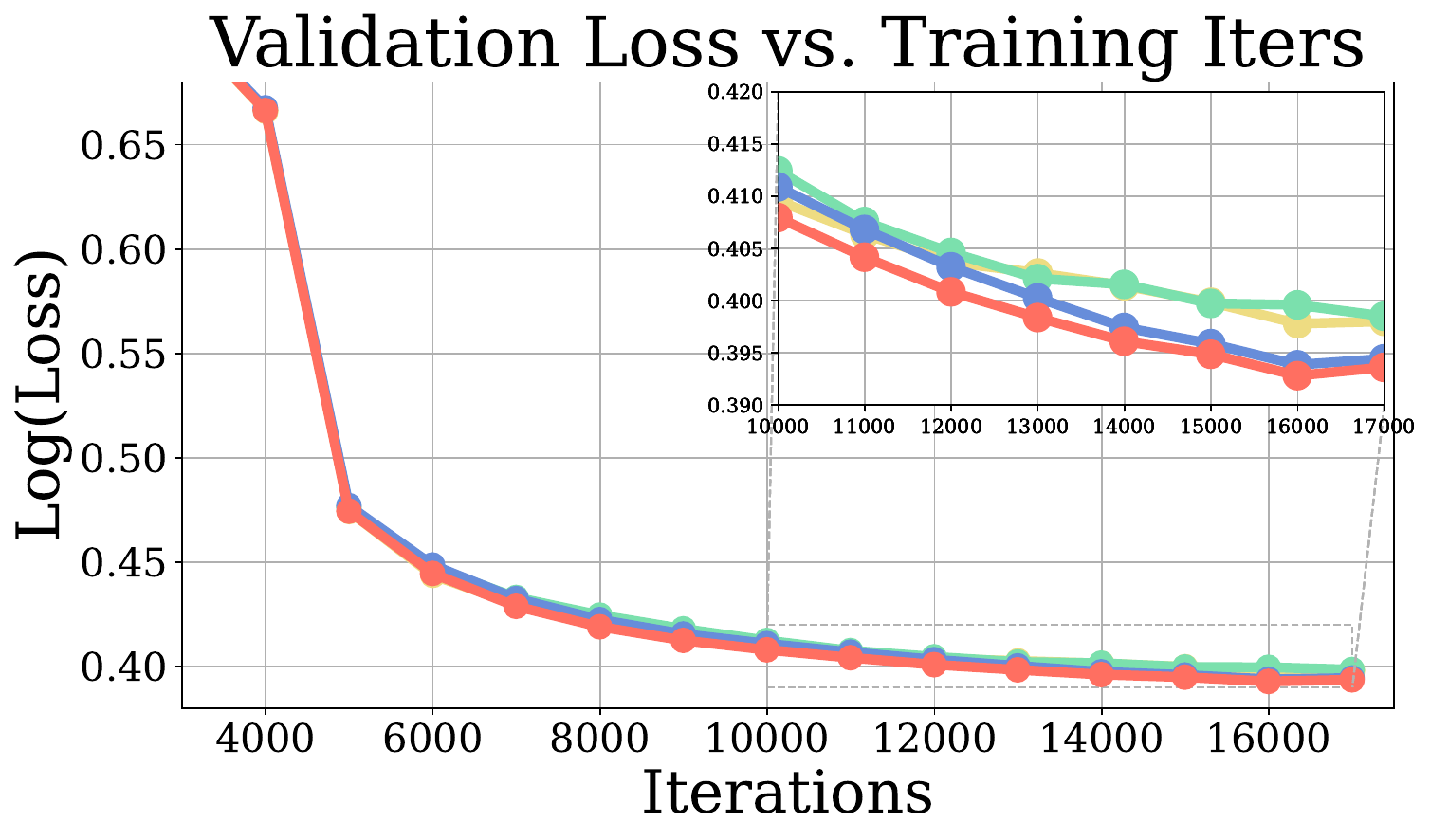}}
    \caption{350M (Iterations).}
    \end{subfigure}%
    \begin{subfigure}[t]{0.25\linewidth}\centering{\includegraphics[width=1\linewidth,trim=0 0 0 0,clip]{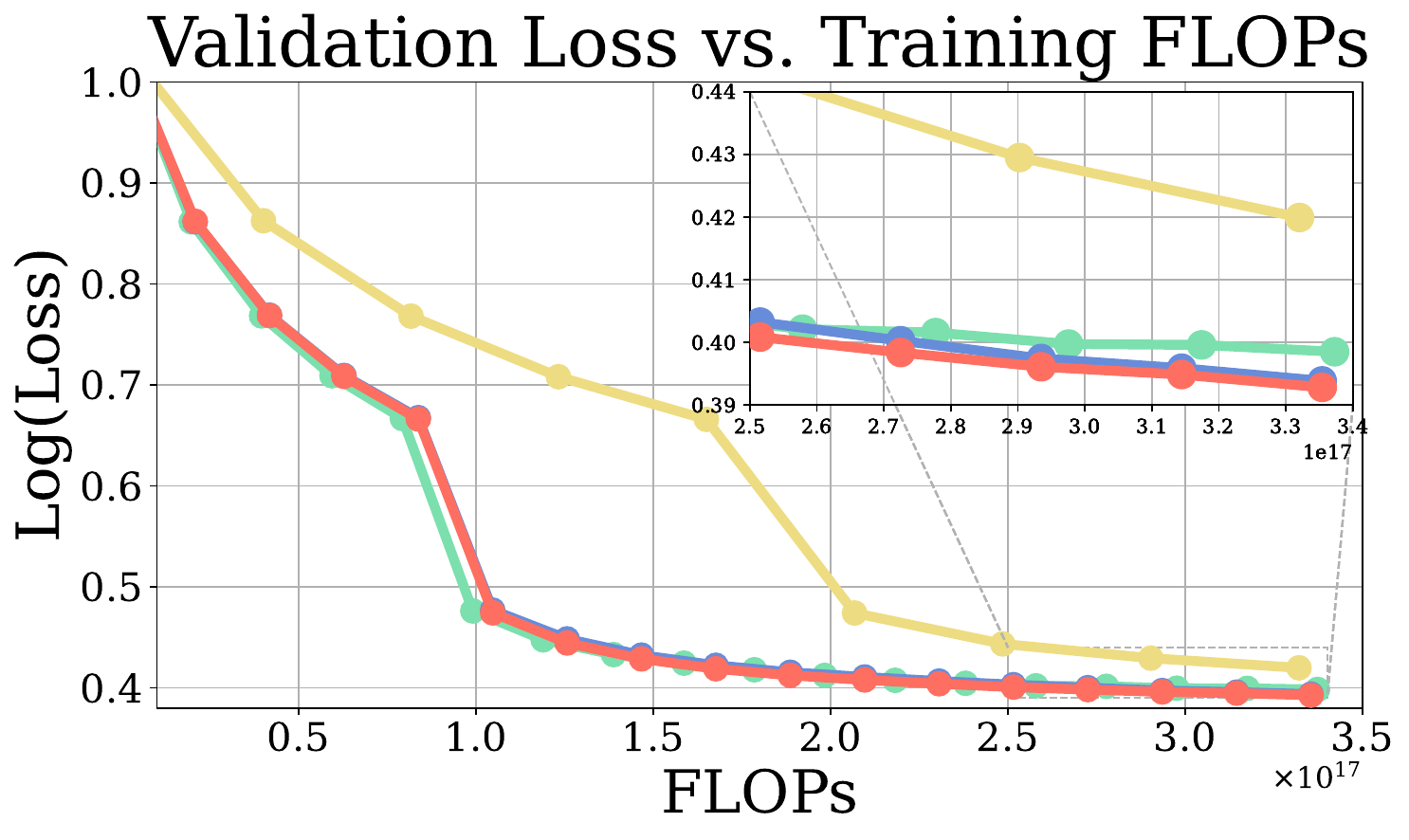}}
    \caption{350M (FLOPs).}
    \end{subfigure}%
    \\
    \begin{subfigure}{0.8\linewidth}\centering{\includegraphics[width=1\linewidth,trim=180 20 160 20,clip]{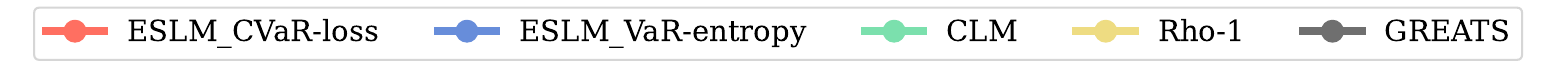}}
    \end{subfigure}%
    \setlength{\belowcaptionskip}{-1pt}
    \caption{\textbf{Validation loss convergence.} We report convergence of validation loss versus training FLOPs/iterations of models trained on SlimPajama-6B-Unif mixture. \textsc{Eslm} variants provide faster convergence to lower loss values in terms of iterations, while requiring fewer FLOPs than baselines at the convergence. See Appendix~\ref{app:val-loss-vs-flops-results} for additional results on other pretraining corpora.
    }
\label{fig:val-loss-vs-flops}
\end{figure*}
\paragraph{Performance metrics.}
We assess our method concerning training efficiency and generalization ability by tracking the metrics: $(i)$ training FLOPs to converge to target validation perplexity as FLOPs are model-agnostic proxy for compute cost \citep{hoffmann2022training, chowdhery2023palm}, $(ii)$ validation loss versus training FLOPs/iterations, and $(iii)$ zero-/few-shot accuracy in downstream benchmarks from the \texttt{lm-eval-harness} \citep{eval-harness} suite, spanning QA, reasoning, and generation. 
Convergence curves use a running average with a window size of 5 evaluation points.
We evaluate performance across model sizes and dataset mixtures with FLOPs estimated as in Section~\ref{sec:eslm}.
Full metric and experimental setup details are in Appendix~\ref{app:experiment-details}.
\subsection{Experimental Results}
\label{sec:results}
In this section, we report the performance of \textsc{Eslm} against baselines, followed by the results of its implementation as a knowledge distillation mechanism and ablation analyses.
\begin{figure}
\centering
\includegraphics[width=0.45\linewidth]{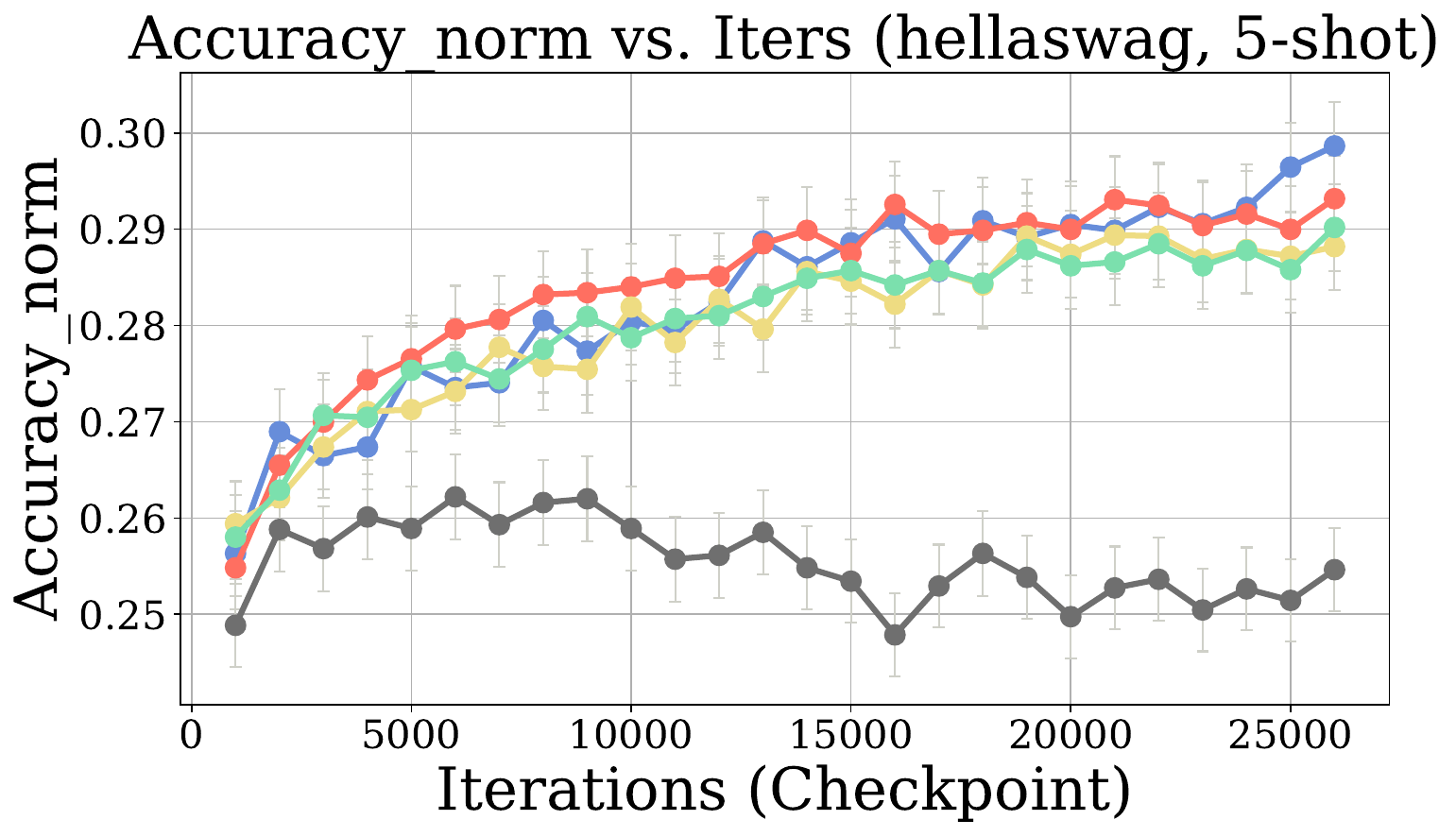} \\
\includegraphics[width=0.4\linewidth,trim=200 25 510 25,clip]{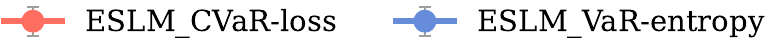}
\includegraphics[width=0.35\linewidth,trim=595 25 170 25,clip]{figures/legend_general.pdf}
\caption{5-shot accuracy (norm) ($\uparrow$) performance on HellaSwag throughout training. \textsc{Eslm} variants discover higher accuracy levels than baselines, with particular gains in the later training stages.
\looseness -1
}
\setlength{\belowcaptionskip}{-5pt}
\setlength{\abovecaptionskip}{-0.2pt}
\vspace{-1\baselineskip}
\label{fig:accuracy-vs-flops-hellaswag}
\end{figure}
\paragraph{Validation loss vs FLOPs/iterations.}
As presented in~\cref{fig:flops-required-for-target-loss}, \textsc{Eslm} consistently requires fewer training FLOPs to reach target validation loss across model sizes and datasets. \textsc{Eslm} provides strong efficiency gains at convergence, reaching the level of, e.g., $-22.5\%$ reduced FLOPs relative to standard training in a 350M setting.
Notably, these gains over CLM are essentially free, i.e., \textsc{Eslm} operates under the same optimizer, dataset, with no external supervision.
Figure~\ref{fig:val-loss-vs-flops} further shows that \textsc{Eslm} accelerates learning convergence compared to the baselines, discovering lower validation loss with fewer iterations. 
While the proxy phase adds a small upfront cost, subsequent instance pruning and token-level loss shaping drive lower losses and net FLOPs savings at convergence across scales compared to CLM.
Unlike Rho-1, which queries an external reference model—adding extra compute overhead and requiring high-quality data—\textsc{Eslm} instead leverages model-internal training dynamics for selection, avoiding such offline preprocessing.
Similarly, while GREATS employs an efficient ghost inner product approximation~\citep{wang2024greats}, it still relies on curated validation data and per-sample gradient estimates, which are impractical at larger model scales due to high memory demands.
In contrast, \textsc{Eslm} operates without gradient tracing and scales naturally.
Moreover, GREATS performs selection at the instance level, often discarding informative tokens within partially useful sequences and leading to worse perplexity.
This highlights the instance and token-level granularity of \textsc{Eslm}, which avoids this limitation by emphasizing valuable sub-sequence information.
\looseness -1
\begin{table*}[ht]
\caption{\textbf{Generalization performance on downstream tasks.} All models (124M) are pretrained under a $\sim$3E17 FLOPs budget on SlimPajama-6B-Unif mixture. We report the best observed accuracy$_{( \text{standard error})}$ or exact match if provided, during training.
\myinlinecolorbox{tableyellow}{\textbf{Highlighted}} values indicate the best performance.
See~\cref{app:additional-downstream-performance} for the results under various model sizes and datasets.
\looseness -1
}
\label{tab:downstream-perf-124m-unif} 
\centering
    \resizebox{1\linewidth}{!}{%
        \begin{tabular}{l|c|cccccccc} \specialrule{1.5pt}{1pt}{1pt}
         \multicolumn{1}{c}{\makecell{\\  \\ \textbf{Benchmark}}} & \multicolumn{5}{c}{\textbf{Method (124M)}} \\ \cmidrule(lr){2-7}
        & \# Shots &  \multicolumn{1}{c}{\textsc{Eslm}-\text{\myinlinecolorbox{salmon!30}{$\operatorname{CVaR}$-loss}}} & \multicolumn{0}{c}{\textsc{Eslm}-\myinlinecolorbox{electric-blue!15}{$\operatorname{VaR}$-entropy}} & \multicolumn{0}{c}{CLM} & \multicolumn{0}{c}{Rho-1} & \multicolumn{0}{c}{GREATS} \\ 
        \specialrule{1.5pt}{1pt}{1pt}
        ARC-E \citep{clark2018think} & 0-shot &   $0.3712_{(0.0099)}$  & \tabye  $0.3766_{(0.0099)}$  & $0.3644_{(0.0099)}$ &   $0.3657_{(0.0099)}$  &   $0.3236_{(0.0096)}$ \\
        LAMBADA \citep{paperno-EtAl:2016:P16-1} & 5-shot & $0.1595_{(0.0051)}$  & \tabye $0.1721_{(0.0053)}$   &   $0.1701_{(0.005)}$   &  $0.1680_{(0.005)}$ & $0.0254_{(0.002)}$ \\
         SciQ \citep{welbl2017crowdsourcing} & 5-shot 
 & \tabye $0.714_{(0.0143)}$  &  $0.71_{(0.0144)}$  &  $0.6970_{(0.0145)}$ &  $0.7000_{(0.0145)}$ & $0.4350_{(0.0157)}$  \\
        HellaSwag \citep{zellers2019hellaswag} &  5-shot &  $0.2930_{(0.0045)}$  & \tabye $0.2964_{(0.0046)}$  &  $0.2901_{(0.0045)}$ &  $0.2893_{(0.0045)}$ &  $0.2621_{(0.0044)}$ \\
        TriviaQA \citep{2017arXivtriviaqa} & 1-shot & \tabye $0.0128_{(0.0001)}$ &  $0.0066_{(0.0006)}$ & $0.0078_{(0.0007)}$ &   $0.0090_{(0.0007)}$ & $0.0007_{(0.0002)}$ \\
        COPA \citep{NEURIPS2019_4496bf24} & 5-shot &  \tabye $0.64_{(0.0482)}$ &  $0.62_{(0.0488)}$ &  $0.62_{(0.0488)}$ & $0.62_{(0.0488)}$ & $0.64_{(0.0482)}$\\
        MultiRC \citep{NEURIPS2019_4496bf24} & 5-shot & \tabye $0.5633_{(0.0071)}$ &  $0.5429_{(0.0072)}$ & $0.5338_{(0.0072)}$ &   $0.5338_{(0.0072)}$ &    $0.5497_{(0.0071)}$ \\
        OpenBookQA \citep{mihaylov2018can} & 5-shot &  $0.164_{(0.0166)}$  & $0.17_{(0.0168)}$  &  \tabye $0.174_{(0.017)}$  &  $0.164_{(0.0166)}$ & $0.148_{(0.0159)}$ \\
        PiQA \citep{bisk2020piqa}  & 5-shot & \tabye $0.6240_{(0.0113)}$ &  $0.6191_{(0.0113)}$  &  $0.6099_{(0.0114)}$ &   $0.6180_{(0.0113)}$ & $0.5571_{(0.0116)}$ \\
        \specialrule{1.5pt}{1pt}{1pt}
        \multicolumn{1}{c}{\textbf{Average} ($\uparrow$)} &    & \tabye \textbf{0.39353} & 	 \textbf{0.39041}	& 0.38523	& 0.38531 & 0.32684 \\
        \specialrule{1.5pt}{1pt}{1pt}
    \end{tabular}
    }
\end{table*}
\paragraph{Downstream performance.} 
Table~\ref{tab:downstream-perf-124m-unif} summarizes the best zero-/few-shot accuracy achieved by 124M models trained on SlimPajama-6B-Unif under $\sim$3E17 FLOPs compute budget. 
\textsc{Eslm} variants significantly outperform baselines in average accuracy, with consistent gains over GREATS and Rho-1 across all tasks.
Figure~\ref{fig:accuracy-vs-flops-hellaswag} further illustrates the accuracy convergence on the HellaSwag benchmark, where \textsc{Eslm}-\myinlinecolorbox{salmon!30}{$\operatorname{CVaR}$-loss} achieves faster early gains, while \textsc{Eslm}-\myinlinecolorbox{electric-blue!15}{$\operatorname{VaR}$-entropy} surpasses baselines in later training stages.
These results, including additional evaluations in~\cref{app:additional-downstream-performance}, show that \textsc{Eslm} improves generalization.

\begin{figure*}[ht]
    \centering  
    \begin{subfigure}[t]{0.4\linewidth}\centering{\includegraphics[width=1\linewidth,trim=0 0 0 0,clip]{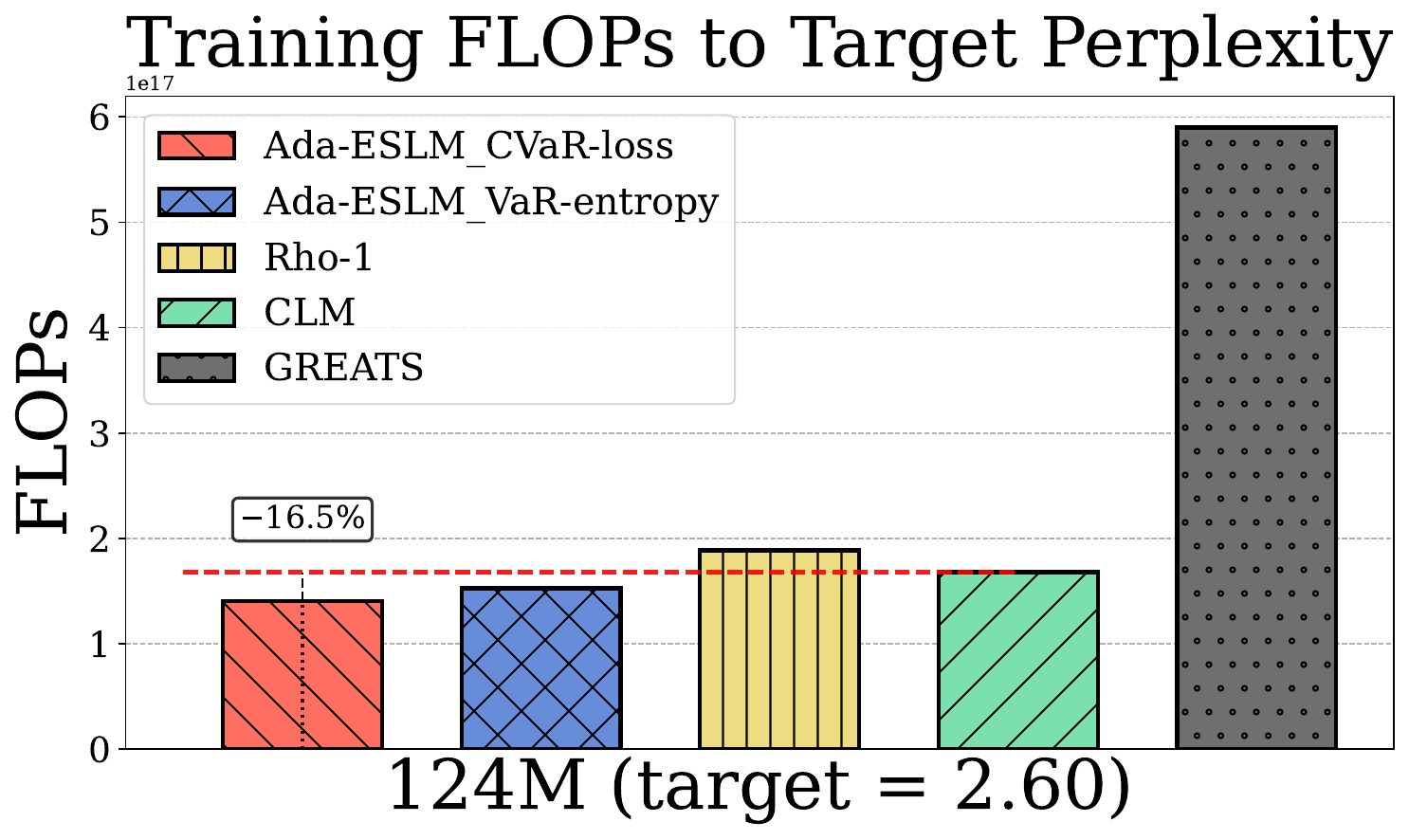}}
    \caption{\textsc{Ada-Eslm}.}
    \label{fig:ada-eslm-result}
    \end{subfigure}%
    \begin{subfigure}[t]{0.4\linewidth}\centering{\includegraphics[width=1\linewidth,trim=0 0 0 0,clip]{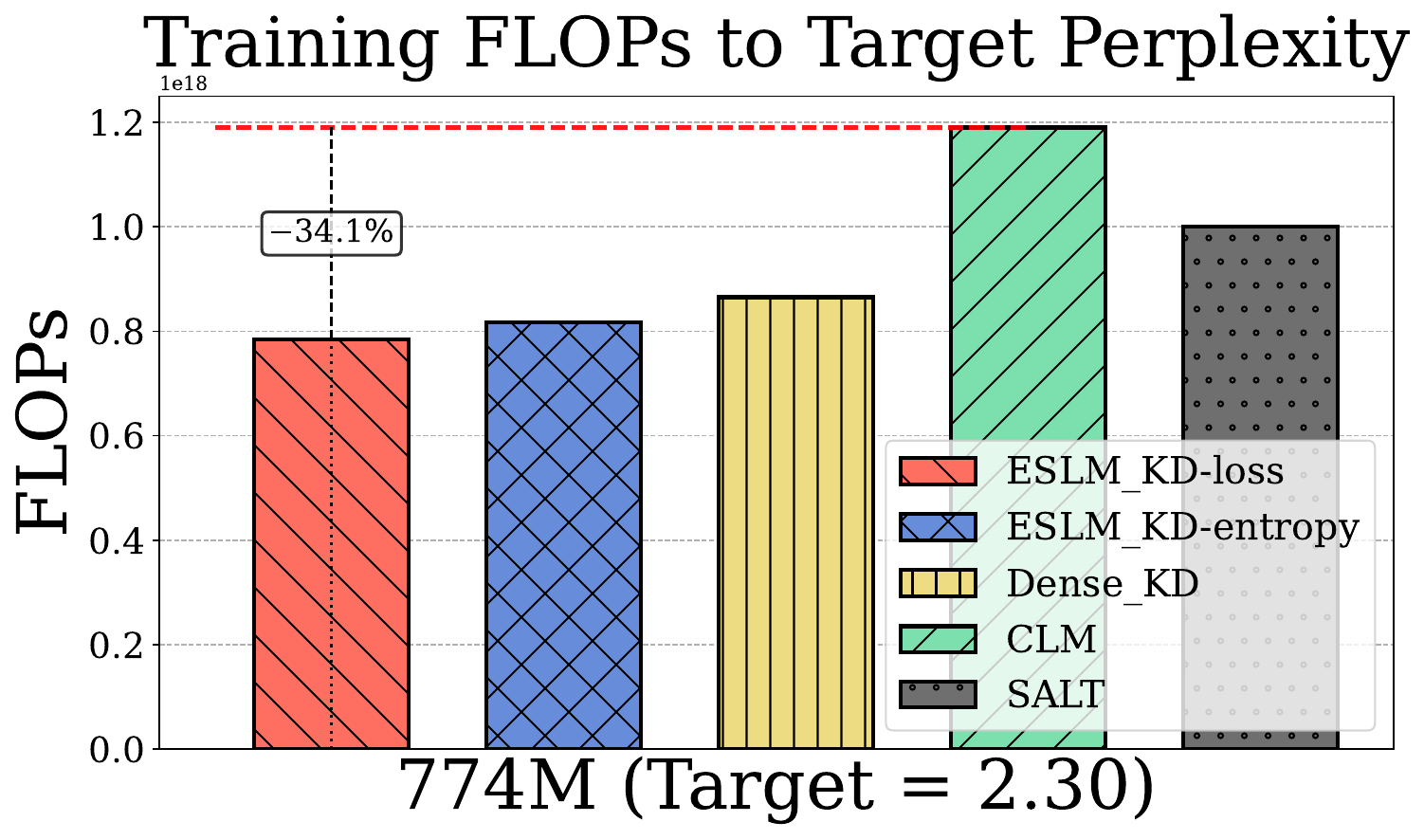}}
    \caption{Knowledge distillation.}
    \label{fig:eslm-kd-result}
    \end{subfigure} \\
    \begin{subfigure}[t]{0.4\linewidth}\centering{\includegraphics[width=1\linewidth,trim=0 0 0 0,clip]{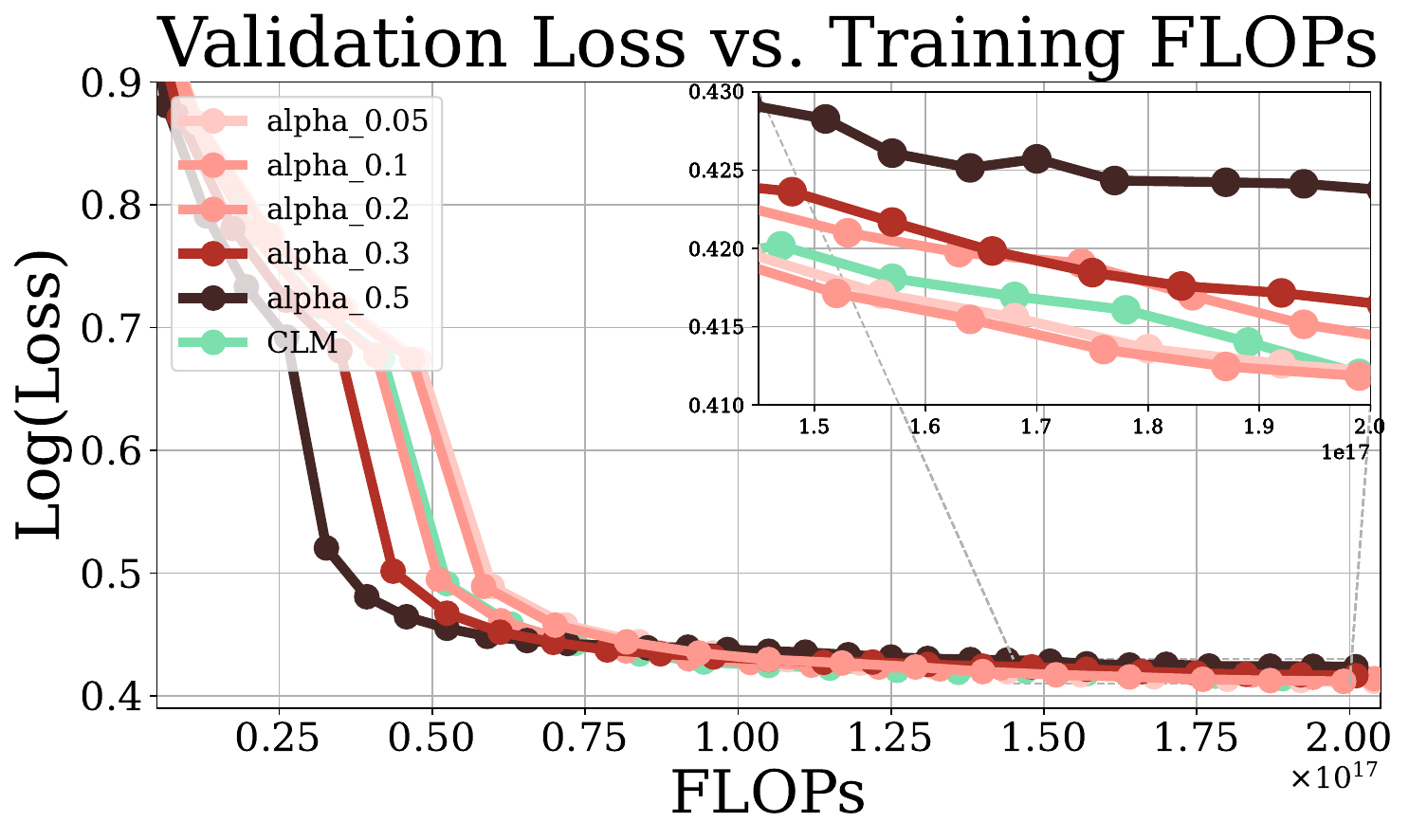}}
    \caption{$\alpha$ ablation.}
    \label{fig:alpha-ablation}
    \end{subfigure}%
    \begin{subfigure}[t]{0.4\linewidth}\centering{\includegraphics[width=1\linewidth,trim=0 0 0 0,clip]{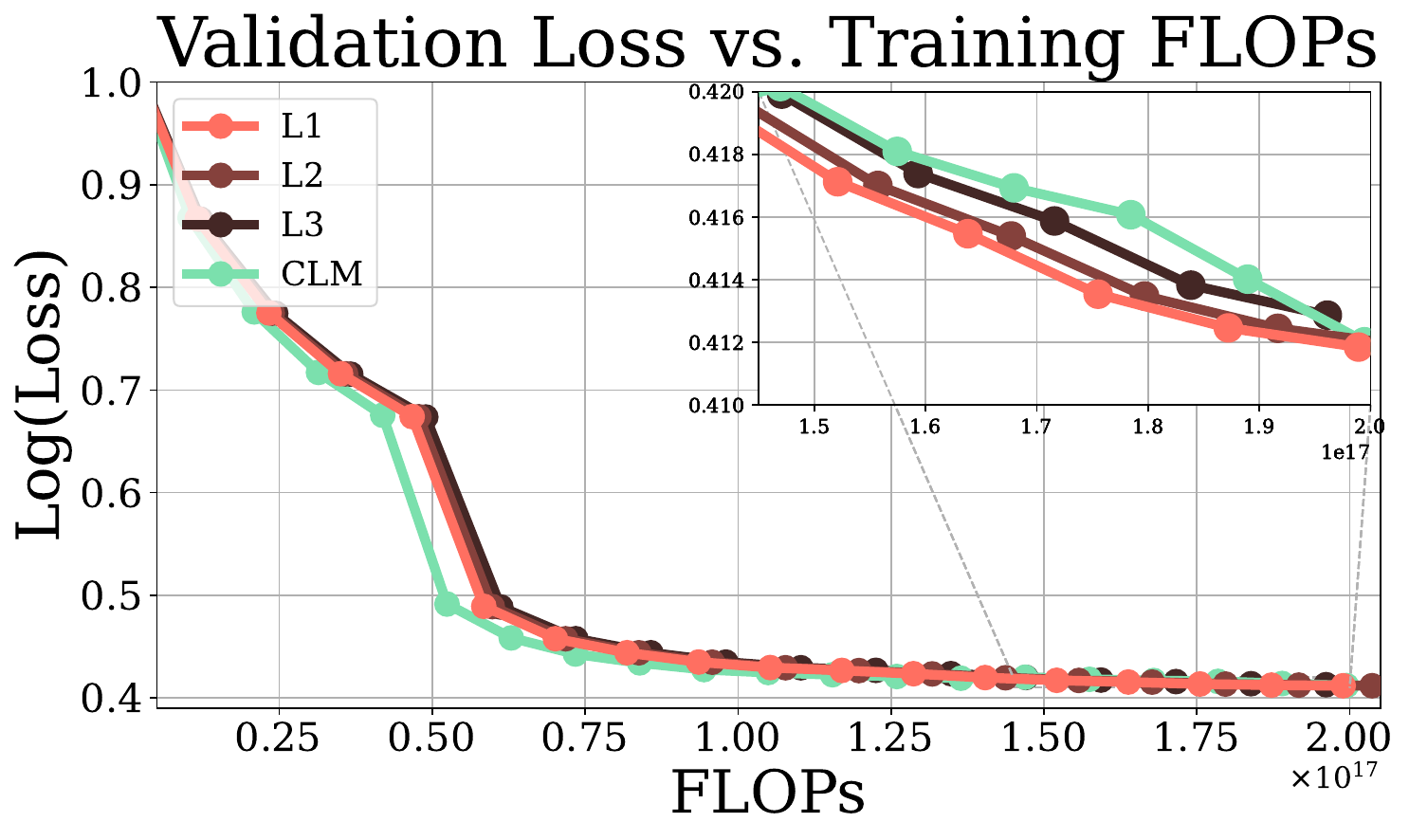}}
    \caption{$L$ ablation.}
    \label{fig:proxy-L-ablation}
    \end{subfigure}%
    \setlength{\belowcaptionskip}{-7pt}
    \caption{\textbf{Extended analyses.} 
    \textbf{(a):} \textsc{Ada-Eslm} reduces FLOPs for convergence to target validation loss ($\downarrow$) by adaptively tuning $\alpha$ based on training dynamics.
    \textbf{(b):} For 774M knowledge distillation, \textsc{Eslm-KD} converges to target validation loss with substantially fewer FLOPs.
    \textbf{(c):} Adjusting $\alpha$ enables flexible control over the trade-off between training efficiency and model quality.
    \textbf{(d):} Proxy depth ablation shows shallow proxies match deeper ones in convergence, indicating minimal proxy computation is sufficient for effective instance selection.
    \looseness -1
    }
\label{fig:additional-analyses}
\end{figure*}

\paragraph{\textsc{Ada-ESLM} experiments.}
We train \textsc{Ada-Eslm} (124M) with $\alpha_0 = 0.1, \gamma = 0.5$ on SlimPajama-6B-Unif.
Figure~\ref{fig:ada-eslm-result} reveals that \textsc{Ada-Eslm} achieves the target validation perplexity with $-16.5\%$ training FLOPs savings at convergence. As detailed in Appendix~\ref{app:ada-eslm} (Figure~\ref{fig:ada-eslm-additional-results}), \textsc{Ada-Eslm} provides an implicit curriculum learning: the training process begins with broader input coverage and gradually shifts focus toward higher-risk tokens, without manual scheduling or external supervision. 
Adaptively adjusting $\alpha$ based on CVaR feedback stabilizes training while offering a principled trade-off between compute-efficiency and generalization.
Downstream evaluations in Figure~\ref{fig:ada-eslm-flops-and-generalization} and Table~\ref{tab:downstream-perf-ada-eslm-124M} (Appendix~\ref{app:ada-eslm}) confirm that \textsc{Ada-Eslm} further achieves higher average downstream accuracy than baselines, improving generalization while maintaining high training efficiency.

\paragraph{Experiments for knowledge distillation.}
To utilize \textsc{Eslm} for risk-aware knowledge distillation (\cref{sec:kd-eslm}), we pretrain a 774M student from a 124M teacher on SlimPajama-6B-Unif.
We set the distillation weight $\lambda = 0.5$ and teacher temperature $\rho = 1.0$ (see Appendix~\ref{app:eslm-kd} for details).
We compare \textsc{Eslm-Kd} against CLM, dense distillation without token selection, and SALT \citep{rawat2024little}, a two-stage distillation-then-pretraining pipeline.
As shown in Figure~\ref{fig:eslm-kd-result}, \textsc{Eslm-Kd} models converge to the target validation perplexity with substantially fewer FLOPs; with $-34\%$ FLOPs savings compared to CLM.
Furthermore, as reported in Appendix~\ref{app:eslm-kd} (Table~\ref{tab:downstream-perf-124m-unif-distillation}), it outperforms baseline models in downstream tasks, demonstrating the effectiveness of \textsc{Eslm} for efficient and generalizable distillation.
\looseness -1

\subsection{Ablation and Additional Analyses} 
\label{sec:ablations} 
\begin{itemize}[left=0.2cm] 
\item \textbf{Confidence level ($\alpha$):}
In~\cref{fig:alpha-ablation}, we assess the sensitivity to varying $\alpha$ values: $\{0.05, 0.1, 0.2, 0.3, 0.5\}$.
Lower $\alpha$ values improve data coverage but increase computation, whereas higher $\alpha$ levels enhance efficiency at the cost of underutilization.
We find $\alpha \in [0.1, 0.2]$ offers a favorable trade-off between compute savings and generalization.
\item \textbf{Proxy depth ($L$):}
In~\cref{fig:proxy-L-ablation}, we assess \textsc{Eslm} under varying $L$ levels: $\{1,2,3\}$ for a 124M model.  \textsc{Eslm} achieves similar validation loss convergence, hence, proxy computation with $L=1$ suffices for effective instance selection while minimizing overhead costs.
\item \textbf{Model size:}
Across 124M, 350M, 774M, and 1.5B GPT-2 models, \textsc{Eslm} consistently improves efficiency and generalization (see Figures~\ref{fig:flops-required-for-target-loss}, \ref{fig:val-loss-vs-flops}, \ref{fig:accuracy-vs-flops-hellaswag}, and Appendix~\ref{app:additional-experiments}).
\looseness -1
\item \textbf{Pretraining corpus:}
We evaluate \textsc{Eslm} on SlimPajama-6B corpus with different domain mixture weights (uniform and DoReMi).
The method generalizes well across corpora (see Figure~\ref{fig:flops-required-for-target-loss}) without requiring domain-specific tuning.
Detailed results are reported in Appendix~\ref{app:additional-experiments}.

\item \textbf{Token selection analysis:}
To better understand the behavior of \textsc{Eslm} variants, we analyze the selected tokens across different domains in the Appendix~\ref{app:token-selection-analysis}, which reveals that \textsc{Eslm} focuses on rare, or contextually ambiguous tokens, supporting its risk-aware design.
\item \textbf{Instance vs token-level selection:} In Appendix \ref{app:ablation-flops-savings}, we analyze \textsc{Eslm}'s FLOPs savings into contributions from proxy instance selection and token-level loss shaping.
\looseness -1

\item \textbf{Robustness to label noise:} In Appendix \ref{app:robustness-label-noise}, we further show the utility of \textsc{Eslm}'s hierarchical
risk-aware filtering, not only in improving efficiency and generalization but also in enhancing robustness to label noise.
\end{itemize}
\paragraph{Limitations.} While \textsc{Eslm} offers an effective approach to selective pretraining, it inherently trades completeness for efficiency by sparsifying backpropagation. Although this provides compute savings (FLOPs), in practice, it may underutilize the full training signal. Importantly, these FLOPs savings do not currently translate into wall-clock speedups: on our hardware, \textsc{Eslm} requires more wall-clock time than standard CLM training to reach the same target validation loss (e.g., 9.9--11.7 hours for \textsc{Eslm} variants vs.\ 5.3 hours for CLM at the 124M scale; Appendix~\ref{app:hardware}, Table~\ref{tab:runtime-comp}), since the irregular, fragmented backpropagation induced by instance and token selection is not well-matched to the dense tensor operations that current accelerators are optimized for. We emphasize, however, that \textsc{Eslm} remains substantially faster in wall-clock time than the other selective-training baselines we compare against---nearly $2\times$ faster than Rho-1 and over $8\times$ faster than GREATS at the same scale (Table~\ref{tab:runtime-comp})---hence the practical overhead relative to CLM should be read as a cost of selective pretraining as a class of methods, rather than one specific to our approach. As discussed in Appendix~\ref{app:hardware}, integrating \textsc{Eslm} with sparsity-aware accelerators could further improve utilization and close this gap. Consistent with recent efficiency works \citep{lasby2024srigl}, achieving practical wall-clock acceleration from FLOPs savings requires optimized hardware support for sparse backpropagation, an engineering direction beyond the scope of this paper.

\looseness -1
\section{Conclusion}
\label{sec:conclusion}
We introduce \textsc{Eslm}, a novel selective language modeling with risk-aware hierarchical batch selection, which improves compute efficiency and generalization of LLM pretraining.
Instead of uniform token updates, \textsc{Eslm} applies a two-phased VaR threshold to prioritize high-utility tokens and skip redundant ones during backpropagation. 
This data-centric strategy effectively improves loss-per-FLOP efficiency, without modifying the model, optimizer, or dataset. 
By focusing optimization on the most informative inputs, \textsc{Eslm} enhances generalization and scalability. As a future work, \textsc{Eslm}—along with its \textsc{Ada-Eslm} variant and integration with knowledge distillation—opens new directions in risk-aware token-level curriculum learning, adaptive compute allocation, and risk-aware data valuation for sustainable and efficient LLM scaling.
\looseness -1

\section*{Acknowledgements}

This research was supported by the Max Planck \& Amazon Science Hub. We also thank the German Research Foundation for the support.

\section*{Broader Impact Statement}
\label{sec:ethics-statement}
Our work introduces \textsc{Eslm}, a selective language modeling framework that improves training efficiency and generalization performance in LLM pretraining via risk-aware instance and token selection.
On the one hand, \textsc{ESLM} enables compute-efficient training by focusing optimization on the most informative parts of the input.
This could reduce the energy footprint of large-scale training runs, make LLM development more accessible to institutions with limited compute budgets, and improve model robustness, particularly in out-of-distribution scenarios.

On the other hand, improved training efficiency may accelerate the development of powerful generative models, some of which could be misused for disinformation, synthetic media, or other harmful applications. In addition, token-level filtering methods—if miscalibrated—may reinforce spurious patterns or underrepresent minority language phenomena, inadvertently encoding or amplifying societal biases in the training data.
Although \textsc{Eslm} is not tied to a specific application, its performance gains could boost the downstream impact of any application built upon the pretrained models.
As our method provides purely training-time improvement, it does not increase model capacity or inference capability directly, which partially limits its risk surface.

Finally, our study does not involve intervention with or collection of data from human participants.
Our experiments use widely adopted language-model pretraining corpora or mixtures assembled from publicly available sources under their respective licenses. Corresponding references are provided.
\looseness -1

\bibliography{refs}
\bibliographystyle{tmlr}

\appendix
\section*{Appendix}

\section{Ada-ESLM: Adaptive Confidence Thresholding}
\label{app:ada-eslm}

In Algorithm~\ref{alg:adaptive-eslm}, we provide the algorithmic description of \textsc{Ada-Eslm}.
Instead of using a fixed confidence level $\alpha$ throughout training, \textsc{Ada-Eslm} introduces a feedback-driven update mechanism that adjusts $\alpha$ based on the evolving difficulty of the training process. The underlying principle is to achieve a steady state through stabilizing the CVaR signal over time, allowing the model to gradually shift from broad token coverage to a more focused, high-risk subset.
When CVaR increases over training intervals, it signals that the model is encountering more difficult (high-risk) examples, requiring a broader coverage. Conversely, a decrease in CVaR suggests that the model is improving on difficult tokens and can afford to focus more narrowly.

\begin{figure}[ht]
    \centering
    \includegraphics[width=0.5\linewidth]{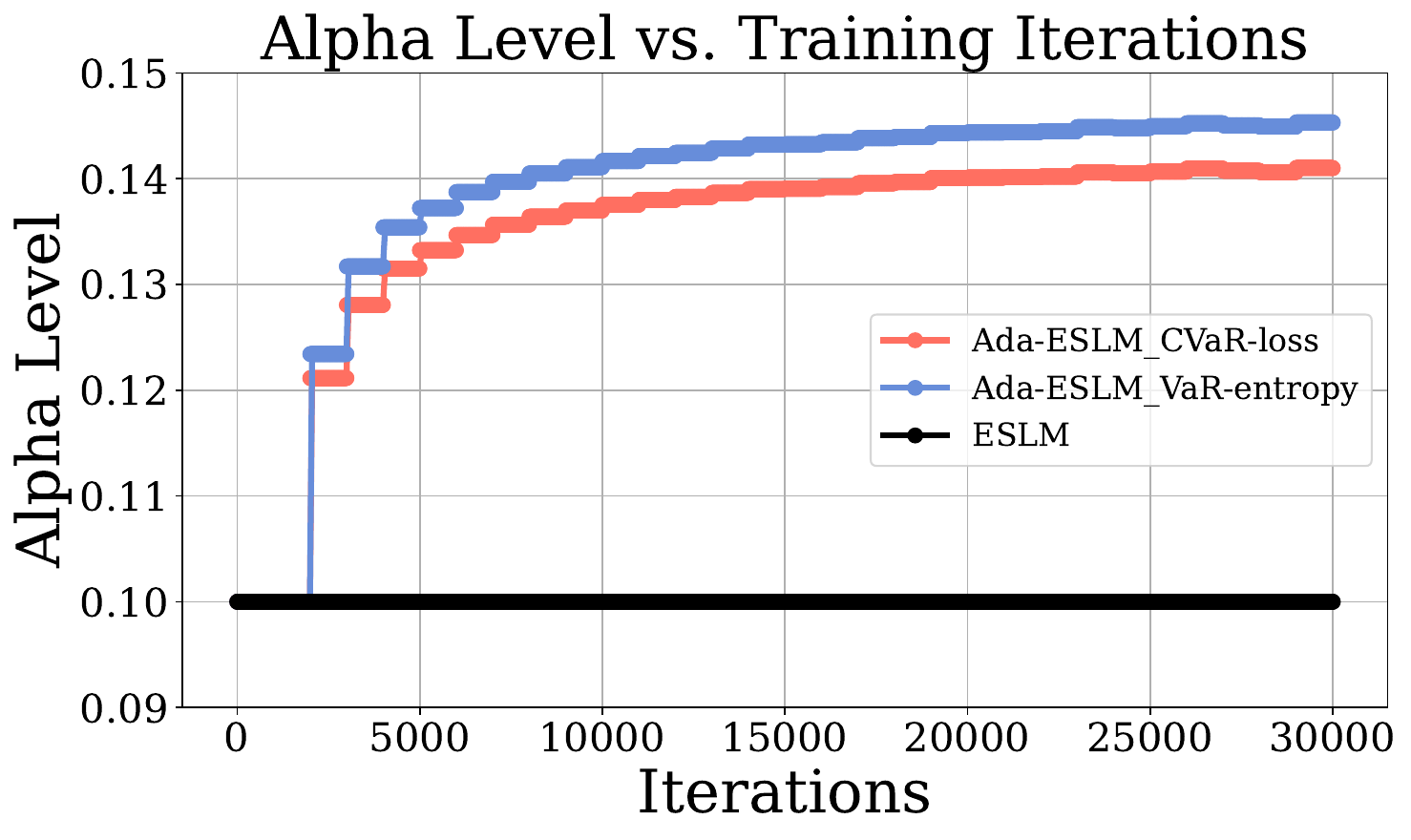}
    \caption{\textbf{\textsc{Ada-ESLM} confidence level ($\alpha$) during training.}  \textsc{Ada-Eslm} adjusts $\alpha$ dynamically using a CVaR-based controller to stabilize training.
    The learned $\alpha$ values converge to the $[0.1, 0.2]$ range—previously shown in Section~\ref{sec:ablations} (Figure~\ref{fig:alpha-ablation}) to balance training efficiency and data utilization.
    \looseness -1
}
\setlength{\belowcaptionskip}{-5pt}
\setlength{\abovecaptionskip}{-1pt}
\label{fig:ada-eslm-additional-results}
\end{figure}

Concretely, at each evaluation step $k$ (defined by the interval $T_{\text{eval}}$), \textsc{Ada-Eslm} measures the change in CVaR average tail token-level risk scores, which is a proxy for difficulty. Let $\operatorname{CVaR}_{\alpha_k}$ denote the CVaR value at iteration $k$ computed via (\ref{eq:CVaR-definition}). We define the normalized CVaR change, $\Delta_{\text{norm}}$, via:
\[
\Delta_{\text {norm }}\left(\alpha_k\right):=\frac{\mathrm{CVaR}_{\alpha_k}-\mathrm{CVaR}_{\alpha_{k-1}}}{\mathrm{CVaR}_{\alpha_{k-1}}+\varepsilon},
\]
which is a dimension and scale-independent feedback signal, and $\varepsilon > 0$ is a small constant for numerical stability.
The controller updates the confidence level $\alpha$ multiplicatively using:
\[
\alpha_{k+1}=\alpha_k \cdot \exp \left(-\gamma \cdot \Delta_{\text {norm }}\left(\alpha_k\right)\right),
\]
where $\gamma > 0$ controls the update rate. The update rule captures the key intuition:
\begin{itemize}
    \item If $\Delta_{\text{norm}} > 0$ (CVaR increases), then $\alpha$ is decreased to expand the input selection.
    \item If $\Delta_{\text{norm}} < 0$ (CVaR decreases), then $\alpha$ is increased to narrow focus to high-risk inputs.
\end{itemize}
This dynamic adjustment results in a form of \textit{token-level curriculum learning} in which the model begins with broad exposure and progressively narrows focus to the most informative regions of the data. As we further show in Figure~\ref{fig:ada-eslm-additional-results}, \textsc{Ada-Eslm} gradually increases $\alpha$ over training and converges to a stable operating regime in the range $[0.1, 0.2]$—a region empirically shown to yield a strong trade-off between training efficiency and data utility (Section~\ref{sec:ablations}, Figure~\ref{fig:alpha-ablation}).

\begin{algorithm}[ht]
\caption{\textsc{Ada-Eslm}}
\label{alg:adaptive-eslm}
\begin{algorithmic}[1]
\STATE {\bfseries Input:} Language model $\theta$, dataset $\mathcal{D}$, learning rate $\eta$, initial confidence level $\alpha_0 \in (0,1)$, proxy $L$ layers, sensitivity $\gamma > 0$, evaluation interval $T_{\text{eval}}$, batch size $M$, small constant $\varepsilon > 0$.
\STATE Initialize: $\mathrm{CVaR}_{0} \gets 0$.
\STATE Initialize the list: $\texttt{CVaR\_history} \gets []$.
\STATE Append $\mathrm{CVaR}_{0}$ to $\texttt{CVaR\_history}$.
\FOR{each training iteration $k = 1, \dots, K$}
\STATE Sample a batch of instances $\mathcal{B} = \{x_1, \dots, x_M\} \sim \mathcal{D}$ of length $T$ $\{x_j^t\}_{t=1}^{T}, \forall j\in\{1,...,M\}$.
\STATE Compute per-instance scores $S_{\theta_k^L}(x)$ using early-exit $\theta_k^{L}$ proxy. \hfill {\color{Periwinkle} */ Entropy or loss}
\STATE Compute threshold $S_{\theta_k^L, \alpha}^{\operatorname{VaR}} \gets \operatorname{VaR}_{\alpha} \left( \{ S_{\theta_k^L}(x_j) \}_{j=1}^{M} \right)$ using (\ref{eq:VaR-definition}).
\STATE $\Bar{\mathcal{B}} \gets \{ x_j \in \mathcal{B} \mid S_{\theta_k^L}(x_j) \geq S_{\theta_k^L, \alpha}^{\operatorname{VaR}} \}$. \hfill {\color{Periwinkle} */ High-risk instance selection}
    \STATE Compute per-token statistics $S_{\theta_k}(x^t)$: \hfill {\color{Periwinkle} */ Entropy or loss } \\
    $
    S_{\theta_k}(x_j^t) = 
    \begin{cases} 
      H_{\theta_k}(x_j^t) \text{ as in \ref{entropy-defn}}, & \text{(\myinlinecolorbox{electric-blue!15}{$\operatorname{VaR}$-entropy})} \\
      \ell_{\theta_k}(x_j^t) \text{ as in \ref{loss-defn}}, & \text{(\myinlinecolorbox{salmon!30}{$\operatorname{CVaR}$-loss})} 
    \end{cases}
    $
    \STATE Compute threshold $S_{\theta_k, \alpha}^{\operatorname{VaR}} \gets \operatorname{VaR}_{\alpha} \left( \{\{ S_{\theta_k}(x_j^t) \}_{j \in \Bar{\mathcal{B}}}\}_{t=1}^{T} \right)$ using (\ref{eq:VaR-definition}).
    \STATE $\tilde{\mathcal{B}} \gets \{ x_j^t \in \Bar{\mathcal{B}} \mid S_{\theta_k}(x_j^t) \geq S_{\theta_k, \alpha}^{\operatorname{VaR}} \}$. \hfill {\color{Periwinkle} */ High-risk token selection}
    \STATE Compute loss over selected tokens:\hfill {\color{Periwinkle} */ Shaped loss}
    $ 
    \mathcal{L}_{\tilde{\mathcal{B}}}(x;\theta_k) = 
    \begin{cases} 
      \mathbb{E}[\ell_{\theta_k}(x_j^t) \mid x_j^t \in \tilde{\mathcal{B}}], & \text{(\myinlinecolorbox{electric-blue!15}{$\operatorname{VaR}$-entropy}}) \\
      \operatorname{CVaR}_{\alpha}(\ell_{\theta_k}(x)) = \mathbb{E}[\ell_{\theta_k}(x_j^t) \mid x_j^t \in \tilde{\mathcal{B}}] \text{ using } (\ref{eq:CVaR-definition}), & \text{(\myinlinecolorbox{salmon!30}{$\operatorname{CVaR}$-loss}}) \\
    \end{cases}
    $
    \STATE Update model parameters using optimizer $O$: $\theta_{k+1} \leftarrow O(\theta_{k}, \nabla_{\theta} \mathcal{L}_{\tilde{\mathcal{B}}}(x;\theta_k), \eta)$.
    \IF[\hfill {\color{Periwinkle} */ Update $\alpha$}]{$k \bmod T_{\text{eval}} = 0$} 
        \STATE Compute $\mathrm{CVaR}_{\alpha_k} \gets \operatorname{CVaR}_{\alpha_k}(S_{\theta_k})$ \text{ using } (\ref{eq:CVaR-definition}).
            \STATE Retrieve $\operatorname{CVaR}_{\alpha_{k-1}} \gets \texttt{CVaR\_history}[-1]$.
            \STATE Compute normalized CVaR change:
            \[
            \Delta_{\text{norm}}(\alpha_k) \gets \frac{\operatorname{CVaR}_{\alpha_k} - \operatorname{CVaR}_{\alpha_{k-1}}}{|\operatorname{CVaR}_{\alpha_{k-1}}| + \varepsilon}
            \]
            \STATE Update confidence level: 
            $\alpha_{k+1} \gets \alpha_k \cdot \exp(-\gamma \cdot \Delta_{\text{norm}}(\alpha_k)).$
        \STATE Append $\operatorname{CVaR}_{\alpha_k}$ to \texttt{CVaR\_history}.
    \ENDIF
\ENDFOR
\RETURN $\theta_K$
\end{algorithmic}
\end{algorithm}

 \begin{figure}[ht]
    \centering  
    \begin{subfigure}[t]{0.45\linewidth}\centering{\includegraphics[width=1\linewidth,trim=0 0 0 0,clip]{figures/ada_eslm_flops_to_target_new.pdf}}
    \caption{Training FLOPs required ($\downarrow$).}
    \label{fig:ada-eslm-perp-result-app}
    \end{subfigure}%
    \begin{subfigure}[t]{0.45\linewidth}\centering{\includegraphics[width=1\linewidth,trim=0 0 0 0,clip]{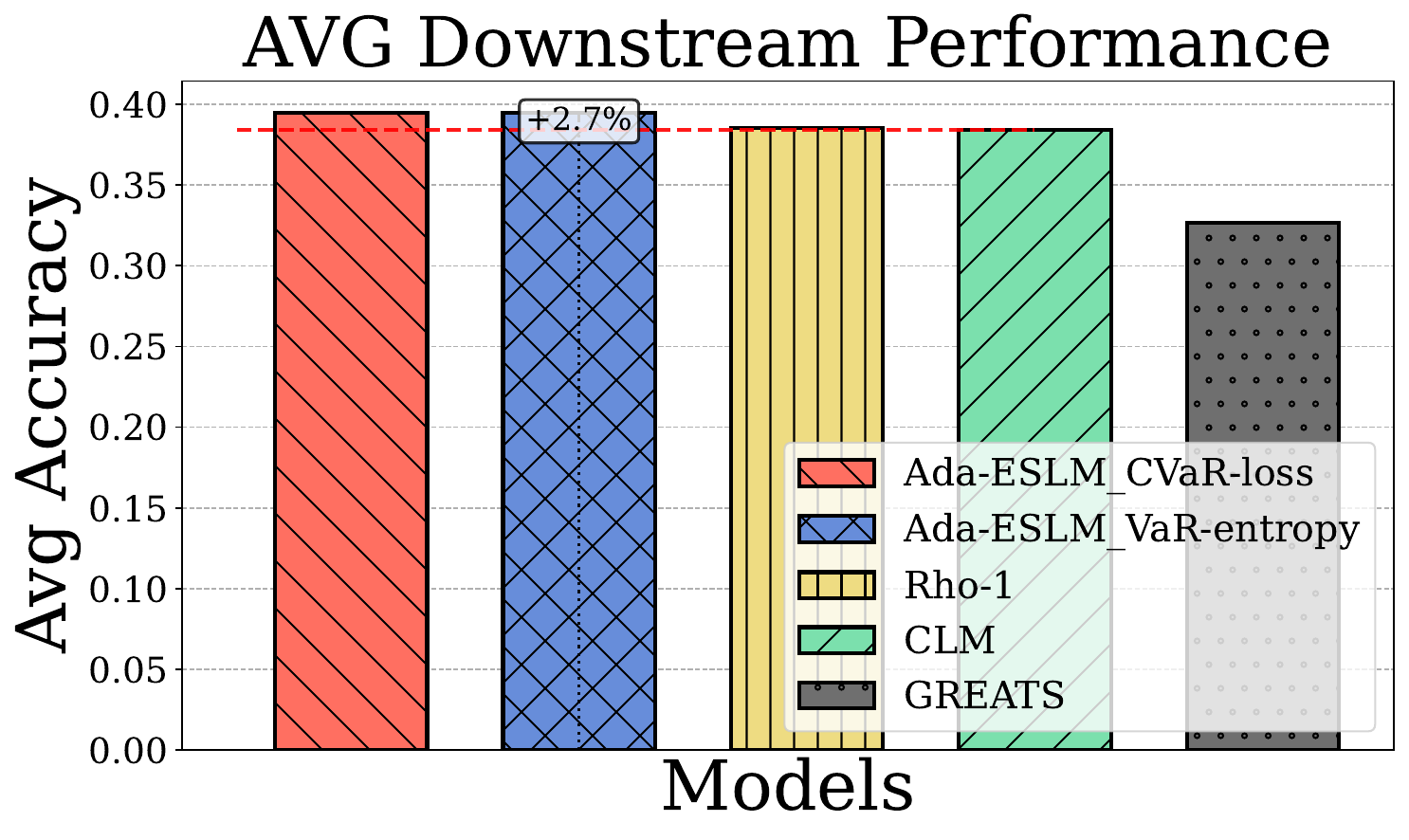}}
    \caption{Generalization accuracy ($\uparrow$).}
    \label{fig:ada-eslm-general-result-app}
    \end{subfigure}%
    \caption{\textbf{\textsc{Ada-ESLM} efficiency and generalization performance.} \textbf{(a):} \textsc{Ada-Eslm} adaptively tunes the $\alpha$ level based on training dynamics, achieving the target validation (log) perplexity with fewer training FLOPs compared to baselines. \textbf{(b):} \textsc{Ada-Eslm} further improves generalization on downstream benchmarks reported in detail in Table~\ref{tab:downstream-perf-ada-eslm-124M}, achieving higher average accuracy than baselines, trained under the same compute budget.
    }
\label{fig:ada-eslm-flops-and-generalization}
\end{figure}

In Table~\ref{tab:downstream-perf-ada-eslm-124M}, we present the downstream performance of \textsc{Ada-Eslm} across standard benchmarks. Notably, the adaptive variant \textsc{\textbf{Ada}-Eslm}-\myinlinecolorbox{electric-blue!15}{$\operatorname{VaR}$-entropy} consistently outperforms both baseline methods (CLM, Rho-1, GREATS) and fixed-$\alpha$ \textsc{Eslm} variants. These results highlight the benefit of dynamically adjusting the token selection threshold during training, demonstrating that \textsc{Ada-Eslm} improves generalization while maintaining high training efficiency.

\begin{table}[ht]
\caption{\textbf{Generalization performance of \textsc{\textbf{Ada}-ESLM} on downstream tasks.} All models (124M) are pretrained under a $\sim$3E17 FLOPs budget on SlimPajama-6B-Unif mixture. We report the best observed accuracy$_{( \text{standard error})}$ or exact match if provided, during training.
\myinlinecolorbox{tableyellow}{\textbf{Highlighted}} values indicate the best performance.
Results demonstrate that the dynamic CVaR-driven adjustment of $\alpha$ level leads to improved generalization over baselines, particularly with \textsc{\textbf{Ada}-Eslm}-\myinlinecolorbox{electric-blue!15}{$\operatorname{VaR}$-entropy} setting.
\looseness -1
}
\label{tab:downstream-perf-ada-eslm-124M} 
\centering
    \resizebox{1\linewidth}{!}{%
        \begin{tabular}{l|c|ccccccc} \specialrule{1.5pt}{1pt}{1pt}
         \multicolumn{1}{c}{\makecell{\\  \\ \textbf{Benchmark}}} & \multicolumn{7}{c}{\textbf{Method (124M)}} \\ \cmidrule(lr){2-9}
        & \# Shots &  \multicolumn{1}{c}{\textsc{\textbf{Ada}-Eslm}-\text{\myinlinecolorbox{salmon!30}{$\operatorname{CVaR}$-loss}}} & \multicolumn{0}{c}{\textsc{\textbf{Ada}-Eslm}-\myinlinecolorbox{electric-blue!15}{$\operatorname{VaR}$-entropy}} &  \multicolumn{1}{c}{\textsc{Eslm}-\text{\myinlinecolorbox{salmon!30}{$\operatorname{CVaR}$-loss}}} & \multicolumn{0}{c}{\textsc{Eslm}-\myinlinecolorbox{electric-blue!15}{$\operatorname{VaR}$-entropy}} & \multicolumn{0}{c}{CLM} & \multicolumn{0}{c}{Rho-1} & \multicolumn{0}{c}{GREATS} \\ 
        \specialrule{1.5pt}{1pt}{1pt}
        ARC-E  \citep{clark2018think} & 0-shot & $0.3728_{(0.0099)}$ &  \tabye $0.3884_{(0.01)}$ & $0.3712_{(0.0099)}$  &    $0.3766_{(0.0099)}$  & $0.3644_{(0.0099)}$ &   $0.3657_{(0.0099)}$  &   $0.3236_{(0.0096)}$ \\
        LAMBADA \citep{paperno-EtAl:2016:P16-1} & 5-shot &  $0.1631_{(0.0051)}$ & $0.1628_{(0.0051)}$ & $0.1595_{(0.0051)}$  & \tabye  $0.1721_{(0.0053)}$   &   $0.1701_{(0.0051)}$   &  $0.1680_{(0.005)}$ & $0.0254_{(0.002)}$ \\
         SciQ \citep{welbl2017crowdsourcing} & 5-shot 
 & $0.6960_{(0.0146)}$ &  $0.7060_{(0.0144)}$ & \tabye $0.7140_{(0.0143)}$  &  $0.7100_{(0.0144)}$  &  $0.6970_{(0.0145)}$ &  $0.7000_{(0.0145)}$ & $0.4350_{(0.0157)}$  \\
        HellaSwag \citep{zellers2019hellaswag} &  5-shot & \tabye $0.2976_{(0.0046)}$ &  $0.2952_{(0.0046)}$ & $0.2930_{(0.0045)}$  & $0.2964_{(0.0046)}$  &  $0.2901_{(0.0045)}$ &  $0.2893_{(0.0045)}$ &  $0.2621_{(0.0044)}$ \\
        TriviaQA \citep{2017arXivtriviaqa} & 1-shot & $0.0082_{(0.0007)}$ &  \tabye $0.0128_{(0.0008)}$ & $0.0128_{(0.0001)}$ &  $0.0066_{(0.0006)}$ & $0.0078_{(0.0007)}$ &   $0.0090_{(0.0007)}$ & $0.0007_{(0.0002)}$ \\
        COPA \citep{NEURIPS2019_4496bf24} & 5-shot & \tabye $0.6600_{(0.0476)}$ & $0.6300_{(0.0485)}$ &  $0.6400_{(0.0482)}$ &  $0.6200_{(0.0488)}$ &  $0.6200_{(0.0488)}$ & $0.6200_{(0.0488)}$ & $0.6400_{(0.0482)}$\\
        MultiRC \citep{NEURIPS2019_4496bf24} & 5-shot &  $0.5600_{(0.0071)}$ & \tabye $0.5668_{(0.0071)}$&  $0.5633_{(0.0071)}$ &  $0.5429_{(0.0071)}$ & $0.5338_{(0.0072)}$ &   $0.5338_{(0.0072)}$ &    $0.5497_{(0.0071)}$ \\
        OpenBookQA \citep{mihaylov2018can} & 5-shot & $0.174_{(0.017)}$ & \tabye $0.178_{(0.0171)}$ &$0.164_{(0.0166)}$  &  $0.170_{(0.0168)}$  &  $0.166_{(0.0167)}$  &  $0.164_{(0.0166)}$ & $0.148_{(0.0159)}$ \\
        PiQA \citep{bisk2020piqa} & 5-shot & $0.6191_{(0.0113)}$ &  $0.6115_{(0.0114)}$ & \tabye $0.6240_{(0.0113)}$ &  $0.6191_{(0.0113)}$  &  $0.6099_{(0.0114)}$ &   $0.6180_{(0.0113)}$ & $0.5571_{(0.0116)}$ \\
        \specialrule{1.5pt}{1pt}{1pt}
        \multicolumn{1}{c}{\textbf{Average} ($\uparrow$)} &  & \textbf{0.39453} &    \textbf{0.39461}  & 0.39353 & 	 0.39041	& 0.38434	& 0.38531 & 0.32684 \\
        \specialrule{1.5pt}{1pt}{1pt}
    \end{tabular}
    }
\end{table}

\clearpage
\section{Risk-Aware Knowledge Distillation with \textsc{ESLM-KD}}
\label{app:eslm-kd}

We provide the implementation for our knowledge distillation setup, namely \textsc{Eslm-Kd}, in Algorithm~\ref{alg:eslm-kd}. 
The student model $\theta$ computes per-token risk scores over each batch, and high-risk tokens are selected via $\operatorname{VaR}_\alpha$ thresholding. 
The student is then supervised only on these informative tokens using a combined loss: a weighted sum of KL divergence to the teacher ($\phi$) and standard cross-entropy. 

In our experiments on knowledge distillation (\cref{sec:results}), we used a 124M GPT-2 model pretrained with the CLM objective (checkpoint 40,000) as the teacher to train a 774M student models on the SlimPajama-6B-Unif dataset. 
Based on hyperparameter tuning, we set the distillation weight to $\lambda = 0.5$ and the teacher temperature to $\rho = 1.0$.
We compare \textsc{Eslm-Kd} against three 774M baselines with the same teacher model: standard CLM training, Dense-KD (dense knowledge distillation without token selection), and SALT \citep{rawat2024little}, a two-staged distillation method which employs distillation in the first stage and then transitions to standard pretraining. For the SALT baseline, we set the distillation iterations to 12,000. We trained all distillation-based models with a compute budget of 1E18 FLOPs.

The experimental results in Figure~\ref{fig:eslm-kd-result} (Section~\ref{sec:results}) show that \textsc{Eslm-Kd} achieves the target validation loss with significantly less training FLOPs, demonstrating its efficiency and effectiveness in large-scale distillation.
Table~\ref{tab:downstream-perf-124m-unif-distillation} further compares the generalization performance of \textsc{Eslm-Kd} against dense distillation using the same teacher model.
The results show that integrating risk-aware token selection into distillation not only reduces compute cost but also improves downstream accuracy over full-token distillation.
\looseness -1
\begin{algorithm}[ht]
\caption{\textsc{Eslm-Kd}}
\label{alg:eslm-kd}
\begin{algorithmic}[1]
\STATE {\bfseries Input:} Teacher LM parameters $\phi$, student LM parameters $\theta$, dataset $\mathcal{D}$, learning rate $\eta$, confidence level $\alpha \in (0,1)$, batch size $M$, teacher temperature $\rho > 0$, distillation loss weight $\lambda \in [0,1]$.
\FOR{each training iteration $k = 1, \dots, K$}
    \STATE Sample a batch of tokens $\mathcal{B} = \{x_1, \dots, x_M\} \sim \mathcal{D}$.
    \STATE Compute per-token statistics $S_{\theta}(x_j)$ using the student model:\hfill {\color{Periwinkle} */ Entropy or loss}
    \[
    S_{\theta}(x_j) = 
    \begin{cases} 
        H_{\theta}(x_j)\text{ as in \ref{entropy-defn}}, & \text{(\myinlinecolorbox{electric-blue!15}{$\operatorname{VaR}$-entropy})} \\
        \ell_{\theta}(x_j)\text{ as in \ref{loss-defn}}, & \text{(\myinlinecolorbox{salmon!30}{$\operatorname{CVaR}$-loss})} 
    \end{cases}
    \]
    \STATE Compute VaR threshold: $S_{\theta, \alpha}^{\operatorname{VaR}} \gets \operatorname{VaR}_{\alpha} \left( \{ S_{\theta}(x_j) \}_{j=1}^{M} \right)$  using (\ref{eq:VaR-definition}).
    \STATE Select high-risk tokens: $\tilde{\mathcal{B}} \gets \{ x_j \in \mathcal{B} \mid S_{\theta}(x_j) \geq S_{\theta, \alpha}^{\operatorname{VaR}} \}$.
    \STATE Compute combined student loss on selected tokens:\hfill {\color{Periwinkle} */ Distillation + cross-entropy loss}
    \[
    \mathcal{L}_{\textsc{Eslm-Kd}} = \frac{1}{|\tilde{\mathcal{B}}|} \sum_{x_j \in \tilde{\mathcal{B}}} \left[ \lambda \cdot \mathrm{KL}\left(P^{\phi}_\rho(x_j \mid x_{<j}) \,\|\, P^{\theta}_\rho(x_j \mid x_{<j})\right) + (1 - \lambda) \cdot \ell_{\theta_k}(x_j) \right]
    \]
    \STATE Update student parameters using optimizer $O$: $\theta_{k+1} \leftarrow O(\theta_k, \nabla_\theta \mathcal{L}_{\textsc{Eslm-Kd}}, \eta)$.
\ENDFOR
\RETURN $\theta_K$.
\end{algorithmic}
\end{algorithm}

\begin{table}[t]
\caption{\textbf{Generalization performance of \textsc{ESLM-KD} on downstream tasks.} All models (774M) are pretrained under a $\sim$1E18 FLOPs budget on SlimPajama-6B-Unif mixture, using the same teacher model. We report the best observed accuracy$_{( \text{standard error})}$ or exact match if provided, during training.
\myinlinecolorbox{tableyellow}{\textbf{Highlighted}} values indicate the best performance.
\looseness -1
}
\label{tab:downstream-perf-124m-unif-distillation} 
\centering
    \resizebox{1\linewidth}{!}{%
        \begin{tabular}{l|c|cccccccc} \specialrule{1.5pt}{1pt}{1pt}
         \multicolumn{1}{c}{\makecell{\\  \\ \textbf{Benchmark}}} & \multicolumn{4}{c}{\textbf{Method (774M)}} \\ \cmidrule(lr){2-6}
        & \# Shots & \multicolumn{1}{c}{\textsc{Eslm-Kd}-\myinlinecolorbox{salmon!30}{$\operatorname{CVaR}$-loss}} & \multicolumn{0}{c}{\textsc{Eslm-Kd}-\myinlinecolorbox{electric-blue!15}{$\operatorname{VaR}$-entropy}} 
        & \multicolumn{0}{c}{Dense-KD} & \multicolumn{0}{c}{SALT} \\
        \specialrule{1.5pt}{1pt}{1pt}
        ARC-E  \citep{clark2018think} & 0-shot &   $0.3901_{(0.01)}$  &   $0.3935_{(0.01)}$  & $0.3909_{(0.01)}$ &   \tabye $0.3947_{(0.01)}$   \\
        LAMBADA \citep{paperno-EtAl:2016:P16-1} & 5-shot & $0.2472_{(0.006)}$  & \tabye $0.2480_{(0.006)}$   &   $0.2429_{(0.006)}$   &  $0.2258_{(0.0058)}$ \\
         SciQ \citep{welbl2017crowdsourcing} & 5-shot 
& $0.766_{(0.0134)}$  &  \tabye $0.773_{(0.0133)}$   &  $0.759_{(0.0135)}$   & $0.770_{(0.0133)}$ \\
        HellaSwag \citep{zellers2019hellaswag} &  5-shot & $0.3200_{(0.0047)}$  & $0.3189_{(0.0047)}$   &   $0.3179_{(0.0046)}$   & \tabye $0.3313_{(0.0047)}$ \\
        TriviaQA \citep{2017arXivtriviaqa} & 1-shot & $0.0273_{(0.0012)}$  & $0.0280_{(0.0012)}$   &  $0.0231_{(0.0011)}$   &  \tabye $0.0299_{(0.0013)}$ \\
        COPA \citep{NEURIPS2019_4496bf24} & 5-shot & $0.68_{(0.0469)}$  & $0.66_{(0.0476)}$   &  \tabye $0.69_{(0.0465)}$   &  $0.67_{(0.0473)}$ \\
        MultiRC \citep{NEURIPS2019_4496bf24} & 5-shot & $0.5408_{(0.0072)}$  & $0.5406_{(0.0072)}$   &   $0.5360_{(0.0072)}$   & \tabye $0.5420_{(0.0072)}$ \\
        OpenBookQA \citep{mihaylov2018can} & 5-shot & $0.270_{(0.0199)}$  & \tabye $0.290_{(0.0203)}$   &   $0.278_{(0.0201)}$   &  $0.278_{(0.02)}$ \\
        PiQA \citep{bisk2020piqa} & 5-shot & $0.6300_{(0.0113)}$  & $0.6245_{(0.0113)}$   &   \tabye $0.6327_{(0.0112)}$   &  $0.6322_{(0.0113)}$ \\
        \specialrule{1.5pt}{1pt}{1pt}
        \multicolumn{1}{c}{\textbf{Average} ($\uparrow$)} &    &  0.4301 & 	\tabye \textbf{0.4307}	& 0.4299	&  \textbf{0.4304} \\
        \specialrule{1.5pt}{1pt}{1pt}
    \end{tabular}
    }
\vspace{-1.5em}
\end{table}

\section{Experiment Details}
\label{app:experiment-details}

\subsection{Experimental Setup}
\label{app:exp-setup}

We set the training hyperparameters as in Table~\ref{tab:params-and-hyperparams}.
We train GPT-2 models \citep{radford2019language}
of sizes 124M, 350M, 774M, and 1.5B parameters, with architecture details reported in Table~\ref{tab:hyperparams-model-architecture}.

\begin{table}[ht]
\centering
\caption{Training and evaluation hyperparameters used in all experiments.}
\label{tab:params-and-hyperparams}
\resizebox{0.75\textwidth}{!}{
\begin{tabular}{ll}
\specialrule{1.5pt}{1pt}{1pt}
\textbf{Hyperparameter} & \textbf{Value} \\
\specialrule{1.5pt}{1pt}{1pt}

\multicolumn{2}{c}{\textbf{\textsc{ESLM}-Specific}} \\
\midrule
Confidence level ($\alpha$) & Optimized over \{0.05, 0.1, 0.2\}  \\
Early-exit layers for proxy phase ($L$) & Optimized over \{1, 2, 3\} \\
\midrule

\multicolumn{2}{c}{\textbf{General Setup}} \\
\midrule
Mini-batch size ($M$) in tokens & $\{$  12 (124M), 8 (350M/774M), 4 (1.5B) $\}$ $\times$ 1024 \\
Gradient accumulation steps & 40 \\
Effective batch size in tokens  & $\{$ 480 (124M), 320 (350M/774M), 160 (1.5B) $\}$  $\times$ 1024  \\
Sequence length ($T$) & 1024 \\
Vocabulary size ($|\mathcal{V}|$) & 50304 \\
Dropout & 0 \\
Evaluation interval ($T_{\text{eval}}$) & 1000 iterations \\
Evaluation steps & 200 iterations \\
\midrule

\multicolumn{2}{c}{\textbf{Optimization}} \\
\midrule
Optimizer ($O$) & AdamW with $\beta_1=0.9$, $\beta_2=0.95$ \\
Learning rate schedule & Cosine annealing with warmup \\
Max. learning rate ($\eta$) & 0.0006 (124M/350M), 0.0001 (774M/1.5B) \\
Min. learning rate & 0.00006 \\
Warmup steps & 2000 \\
Decay iterations & 200{,}000 \\
Weight decay & 0.1 \\
Gradient clipping & 1.0 \\
\midrule

\multicolumn{2}{c}{\textbf{Knowledge Distillation}} \\
\midrule
Teacher temperature ($\rho$) & 1.0 \\
Distillation loss weight ($\lambda$) & 0.5 \\

\specialrule{1.5pt}{1pt}{1pt}
\end{tabular}
}
\end{table}
\begin{table}[ht]
\caption{Architecture hyperparameters for GPT-2 model sizes.}
\centering
\resizebox{0.55\textwidth}{!}{
\begin{tabular}{cccc}
\specialrule{1.5pt}{1pt}{1pt} 
\textbf{Model size} & \textbf{Layers} & \textbf{Attention heads} & \textbf{Embed dimension}  \\
\specialrule{1.5pt}{1pt}{1pt} 
124M & 12 &  12 & 768  \\
350M & 24 & 16& 1024 \\
774M & 36 & 20 & 1280\\
1.5B & 48 & 25 & 1600\\
\specialrule{1.5pt}{1pt}{1pt}
\end{tabular}
}
\label{tab:hyperparams-model-architecture}
\end{table}
\subsection{Pretraining Corpus}
\label{app:corpus}
We utilize the SlimPajama-6B dataset for our experiments.
SlimPajama-6B \citep{cerebras2023slimpajama} mixture consists of seven data domains: \{Arxiv, Book, CommonCrawl, C4, Github, Stackexchange, Wikipedia\} with two weighted versions: uniform domain weights (SlimPajama-6B-Unif) and DoReMi \citep{xie2023doremi} domain weights (SlimPajama-6B-DoReMi).

In Table~\ref{tab:data-mixture-weights}, we report the domain weights for the experiments under SlimPajama-6B \citep{cerebras2023slimpajama} mixture, using DoReMi \citep{xie2023doremi} and uniform weights.

\begin{table*}[ht]
\caption{Domain weights used for experiments on the  SlimPajama-6B mixture.
}
\label{tab:data-mixture-weights} 
\centering
    \resizebox{0.35\linewidth}{!}{%
        \begin{tabular}{lll} 
        \specialrule{1.5pt}{1pt}{1pt}
        \textbf{Domain} & \textbf{DoReMi} & \textbf{Unif} \\
        \specialrule{1.5pt}{1pt}{1pt}
        Arxiv & 0.04235 & 0.1428 \\
        Book & 0.08201 & 0.1428 \\
        CC & 0.381 & 0.1428 \\
        C4 & 0.1141 & 0.1428 \\
        Github & 0.0654 & 0.1428 \\ 
        Stackexchange & 0.0847 & 0.1428 \\
        Wikipedia & 0.2305 & 0.1428 \\
        \specialrule{1.5pt}{1pt}{1pt}
    \end{tabular}
    }
\end{table*}
\subsection{Hardware \& Computational Overhead}
\label{app:hardware}
All experiments were conducted on the HTCondor-managed cluster equipped with NVIDIA A100 GPUs (80GB). Model pretraining and evaluation were parallelized using PyTorch's Distributed Data Parallel (DDP) framework \citep{pytorch} with the NCCL backend and mixed-precision (bfloat16) training. 
We used $4\times$A100 GPUs for experiments on the SlimPajama-6B mixtures.

\paragraph{Runtime analysis.}
\begin{wraptable}{R}{0.5\textwidth}
\centering
\caption{Runtime comparison of 124M models trained on SlimPajama-6B-Unif, for convergence to a target validation loss value of $2.60$. The overhead compared to the standard training is mainly due to the mismatch between sparsity introduced via instance and token selection and current hardware optimizations.}
\label{tab:runtime-comp}
\resizebox{1\linewidth}{!}{%
\begin{tabular}{lll}
\specialrule{1.5pt}{1pt}{1pt}
\textbf{Method} & \textbf{Wall-clock time (hrs)} & \textbf{GPU utilization (avg)} \\
\specialrule{1.5pt}{1pt}{1pt}
\textsc{Eslm}-\myinlinecolorbox{electric-blue!15}{$\operatorname{VaR}$-entropy} & 11.73 & 91\%\\ 
\textsc{Eslm}-\myinlinecolorbox{salmon!30}{$\operatorname{CVaR}$-loss} & 9.91 & 93\% \\  
CLM & 5.33 & 96\% \\
Rho-1 & 17.91 & 88\%\\
GREATS & 99.89 & 75\%\\
\specialrule{1.5pt}{1pt}{1pt}
\end{tabular}
}
\end{wraptable} 
In Table~\ref{tab:runtime-comp}, we compare the wall-clock time of 124M models trained on the SlimPajama-6B-Unif, for convergence to a target validation loss value of $2.60$. 
While \textsc{Eslm} achieves substantial reductions in training FLOPs, reaches lower validation loss, and stronger downstream performance, it incurs higher wall-clock time compared to standard training. However, it remains significantly more efficient than GREATS and Rho-1 baselines--nearly \textbf{twice as fast as Rho-1} and \textbf{over 8× faster than GREATS}.
We attribute this overhead to mismatches between sparse training operations and current hardware optimizations.
This is corroborated directly by the GPU utilization figures reported in Table~\ref{tab:runtime-comp}: CLM, whose dense, uniform per-token operations map cleanly onto accelerator hardware, sustains the highest average utilization (96\%), while \textsc{Eslm} variants operate modestly below this (91--93\%), consistent with the irregular, fragmented backpropagation paths induced by instance- and token-level selection. Rho-1 (88\%) and, more substantially, GREATS (75\%) show lower utilization still, which we attribute to additional non-fused operations layered on top of the standard training step—Rho-1's external reference-model forward pass, and GREATS' per-sample gradient and ghost inner-product computations. Notably, the utilization ordering tracks the wall-clock overhead ordering closely, reinforcing that reduced hardware utilization, rather than any increase in the number of operations performed, is the primary driver of \textsc{Eslm}'s wall-clock gap relative to CLM despite its lower FLOPs count.

As also explained by the recent efficiency works \citep{lasby2024srigl}, achieving practical acceleration from FLOPs savings requires optimized hardware support for sparse backpropagation—an engineering direction which is beyond our paper's scope.
Although \textsc{Eslm}'s per-token risk scores are computed during the forward pass via sorting with $O(M \log M)$ complexity per batch—without requiring additional external inference or backpropagation—its proxy pass requires a shallow forward call, and VaR-based token filtering introduces sparsity into the training process. This sparsity, while beneficial for compute efficiency, leads to irregular and fragmented backpropagation paths that underutilize the dense compute capabilities of modern accelerators.
Unlike the uniform operations of standard CLM, \textsc{Eslm}'s selective masking disrupts efficient tensor fusion, resulting in slower wall-clock runtime despite using fewer FLOPs. Nonetheless, \textsc{Eslm}  provides a favorable trade-off: improved efficiency per FLOP and enhanced generalization.\textbf{ We expect future work leveraging sparsity-aware hardware or sparse accelerators to further reduce this overhead \citep{lasby2024srigl} and unlock the full potential of selective training}.

\subsection{Baselines}
\label{app:baselines}

We identify the baseline methods against which we compare our \textsc{Eslm} approach, specifically from online batch selection methods for LLM pretraining and standard training as discussed in Section~\ref{sec:experiments}.
We provide the baseline implementation details below:
\begin{itemize}[left=0.2cm]
    \item For the Rho-1 baseline \citep{lin2024rho}, we used pretrained GPT-2 models trained via CLM objective as the reference model. 
    Since training a high-quality reference model is the main bottleneck of the Rho-1 method, we used the last checkpoints of pretrained models as proxy models. 
    For pretraining on SlimPajama-6B mixtures (Unif and DoReMi), we used the last saved checkpoint of CLM GPT-2 models as the reference models. Specifically, we utilized 40000, 30000, and 30000 checkpoints for 124M, 350M, 774M models, respectively. We set the loss threshold parameter to $0.1$. For Rho-1's total FLOPs calculation, we include additional FLOPs from the forward call on the reference model.

    \item For the GREATS baseline, we follow the original setup by \citet{wang2024greats}, using a small validation set ($0.5\times$ the batch size) and setting the batch selection budget to $0.9$, aligning with \textsc{Eslm}'s $\alpha=0.1$ level. To compute the training FLOPs for GREATS, we include the forward passes on both training and validation inputs, the backward pass through linear layers to obtain per-example gradients with respect to pre-activation outputs, and the additional FLOPs for computing ghost inner products. While GREATS is evaluated only on the GPT-2 124M model in the original paper, we also restrict our comparison to this setting. Despite adopting their ghost inner product optimization, we found the method to be highly memory-intensive when scaling to larger models, and it could not run stably beyond 124M size.
\end{itemize}

\subsection{Evaluation details}
\label{app:eval-details}
We evaluate pretrained models on a suite of standard language understanding benchmarks in the zero-shot and few-shot settings, using the \texttt{lm-evaluation-harness} evaluation suite \citep{eval-harness}, 
including HellaSwag \citep{zellers2019hellaswag}, LAMBADA \citep{paperno-EtAl:2016:P16-1}, ARC-Easy \citep{clark2018think}, TriviaQA \citep{2017arXivtriviaqa}, SciQ \citep{welbl2017crowdsourcing}, COPA \citep{NEURIPS2019_4496bf24}, MultiRC \citep{NEURIPS2019_4496bf24}, OpenBookQA \citep{mihaylov2018can}, and PiQA \citep{bisk2020piqa} tasks. 
We used the default settings provided by \texttt{lm-evaluation-harness}, which means all evaluations are performed on held-out validation splits, or test splits if provided, and standard errors are calculated using bootstrapping.
Accuracy (norm if provided) or exact match is used as the primary metric.

\subsection{Reproducibility}
\label{app:reproducibility}
For reproducibility, we provide a clear algorithmic description of the \textsc{Eslm} method in Algorithm~\ref{alg:eslm}.
Our implementation builds on the open-source \texttt{NanoGPT} codebase \citep{nanogpt}.
To handle training on the SlimPajama-6B dataset mixture, we adapted the open-source code of DoReMi \citep{xie2023doremi} and DoGE \citep{fan2023doge}. 
As we detail in Section~\ref{sec:eslm}, we estimate the training FLOPs based on the theoretical estimate by \citet{kaplan2020scaling, chowdhery2023palm}. 
To skip redundant gradient computation as a result of token-level loss shaping, we use PyTorch’s backward hook mechanism (\texttt{Tensor.register\_hook(hook)}) for gradient masking.
For downstream evaluation, we utilize the publicly available \texttt{lm-evaluation-harness} suite \citep{eval-harness}.

\section{Additional experimental results}
\label{app:additional-experiments}

In this section, we report additional experimental results, showing validation perplexity convergence in compute space (Appendix~\ref{app:val-loss-vs-flops-results}) and generalization performance in downstream benchmark tasks (Appendix~\ref{app:additional-downstream-performance}) on different datasets across model sizes.

\subsection{Validation loss versus training FLOPs/iterations results}
\label{app:val-loss-vs-flops-results}

\begin{figure*}[t]
    \centering  
    \begin{subfigure}[t]{0.25\linewidth}\centering{\includegraphics[width=1\linewidth,trim=0 0 0 0,clip]{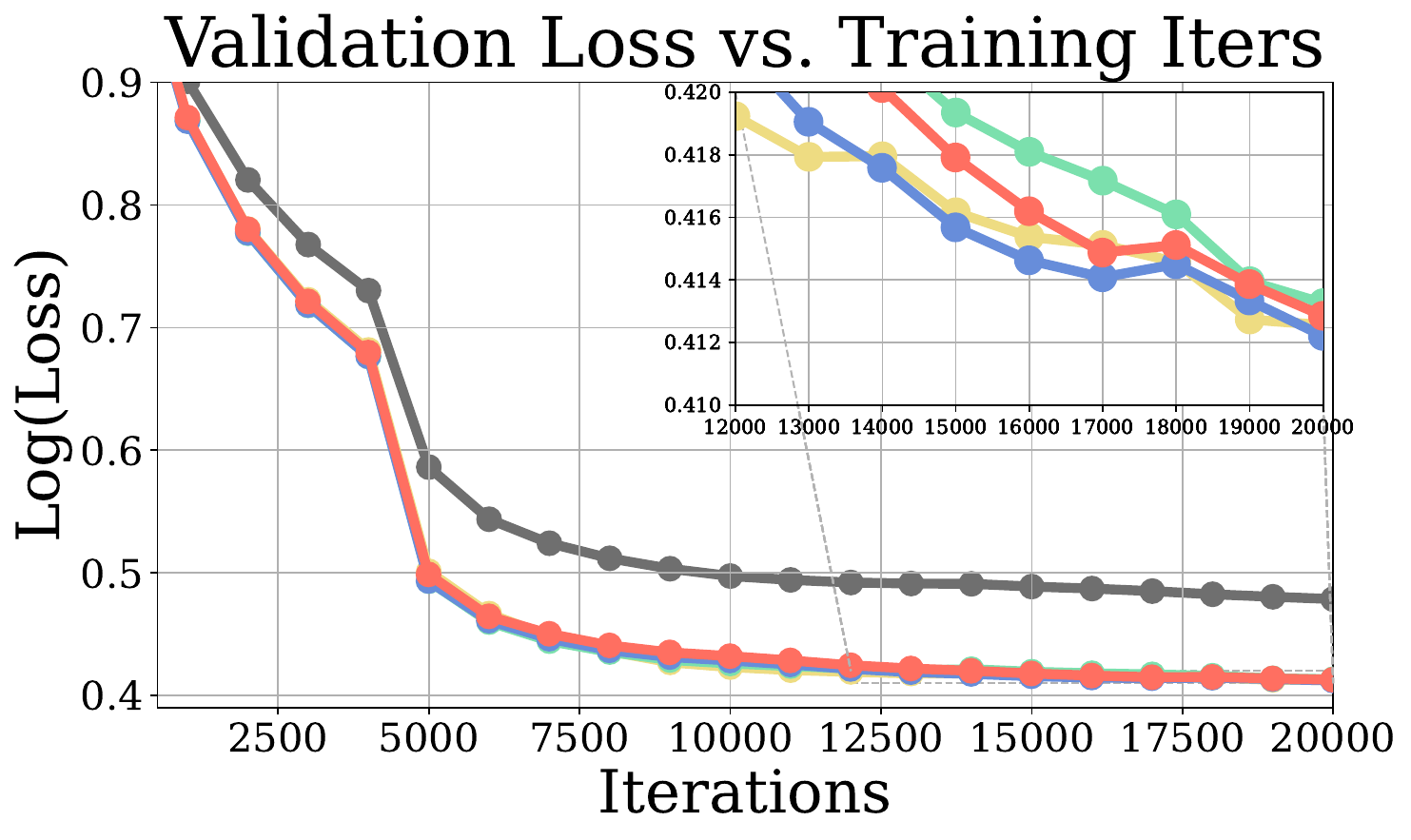}}
    \caption{124M (Iterations).}
    \end{subfigure}%
    \begin{subfigure}[t]{0.25\linewidth}\centering{\includegraphics[width=1\linewidth,trim=0 0 0 0,clip]{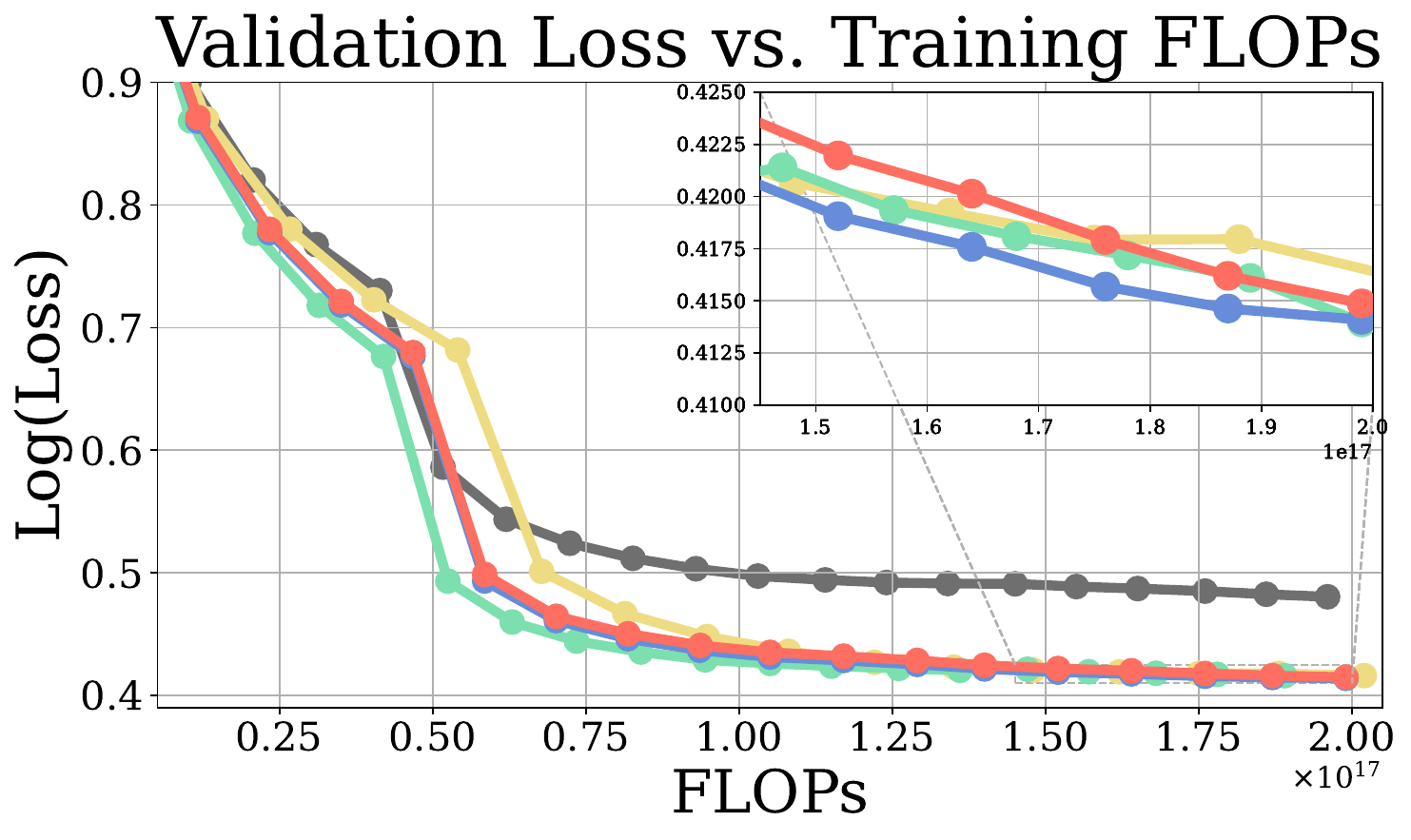}}
    \caption{124M (FLOPs).}
    \end{subfigure}%
    \begin{subfigure}[t]{0.25\linewidth}\centering{\includegraphics[width=1\linewidth,trim=0 0 0 0,clip]{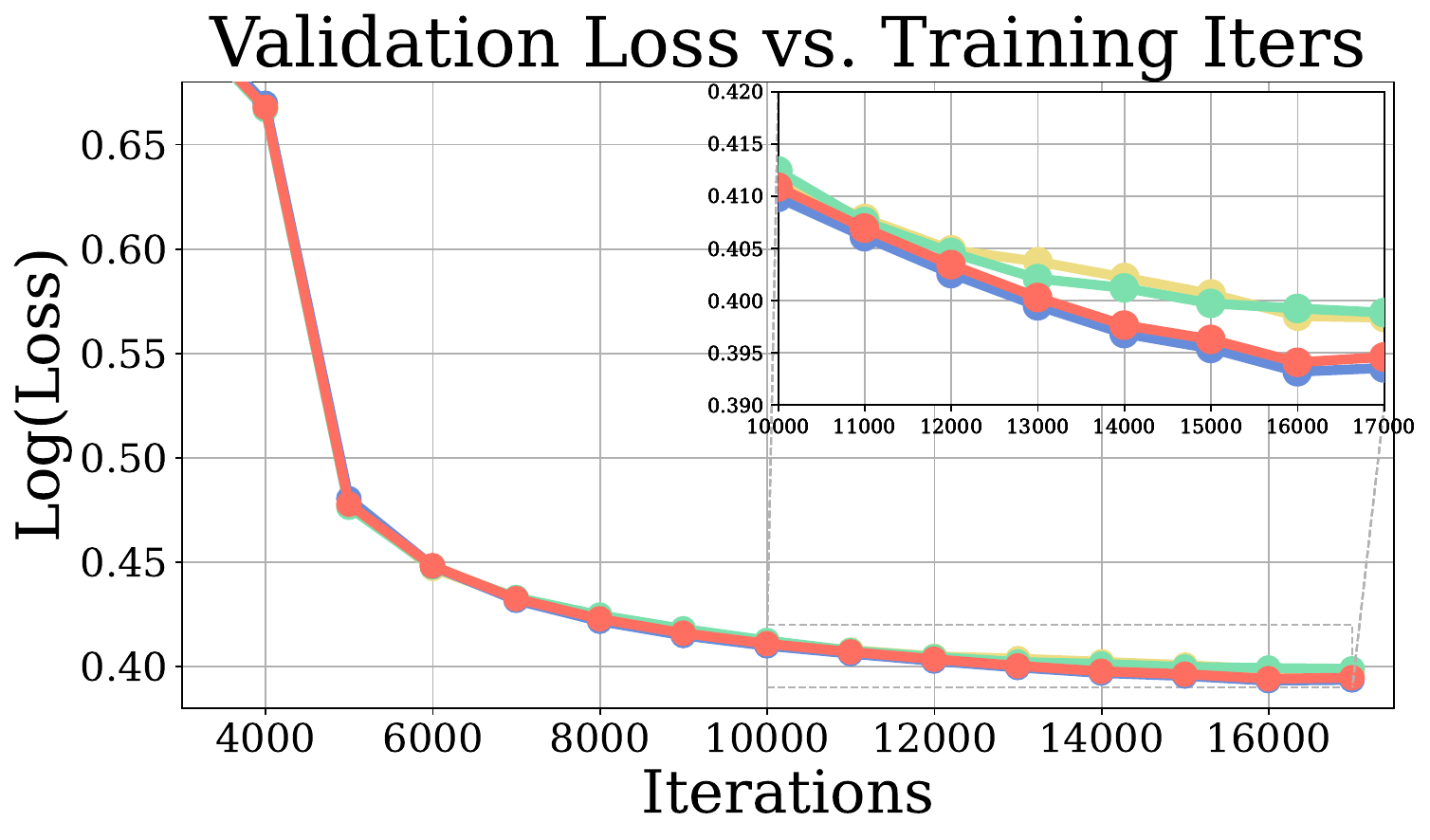}}
    \caption{350M (Iterations).}
    \end{subfigure}%
    \begin{subfigure}[t]{0.25\linewidth}\centering{\includegraphics[width=1\linewidth,trim=0 0 0 0,clip]{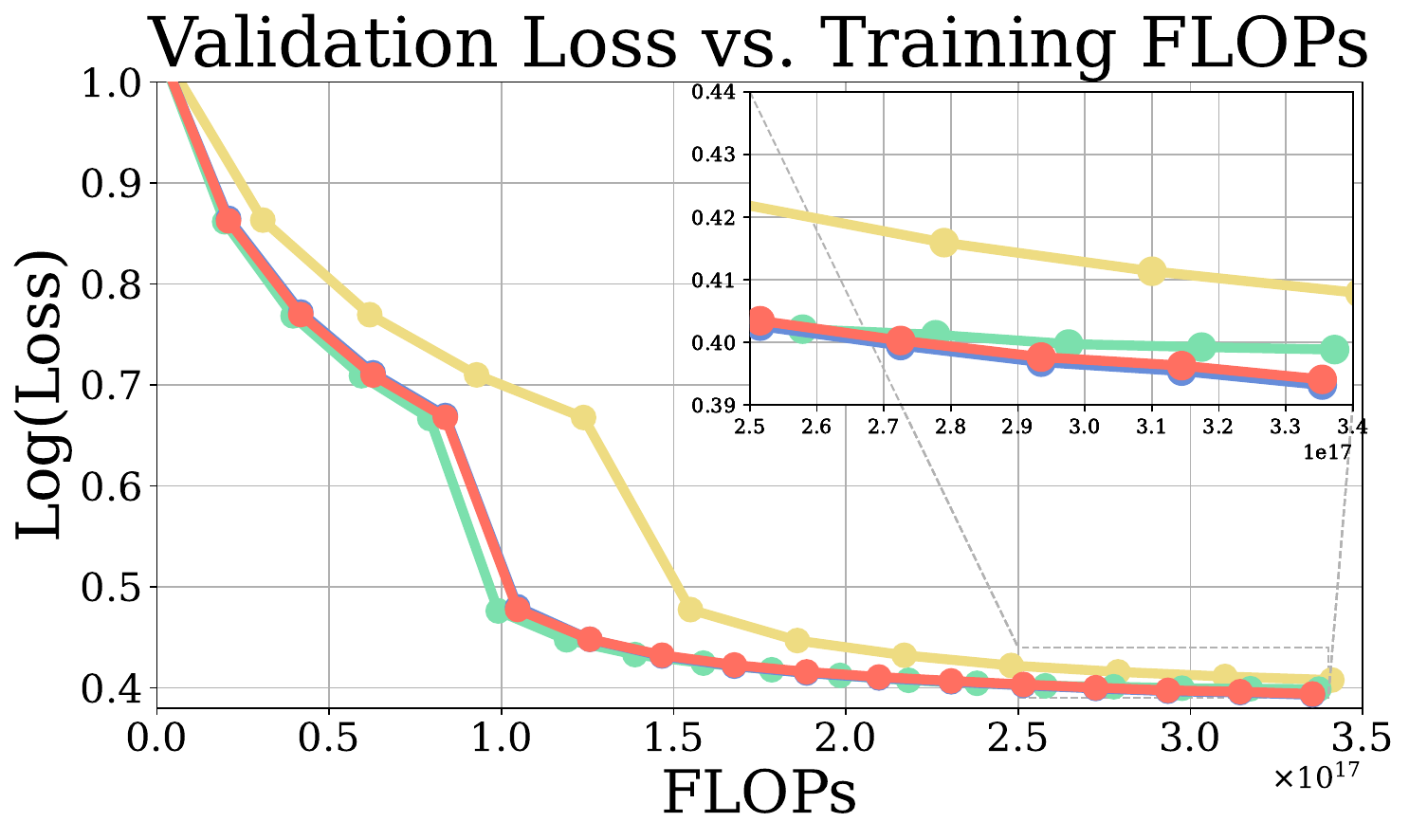}}
    \caption{350M (FLOPs).}
    \end{subfigure}%
    \\
    \begin{subfigure}{0.8\linewidth}\centering{\includegraphics[width=1\linewidth,trim=180 20 160 20,clip]{figures/legend_val_loss_flops.pdf}}
    \end{subfigure}%
    \setlength{\belowcaptionskip}{-5pt}
    \caption{\textbf{Validation loss convergence on SlimPajama-6B-DoReMi.} We report convergence of validation loss versus training FLOPs/iterations of models trained on SlimPajama-6B-DoReMi mixture. \textsc{Eslm} variants provide faster convergence to lower loss values in terms of iterations, while requiring fewer FLOPs than baselines at the convergence. 
    }
\label{fig:val-loss-flops-slimp-doremi}
\end{figure*}

\begin{figure*}[t]
    \centering  
    \begin{subfigure}[t]{0.25\linewidth}\centering{\includegraphics[width=1\linewidth,trim=0 0 0 0,clip]{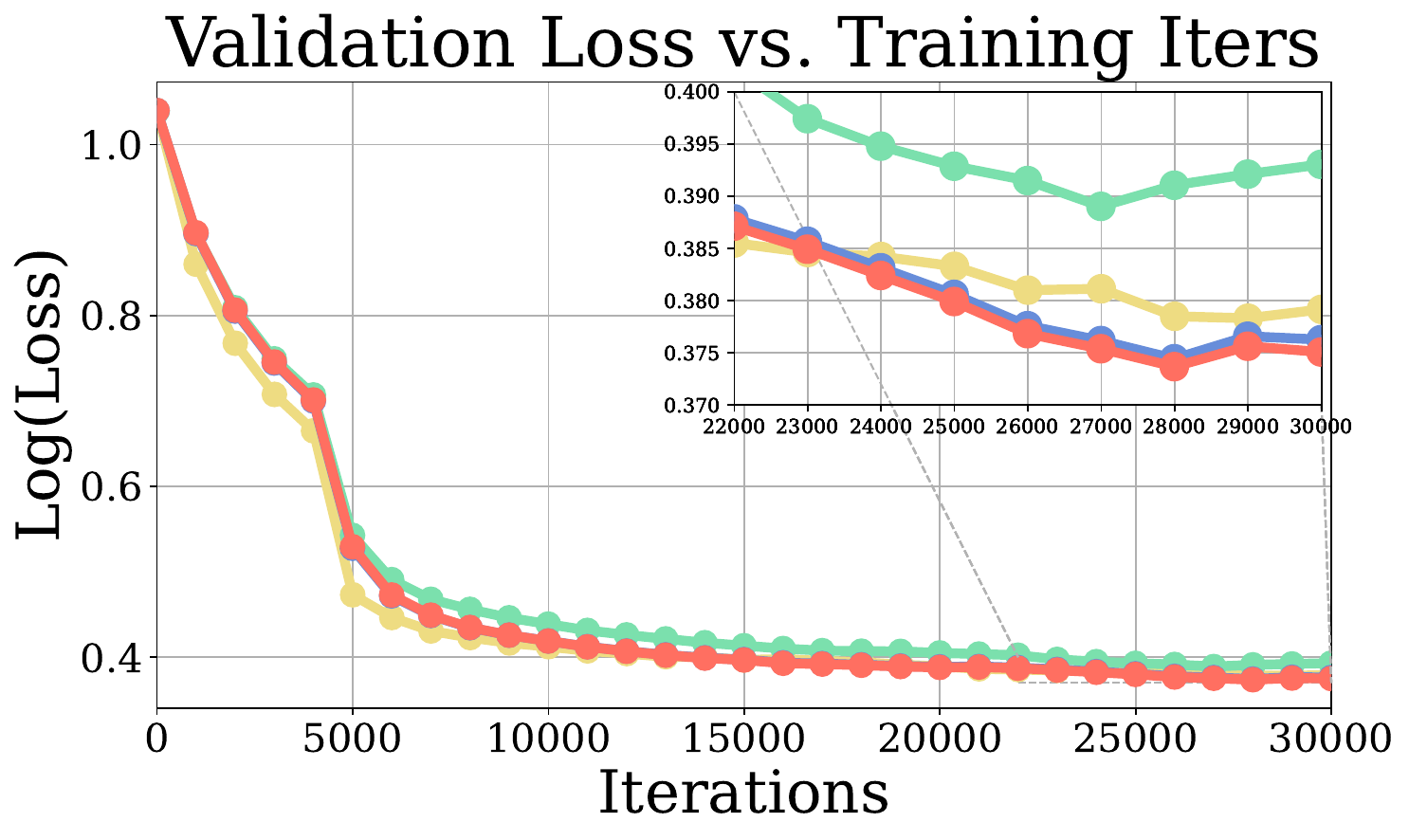}}
    \caption{774M (Iterations).}
    \end{subfigure}%
    \begin{subfigure}[t]{0.25\linewidth}\centering{\includegraphics[width=1\linewidth,trim=0 0 0 0,clip]{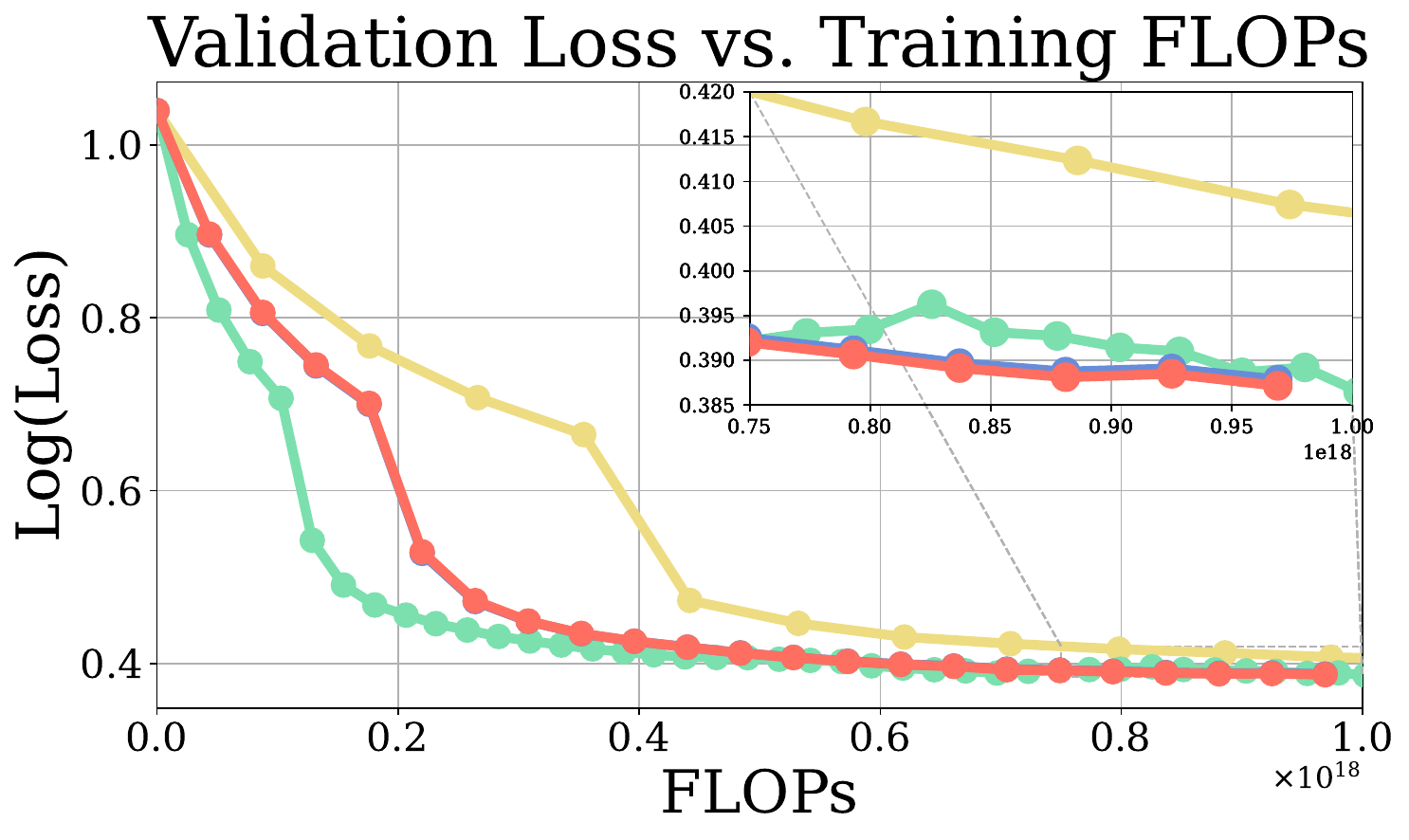}}
    \caption{774M (FLOPs).}
    \end{subfigure}%
    \begin{subfigure}[t]{0.25\linewidth}\centering{\includegraphics[width=1\linewidth,trim=0 0 0 0,clip]{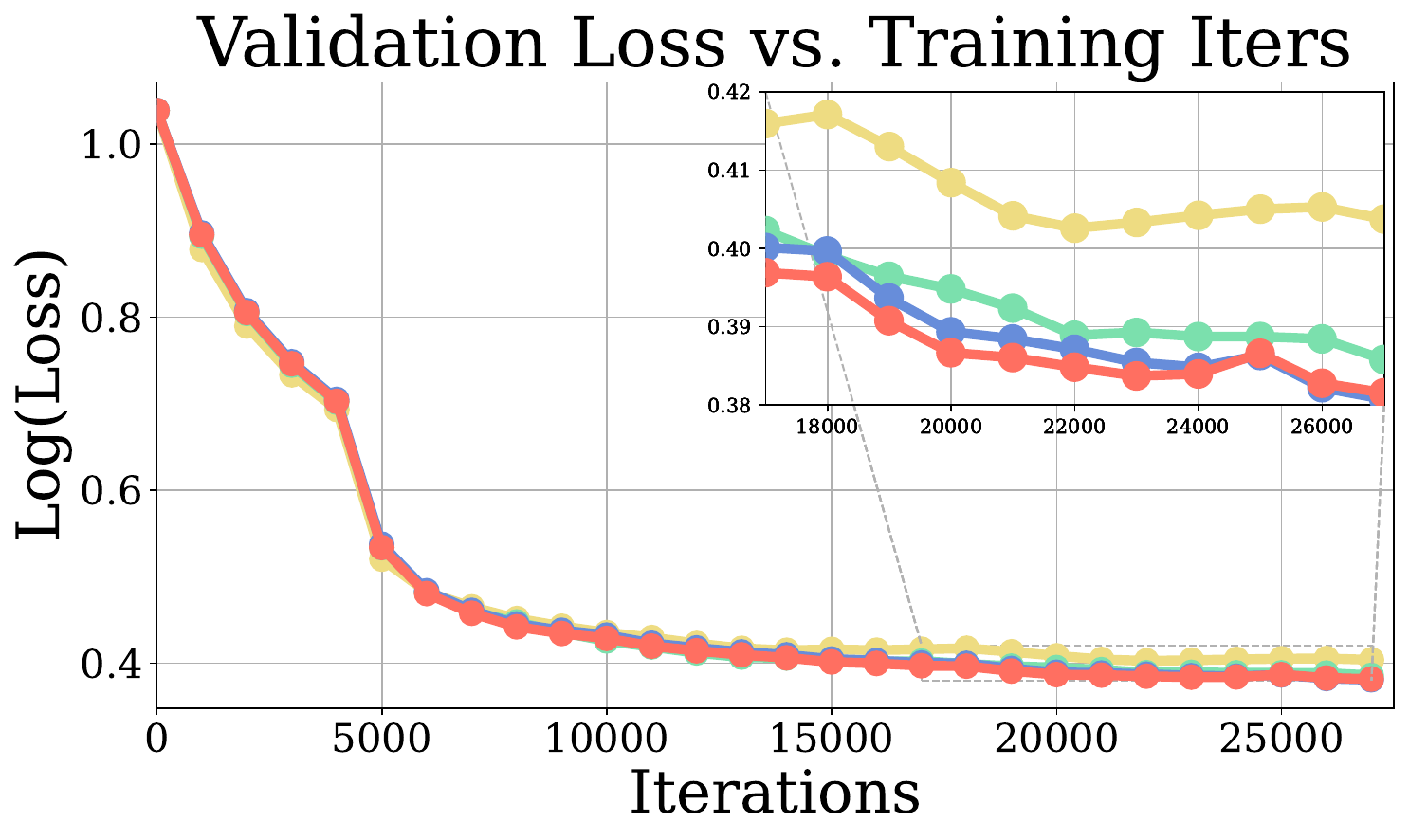}}
    \caption{1.5B (Iterations).}
    \end{subfigure}%
    \begin{subfigure}[t]{0.25\linewidth}\centering{\includegraphics[width=1\linewidth,trim=0 0 0 0,clip]{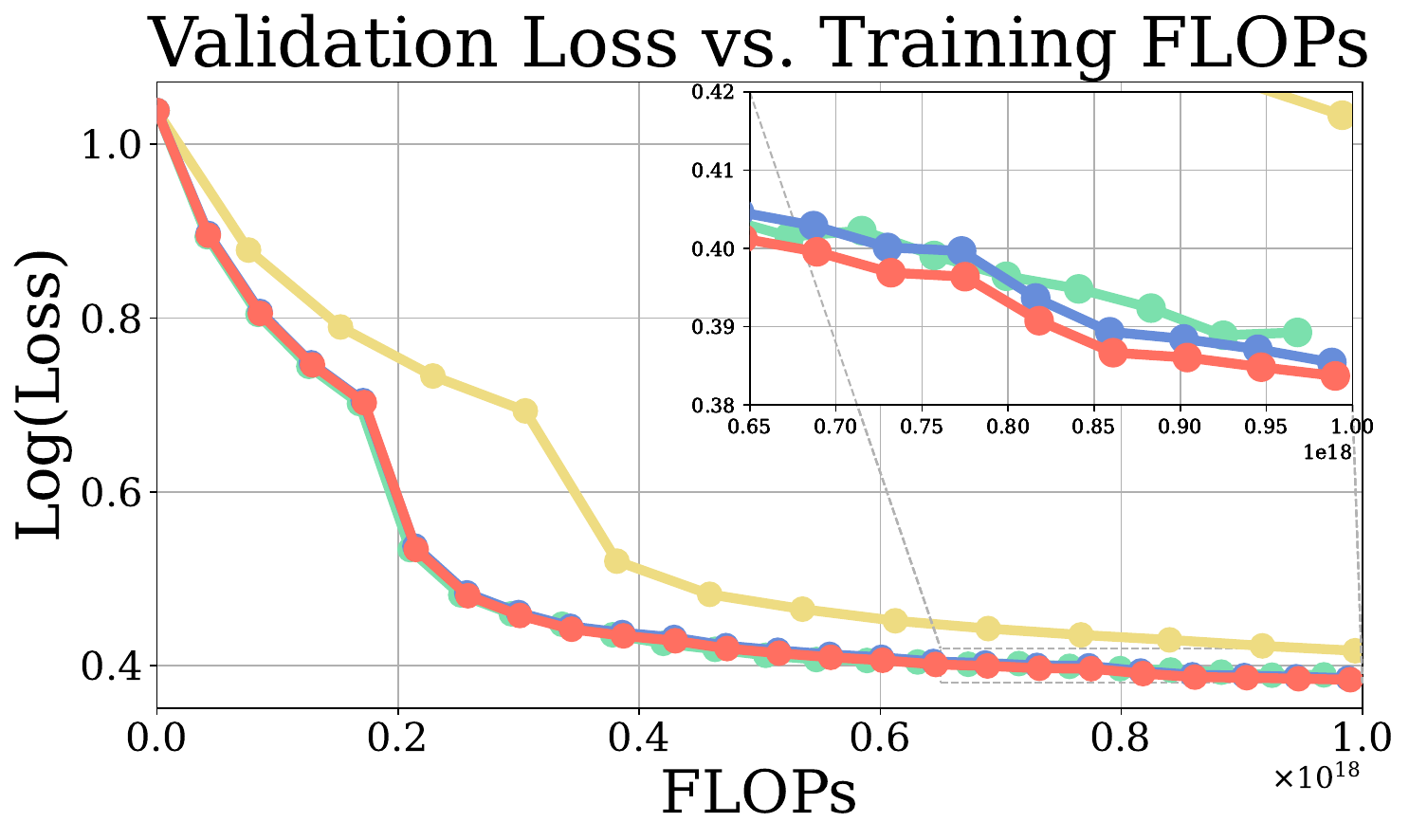}}
    \caption{1.5B (FLOPs).}
    \end{subfigure}%
    \\
    \begin{subfigure}{0.8\linewidth}\centering{\includegraphics[width=1\linewidth,trim=180 20 160 20,clip]{figures/legend_val_loss_flops.pdf}}
    \end{subfigure}%
    \setlength{\belowcaptionskip}{-5pt}
    \caption{\textbf{Validation loss convergence on SlimPajama-6B-Unif.} We report convergence of validation loss versus training FLOPs/iterations of models trained on SlimPajama-6B-Unif mixture, for 774M and 1.5B models. \textsc{Eslm} variants provide faster convergence to lower loss values in terms of iterations, while requiring fewer FLOPs than baselines at the convergence. 
    }
\label{fig:val-loss-flops-slimp-unif-extra}
\end{figure*}

As shown in Figures~\ref{fig:val-loss-flops-slimp-doremi}-\ref{fig:val-loss-flops-slimp-unif-extra}, \textsc{Eslm} variants consistently accelerate validation loss convergence in the compute space, requiring fewer training FLOPs to achieve comparable or superior validation loss relative to baseline models. This efficiency gain holds across diverse pretraining corpora and model scales, highlighting the robustness of \textsc{Eslm} across settings.

\subsection{Downstream performance evaluation results}
\label{app:additional-downstream-performance}

Tables~\ref{tab:downstream-perf-124m-doremi}–\ref{tab:downstream-perf-774m-doremi} present the generalization performance of \textsc{Eslm} models ranging from 124M to 774M parameters, trained on different mixtures and evaluated against baseline models on downstream benchmarks.
All models are trained under a fixed compute budget measured in training FLOPs.
While the most consistent empirical gains of \textsc{Eslm} are observed in validation perplexity, convergence efficiency, and FLOPs-to-target, the downstream evaluations show that \textsc{Eslm} maintains competitive generalization performance while requiring fewer effective training FLOPs. Across the majority of benchmark configurations, \textsc{Eslm} matches or improves upon the CLM baseline in average downstream accuracy, although performance on individual tasks and model configurations naturally exhibits some variability. Together with the results across multiple model scales, two data mixtures (SlimPajama-Unif and SlimPajama-DoReMi), two selective learning variants (VaR-Entropy and CVaR-Loss), the adaptive variant (Ada-\textsc{Eslm}), and knowledge distillation (\textsc{Eslm}-KD), these findings demonstrate that the proposed selective learning framework generalizes across a broad range of pretraining settings without sacrificing downstream performance.

\begin{table}[t]
\caption{\textbf{Generalization performance of 124M models trained on SlimPajama-6B-DoReMi.} All models are pretrained under a $\sim$3E17 FLOPs budget. We report the best observed accuracy$_{( \text{standard error})}$ or exact match if provided, during training.
\myinlinecolorbox{tableyellow}{\textbf{Highlighted}} values indicate the best performance.
\looseness -1
}
\label{tab:downstream-perf-124m-doremi} 
\centering
    \resizebox{0.85\linewidth}{!}{%
        \begin{tabular}{l|c|cccccccc} \specialrule{1.5pt}{1pt}{1pt}
         \multicolumn{1}{c}{\makecell{\\  \\ \textbf{Benchmark}}} & \multicolumn{5}{c}{\textbf{Method (124M)}} \\ \cmidrule(lr){2-7}
        & \# Shots &  \multicolumn{1}{c}{\textsc{Eslm}-\text{\myinlinecolorbox{salmon!30}{$\operatorname{CVaR}$-loss}}} & \multicolumn{0}{c}{\textsc{Eslm}-\myinlinecolorbox{electric-blue!15}{$\operatorname{VaR}$-entropy}} & \multicolumn{0}{c}{CLM} & \multicolumn{0}{c}{Rho-1} & \multicolumn{0}{c}{GREATS} \\ 
        \specialrule{1.5pt}{1pt}{1pt}
        ARC-E  \citep{clark2018think} & 0-shot &  $0.3876_{(0.01)}$  & \tabye  $0.3926_{(0.01)}$  & $0.3838_{(0.01)}$ &   $0.3733_{(0.0099)}$  &   $0.3190_{(0.0096)}$ \\
        LAMBADA \citep{paperno-EtAl:2016:P16-1} & 5-shot &  \tabye $0.1738_{(0.0053)}$  &   $0.1727_{(0.0053)}$  & $0.1581_{(0.0051)}$ &   $0.1672_{(0.0052)}$  &  $0.0362_{(0.0026)}$ \\
         SciQ \citep{welbl2017crowdsourcing} & 5-shot 
 &   $0.731_{(0.014)}$  &   $0.718_{(0.0142)}$  & \tabye $0.735_{(0.014)}$ &   $0.723_{(0.0142)}$  &   $0.465_{(0.0158)}$ \\
        HellaSwag \citep{zellers2019hellaswag} &  5-shot &  $0.2936_{(0.0045)}$  &   $0.2924_{(0.0045)}$  &  \tabye $0.2945_{(0.0045)}$ &   $0.2905_{(0.0045)}$  &   $0.2639_{(0.0044)}$ \\
        TriviaQA \citep{2017arXivtriviaqa} & 1-shot &  \tabye $0.0184_{(0.001)}$  &   $0.0145_{(0.0009)}$  & $0.0100_{(0.0007)}$ &   $0.0112_{(0.0008)}$  &   $0.0007_{(0.0002)}$ \\
        COPA \citep{NEURIPS2019_4496bf24} & 5-shot &   $0.65_{(0.0479)}$  & \tabye  $0.66_{(0.0476)}$  & $0.66_{(0.0476)}$ &   $0.64_{(0.0482)}$  &   $0.65_{(0.0479)}$ \\
        MultiRC \citep{NEURIPS2019_4496bf24} & 5-shot &  $0.5455_{(0.0072)}$  &   $0.5367_{(0.0072)}$  & $0.5449_{(0.0072)}$ & \tabye  $0.5486_{(0.0071)}$  &   $0.5309_{(0.0072)}$ \\
        OpenBookQA \citep{mihaylov2018can} & 5-shot & \tabye  $0.176_{(0.017)}$  &   $0.164_{(0.0166)}$  & $0.174_{(0.017)}$ &   $0.164_{(0.0166)}$  &   $0.148_{(0.0159)}$ \\
        PiQA \citep{bisk2020piqa} & 5-shot & $0.6169_{(0.0113)}$  & \tabye  $0.6175_{(0.0113)}$  & $0.6017_{(0.0114)}$ &   $0.6033_{(0.0114)}$  &   $0.5489_{(0.0116)}$ \\
        \specialrule{1.5pt}{1pt}{1pt}
        \multicolumn{1}{c}{\textbf{Average} ($\uparrow$)} &    & \tabye \textbf{0.3992} & 	 \textbf{0.3964}	& 0.3957	& 0.3912  & 0.3291 \\
        \specialrule{1.5pt}{1pt}{1pt}
    \end{tabular}
    }
\end{table}

\begin{table}[t]
\caption{\textbf{Generalization performance of 350M models trained on SlimPajama-6B-Unif.} All models are pretrained under a $\sim$3.5E17 FLOPs budget. We report the best observed accuracy$_{( \text{standard error})}$ or exact match if provided, during training.
\myinlinecolorbox{tableyellow}{\textbf{Highlighted}} values indicate the best performance.
\looseness -1
}
\label{tab:downstream-perf-350m-unif} 
\centering
    \resizebox{0.85\linewidth}{!}{%
        \begin{tabular}{l|c|ccccc} \specialrule{1.5pt}{1pt}{1pt}
         \multicolumn{1}{c}{\makecell{\\  \\ \textbf{Benchmark}}} & \multicolumn{4}{c}{\textbf{Method (350M)}} \\ \cmidrule(lr){2-6}
        & \# Shots &  \multicolumn{1}{c}{\textsc{Eslm}-\text{\myinlinecolorbox{salmon!30}{$\operatorname{CVaR}$-loss}}} & \multicolumn{0}{c}{\textsc{Eslm}-\myinlinecolorbox{electric-blue!15}{$\operatorname{VaR}$-entropy}} & \multicolumn{0}{c}{CLM} & \multicolumn{0}{c}{Rho-1}  \\ 
        \specialrule{1.5pt}{1pt}{1pt}
        ARC-E  \citep{clark2018think} & 0-shot &  \tabye $0.4078_{(0.0101)}$  &   $0.3973_{(0.01)}$  & $0.4023_{(0.0101)}$ &   $0.4006_{(0.0101)}$  \\
        LAMBADA \citep{paperno-EtAl:2016:P16-1} & 5-shot & \tabye $0.2289_{(0.0059)}$  & $0.2070_{(0.0056)}$   &   $0.2095_{(0.0057)}$   &  $0.1993_{(0.0056)}$  \\
         SciQ \citep{welbl2017crowdsourcing} & 5-shot 
 & $0.749_{(0.0137)}$  & \tabye $0.768_{(0.0134)}$   &   $0.744_{(0.0138)}$   &  $0.760_{(0.0135)}$  \\
        HellaSwag \citep{zellers2019hellaswag} &  5-shot & \tabye $0.3301_{(0.0047)}$  & $0.3298_{(0.0047)}$   &   $0.3242_{(0.0047)}$   &  $0.3207_{(0.0047)}$  \\
        TriviaQA \citep{2017arXivtriviaqa} & 1-shot & \tabye $0.0338_{(0.0013)}$  & $0.0271_{(0.0012)}$   &   $0.0329_{(0.0013)}$   &  $0.0263_{(0.0012)}$  \\
        COPA \citep{NEURIPS2019_4496bf24} & 5-shot & \tabye $0.68_{(0.0469)}$  & $0.68_{(0.0469)}$   &   $0.68_{(0.0469)}$   &  $0.67_{(0.0473)}$  \\
        MultiRC \citep{NEURIPS2019_4496bf24} & 5-shot &  $0.5482_{(0.0071)}$  & $0.5556_{(0.0071)}$   &   $0.5492_{(0.0071)}$   & \tabye $0.5676_{(0.0071)}$  \\
        OpenBookQA \citep{mihaylov2018can} & 5-shot &  $0.276_{(0.02)}$  & \tabye $0.288_{(0.0203)}$   &   $0.280_{(0.0201)}$   &  $0.284_{(0.0202)}$  \\
        PiQA \citep{bisk2020piqa} & 5-shot & $0.6398_{(0.0112)}$  & \tabye $0.6420_{(0.0112)}$   &   $0.6349_{(0.0112)}$   &  $0.6322_{(0.0113)}$  \\
        \specialrule{1.5pt}{1pt}{1pt}
        \multicolumn{1}{c}{\textbf{Average} ($\uparrow$)} &    &  \textbf{0.4326} & \tabye \textbf{0.4327} & 0.4284 & 0.4289 \\
        \specialrule{1.5pt}{1pt}{1pt}
    \end{tabular}
    }
\end{table}

\begin{table}[t]
\caption{\textbf{Generalization performance of 350M models trained on SlimPajama-6B-DoReMi.} All models are pretrained under a $\sim$3.5E17 FLOPs budget. We report the best observed accuracy$_{( \text{standard error})}$ or exact match if provided, during training.
\myinlinecolorbox{tableyellow}{\textbf{Highlighted}} values indicate the best performance.
\looseness -1
}
\label{tab:downstream-perf-350m-doremi} 
\centering
    \resizebox{0.85\linewidth}{!}{%
        \begin{tabular}{l|c|ccccc} \specialrule{1.5pt}{1pt}{1pt}
         \multicolumn{1}{c}{\makecell{\\  \\ \textbf{Benchmark}}} & \multicolumn{4}{c}{\textbf{Method (350M)}} \\ \cmidrule(lr){2-6}
        & \# Shots &  \multicolumn{1}{c}{\textsc{Eslm}-\text{\myinlinecolorbox{salmon!30}{$\operatorname{CVaR}$-loss}}} & \multicolumn{0}{c}{\textsc{Eslm}-\myinlinecolorbox{electric-blue!15}{$\operatorname{VaR}$-entropy}} & \multicolumn{0}{c}{CLM} & \multicolumn{0}{c}{Rho-1}  \\ 
        \specialrule{1.5pt}{1pt}{1pt}
        ARC-E  \citep{clark2018think} & 0-shot &   $0.4170_{(0.0101)}$  &   $0.4048_{(0.0101)}$  & \tabye $0.4196_{(0.0101)}$ &   $0.4090_{(0.0101)}$  \\
        LAMBADA \citep{paperno-EtAl:2016:P16-1} & 5-shot & \tabye $0.2400_{(0.006)}$  & $0.2344_{(0.0059)}$   &   $0.2361_{(0.0059)}$   &  $0.2144_{(0.0057)}$  \\
         SciQ \citep{welbl2017crowdsourcing} & 5-shot 
 &  $0.775_{(0.0132)}$  & \tabye $0.794_{(0.0128)}$   &   $0.782_{(0.0131)}$   &  $0.773_{(0.0133)}$  \\
        HellaSwag \citep{zellers2019hellaswag} &  5-shot & $0.3292_{(0.0047)}$  & $0.3271_{(0.0047)}$   &    \tabye $0.3304_{(0.0047)}$   &  $0.3213_{(0.0047)}$  \\
        TriviaQA \citep{2017arXivtriviaqa} & 1-shot &  $0.0357_{(0.0014)}$  & \tabye $0.0431_{(0.0015)}$   &   $0.0414_{(0.0015)}$   &  $0.0395_{(0.0015)}$  \\
        COPA \citep{NEURIPS2019_4496bf24} & 5-shot &  $0.68_{(0.0469)}$  & \tabye $0.70_{(0.0461)}$   &   $0.69_{(0.0465)}$   &  $0.68_{(0.0469)}$  \\
        MultiRC \citep{NEURIPS2019_4496bf24} & 5-shot &  $0.5457_{(0.0072)}$  & \tabye $0.5645_{(0.0071)}$   &   $0.5548_{(0.0071)}$   &  $0.5558_{(0.0071)}$  \\
        OpenBookQA \citep{mihaylov2018can} & 5-shot & \tabye $0.286_{(0.0202)}$  & $0.284_{(0.0202)}$   &   $0.280_{(0.0201)}$   &  $0.286_{(0.0202)}$  \\
        PiQA \citep{bisk2020piqa} & 5-shot &  $0.6289_{(0.0113)}$  & \tabye $0.6414_{(0.0112)}$   &   $0.6354_{(0.0112)}$   &  $0.6354_{(0.0112)}$  \\
        \specialrule{1.5pt}{1pt}{1pt}
        \multicolumn{1}{c}{\textbf{Average} ($\uparrow$)} &    & 0.4375 & 	 \tabye  \textbf{0.4437}	& 0.4410	& 0.4349 \\
        \specialrule{1.5pt}{1pt}{1pt}
    \end{tabular}
    }
\end{table}

\begin{table}[ht]
\caption{\textbf{Generalization performance of 774M models trained on SlimPajama-6B-Unif.} All models are pretrained under a $\sim$1E18 FLOPs budget. We report the best observed accuracy$_{( \text{standard error})}$ or exact match if provided, during training.
\myinlinecolorbox{tableyellow}{\textbf{Highlighted}} values indicate the best performance.
\looseness -1
}
\label{tab:downstream-perf-774m-unif} 
\centering
    \resizebox{0.85\linewidth}{!}{%
        \begin{tabular}{l|c|ccccc} \specialrule{1.5pt}{1pt}{1pt}
         \multicolumn{1}{c}{\makecell{\\  \\ \textbf{Benchmark}}} & \multicolumn{4}{c}{\textbf{Method (774M)}} \\ \cmidrule(lr){2-6}
        & \# Shots &  \multicolumn{1}{c}{\textsc{Eslm}-\text{\myinlinecolorbox{salmon!30}{$\operatorname{CVaR}$-loss}}} & \multicolumn{0}{c}{\textsc{Eslm}-\myinlinecolorbox{electric-blue!15}{$\operatorname{VaR}$-entropy}} & \multicolumn{0}{c}{CLM} & \multicolumn{0}{c}{Rho-1}  \\ 
        \specialrule{1.5pt}{1pt}{1pt}
        ARC-E  \citep{clark2018think} & 0-shot &  $0.4040_{(0.0101)}$  & $0.4061_{(0.0101)}$   &  \tabye $0.4082_{(0.0101)}$   &  $0.3985_{(0.01)}$  \\
        LAMBADA \citep{paperno-EtAl:2016:P16-1} & 5-shot &  $0.2408_{(0.006)}$  & \tabye $0.2410_{(0.006)}$   &   $0.2336_{(0.0059)}$   &  $0.2404_{(0.006)}$  \\
         SciQ \citep{welbl2017crowdsourcing} & 5-shot 
 &  $0.763_{(0.0135)}$  & $0.760_{(0.0135)}$   &   $0.762_{(0.0135)}$   & \tabye $0.780_{(0.0131)}$  \\
        HellaSwag \citep{zellers2019hellaswag} &  5-shot &  $0.3348_{(0.0047)}$  & \tabye $0.3399_{(0.0047)}$   &   $0.3333_{(0.0047)}$   &  $0.3332_{(0.0047)}$  \\
        TriviaQA \citep{2017arXivtriviaqa} & 1-shot &  \tabye $0.0370_{(0.0014)}$  & $0.0315_{(0.0013)}$   &   $0.0349_{(0.0014)}$   &  $0.0298_{(0.0013)}$  \\
        COPA \citep{NEURIPS2019_4496bf24} & 5-shot &  $0.67_{(0.0473)}$  & $0.68_{(0.0469)}$   &  \tabye $0.69_{(0.0465)}$   &  $0.69_{(0.0465)}$  \\
        MultiRC \citep{NEURIPS2019_4496bf24} & 5-shot &  $0.5474_{(0.0071)}$  & $0.5622_{(0.0071)}$   &   $0.5591_{(0.0071)}$   & \tabye $0.5680_{(0.0071)}$  \\
        OpenBookQA \citep{mihaylov2018can} & 5-shot & \tabye $0.284_{(0.0202)}$  & $0.278_{(0.0201)}$  &   $0.278_{(0.0201)}$   &  $0.280_{(0.0201)}$  \\
        PiQA \citep{bisk2020piqa} & 5-shot &  $0.6458_{(0.0112)}$  & \tabye $0.6468_{(0.0112)}$   &   $0.6436_{(0.0112)}$   &  $0.6392_{(0.0112)}$  \\
        \specialrule{1.5pt}{1pt}{1pt}
\multicolumn{1}{c}{\textbf{Average} ($\uparrow$)} &    &  0.4363 &  \textbf{0.4383} & 0.4380 & \tabye \textbf{0.4399} \\
        \specialrule{1.5pt}{1pt}{1pt}
    \end{tabular}
    }
\end{table}

\begin{table}[ht]
\caption{\textbf{Generalization performance of 774M models trained on SlimPajama-6B-DoReMi.} All models are pretrained under a $\sim$1E18 FLOPs budget. We report the best observed accuracy$_{( \text{standard error})}$ or exact match if provided, during training.
\myinlinecolorbox{tableyellow}{\textbf{Highlighted}} values indicate the best performance.
\looseness -1
}
\label{tab:downstream-perf-774m-doremi} 
\centering
    \resizebox{0.85\linewidth}{!}{%
        \begin{tabular}{l|c|ccccc} \specialrule{1.5pt}{1pt}{1pt}
         \multicolumn{1}{c}{\makecell{\\  \\ \textbf{Benchmark}}} & \multicolumn{4}{c}{\textbf{Method (774M)}} \\ \cmidrule(lr){2-6}
        & \# Shots &  \multicolumn{1}{c}{\textsc{Eslm}-\text{\myinlinecolorbox{salmon!30}{$\operatorname{CVaR}$-loss}}} & \multicolumn{0}{c}{\textsc{Eslm}-\myinlinecolorbox{electric-blue!15}{$\operatorname{VaR}$-entropy}} & \multicolumn{0}{c}{CLM} & \multicolumn{0}{c}{Rho-1}  \\ 
        \specialrule{1.5pt}{1pt}{1pt}
        ARC-E  \citep{clark2018think} & 0-shot &  $0.4132_{(0.0101)}$  & \tabye $0.4158_{(0.0101)}$   &   $0.4128_{(0.0101)}$   &  $0.4141_{(0.0101)}$  \\
        LAMBADA \citep{paperno-EtAl:2016:P16-1} & 5-shot & \tabye $0.2437_{(0.006)}$  & $0.2400_{(0.006)}$   &   $0.2124_{(0.0057)}$   &  $0.2229_{(0.0058)}$  \\
         SciQ \citep{welbl2017crowdsourcing} & 5-shot 
 &  $0.799_{(0.0127)}$  & \tabye $0.801_{(0.0126)}$   &   $0.78_{(0.0131)}$   &  $0.8_{(0.0127)}$  \\
        HellaSwag \citep{zellers2019hellaswag} &  5-shot & \tabye $0.3417_{(0.0047)}$  & $0.3383_{(0.0047)}$   &   $0.3366_{(0.0047)}$   &  $0.3382_{(0.0047)}$  \\
        TriviaQA \citep{2017arXivtriviaqa} & 1-shot &  $0.0457_{(0.0016)}$  &\tabye $0.0470_{(0.0016)}$   &   $0.0412_{(0.0015)}$   &  $0.0388_{(0.0014)}$  \\
        COPA \citep{NEURIPS2019_4496bf24} & 5-shot & \tabye $0.71_{(0.0456)}$  & $0.69_{(0.0465)}$   &   $0.68_{(0.0469)}$   &  $0.67_{(0.0473)}$  \\
        MultiRC \citep{NEURIPS2019_4496bf24} & 5-shot &  $0.5435_{(0.0072)}$  & $0.5602_{(0.0071)}$   &  \tabye $0.5680_{(0.0071)}$   &  $0.5470_{(0.0071)}$  \\
        OpenBookQA \citep{mihaylov2018can} & 5-shot &  $0.288_{(0.0203)}$  & \tabye $0.294_{(0.0204)}$   &   $0.28_{(0.0201)}$   &  $0.284_{(0.0202)}$  \\
        PiQA \citep{bisk2020piqa} & 5-shot &  $0.6360_{(0.0112)}$  & $0.6338_{(0.0112)}$   &   \tabye $0.6430_{(0.0112)}$   &  $0.6289_{(0.0113)}$  \\
        \specialrule{1.5pt}{1pt}{1pt}
        \multicolumn{1}{c}{\textbf{Average} ($\uparrow$)} &    & \tabye \textbf{0.4467} &  \textbf{0.4466}	& 0.4393	& 0.4382 \\
        \specialrule{1.5pt}{1pt}{1pt}
    \end{tabular}
    }
\end{table}

\clearpage

\subsection{Ablation for Instance selection vs Token-level Loss Shaping}
\label{app:ablation-flops-savings}

\begin{figure}[ht]
    \centering  
    \begin{subfigure}[t]{0.4\linewidth}\centering{\includegraphics[width=1\linewidth,trim=0 0 0 0,clip]{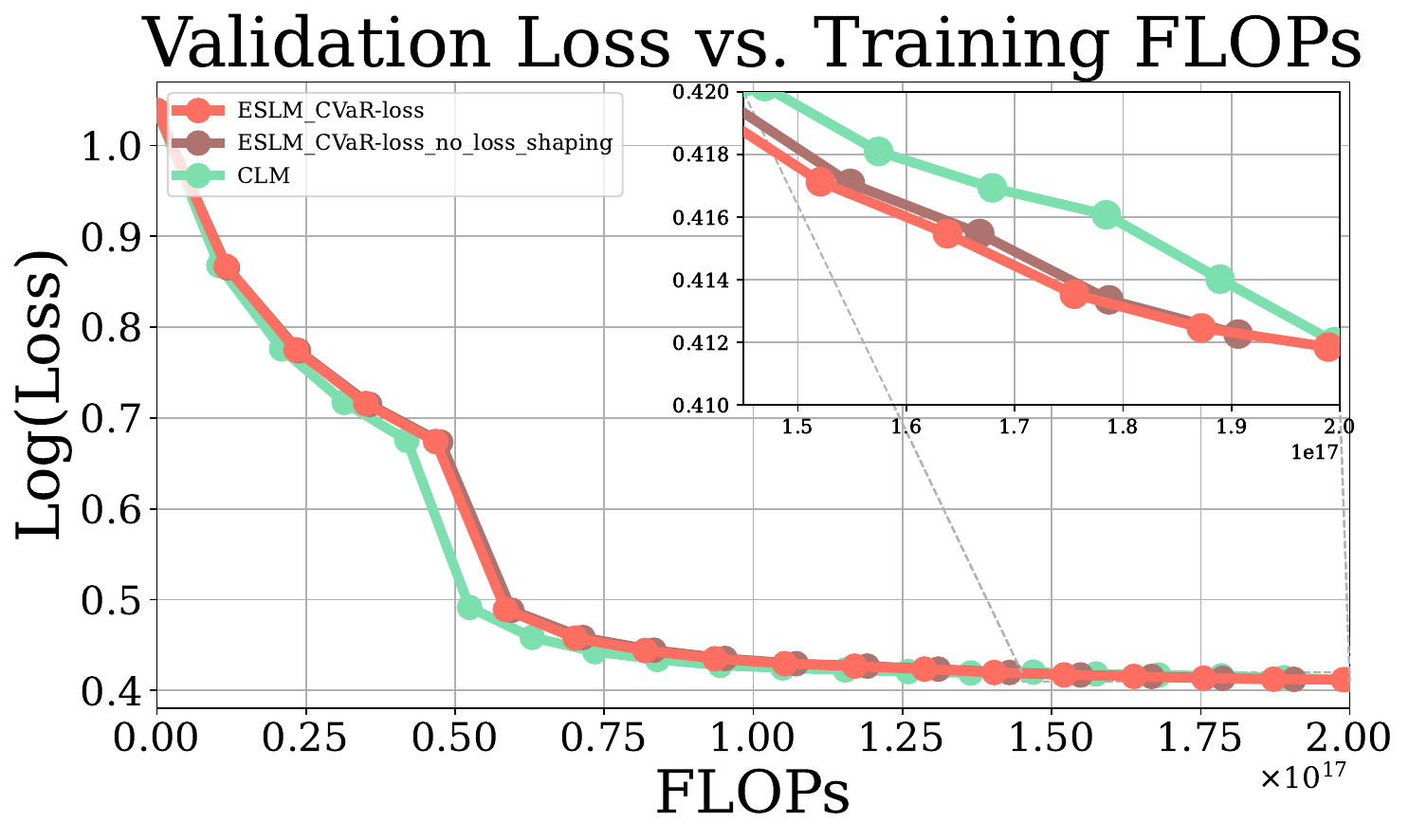}}
    \caption{Validation loss vs training FLOPs.}
    \label{fig:ablation-instance-vs-token-val-loss}
    \end{subfigure}%
    \begin{subfigure}[t]{0.4\linewidth}\centering{\includegraphics[width=1\linewidth,trim=0 0 0 0,clip]{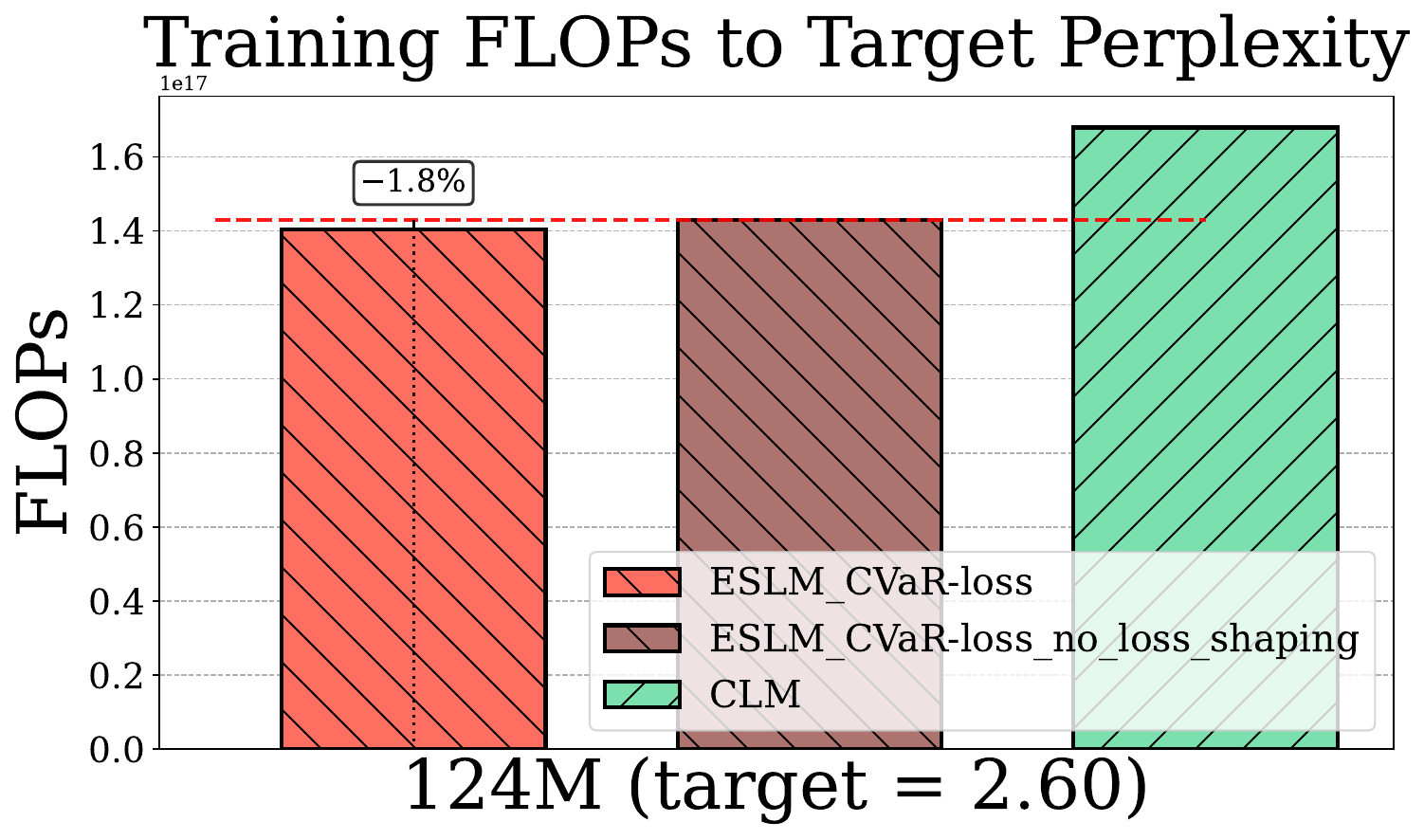}}
    \caption{Training FLOPs required ($\downarrow$).}
    \label{fig:ablation-instance-vs-token-flops}
    \end{subfigure}%
    \caption{\textbf{Ablation on instance selection vs token-level loss shaping.} Applying token-level loss shaping further helps in compute-efficient convergence to the target validation loss with fewer FLOPs compared to the instance-only selection version. 
    }
\label{fig:ablation-instance-vs-token-selection}
\end{figure}

In Figure~\ref{fig:ablation-instance-vs-token-selection}, we decompose the FLOPs savings of \textsc{Eslm} into contributions from the proxy instance-selection phase and the token-level loss-shaping phase on a 124M model trained on the SlimPajama-6B-Unif mixture. The instance selection phase provides the primary reduction by discarding low-value sequences early, while token-level shaping adds complementary savings by pruning redundant gradient updates within the retained sequences. Together, these two components improve pretraining efficiency over standard CLM training without degrading learning convergence.

\subsection{Robustness to Label Noise}
\label{app:robustness-label-noise}

\begin{figure}[ht]
    \centering  
    \begin{subfigure}[t]{0.4\linewidth}\centering{\includegraphics[width=1\linewidth,trim=0 0 0 0,clip]{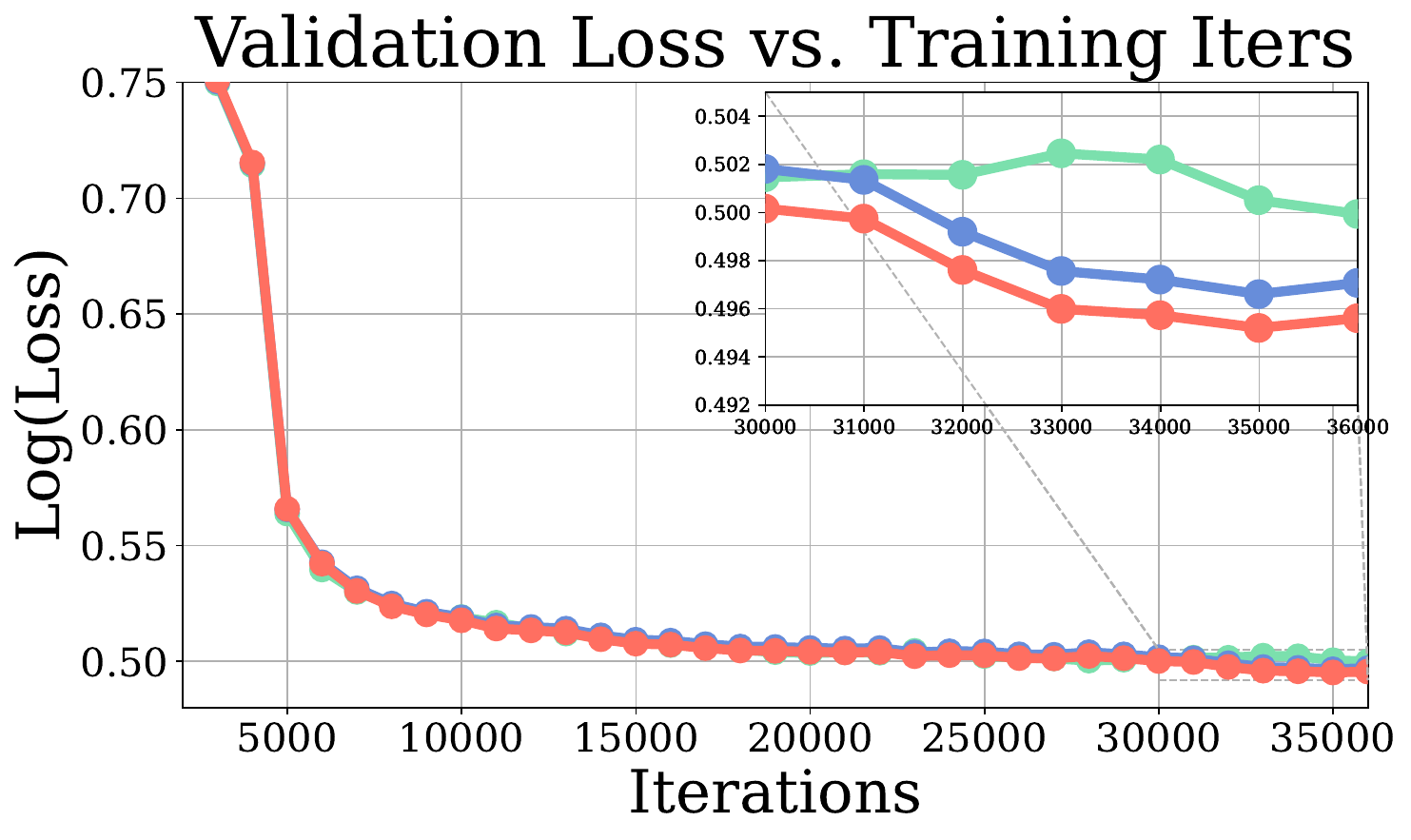}}
    \caption{ 124M (Iterations).}
    \label{fig:label-noise-results-iters-1}
    \end{subfigure}%
    \begin{subfigure}[t]{0.4\linewidth}\centering{\includegraphics[width=1\linewidth,trim=0 0 0 0,clip]{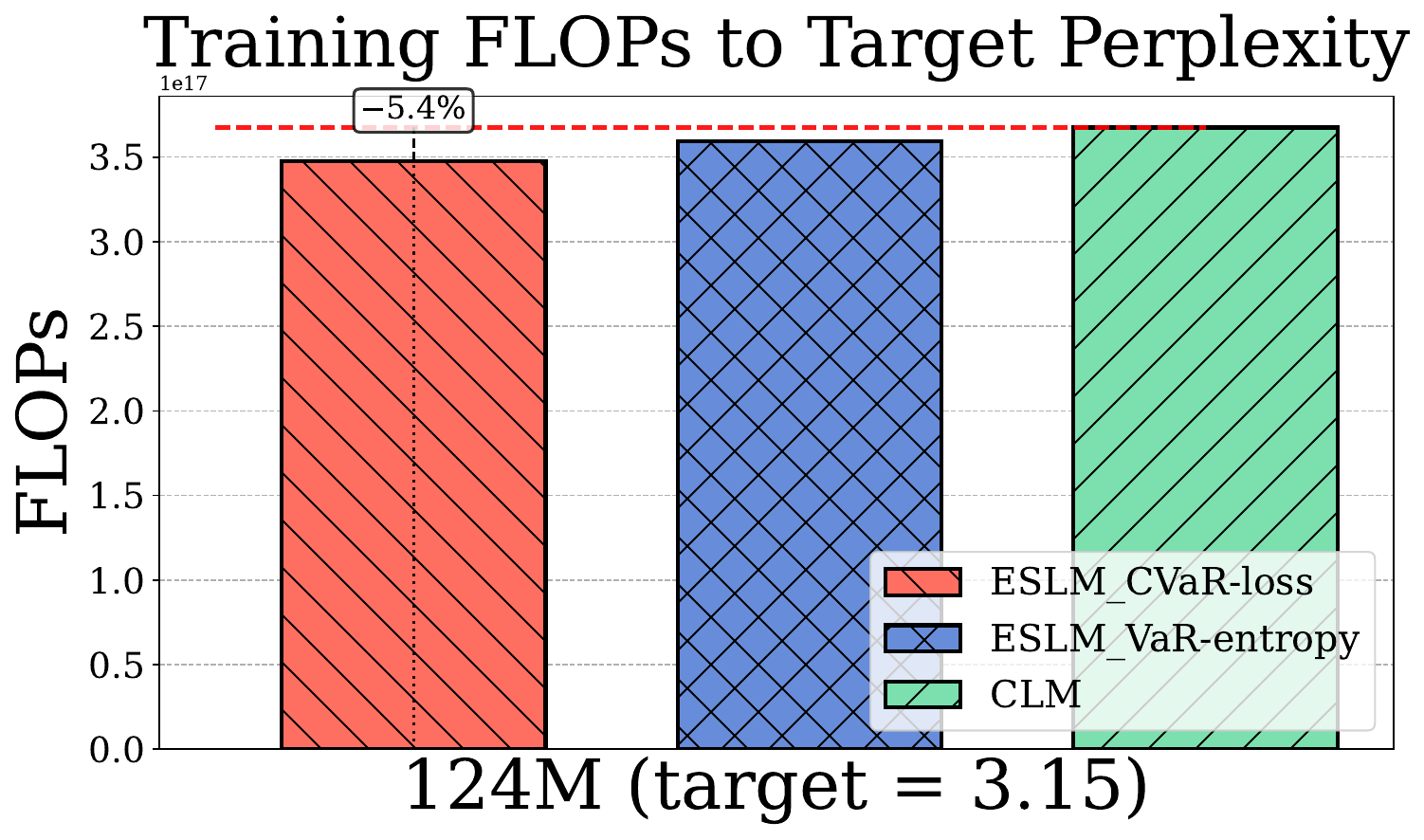}}
    \caption{ 124M (FLOPs).}
    \label{fig:label-noise-results-flops-1}
    \end{subfigure}%
    \setlength{\belowcaptionskip}{-7pt}
    \caption{ \textbf{Pretraining under 5\% label noise.} 
    \textbf{(a):} As shown with the learning trajectory using a running average with a window size of five evaluation points, \textsc{Eslm} variants converge faster to lower validation loss values under label noise compared to standard training.  
    \textbf{(b):} Even under noisy supervision, \textsc{Eslm} achieves the target validation loss with -5.4\% FLOPs savings compared to CLM.
    \looseness -1
    }
\label{fig:label-noise-results}
\end{figure}

To show the broader utility of selective pretraining with high-risk inputs, we further examine the robustness properties of ESLM grounded in the distributional robust optimization framework using label noise injection into the pretraining setup. In practice, real-world pretraining corpora inherently contain noisy tokens from data quality heterogeneity, a challenge in large-scale web data pretraining.
For this, we pretrain the models on SlimPajama-6B-Unif with label noise, randomly permuting a portion of the next-token targets.
We inject label noise by independently replacing each target label with probability $p$ with a uniformly sampled label from the vocabulary; labels not selected for corruption remain unchanged. This yields an expected corruption rate of approximately $p$.
The results in Figure~\ref{fig:label-noise-results} under 5\% label noise show that even under noisy supervision, \textsc{Eslm} variants achieve the target validation loss with fewer FLOPs compared to standard training, demonstrating that its risk-aware token selection mitigates the effect of corrupted gradients.
We further perform the same analysis under 10\% label noise. The results in Figure~\ref{fig:label-noise-results-10} show that \textsc{Eslm} provides a $5.8\%$ saving compared to the standard CLM training.
This experiment further supports the utility of \textsc{Eslm}'s hierarchical risk-aware filtering, not only in improving compute efficiency but also in enhancing robustness to label noise.

{\color{blue}

 \begin{figure}[ht]
    \centering  
    \begin{subfigure}[t]{0.4\linewidth}\centering{\includegraphics[width=1\linewidth,trim=0 0 0 0,clip]{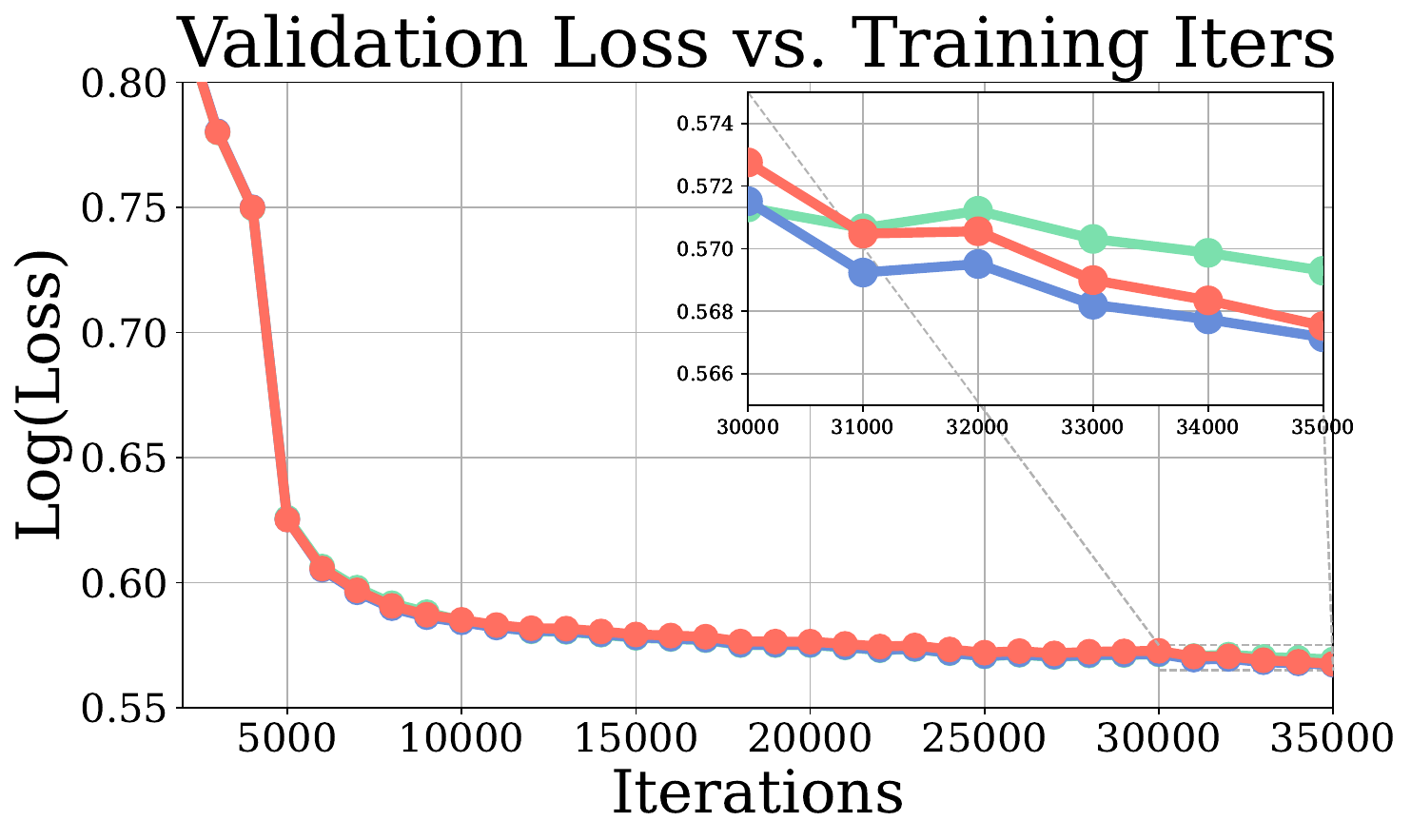}}
    \caption{ 124M (Iterations).}
    \label{fig:label-noise-results-iters-2}
    \end{subfigure}%
    \begin{subfigure}[t]{0.4\linewidth}\centering{\includegraphics[width=1\linewidth,trim=0 0 0 0,clip]{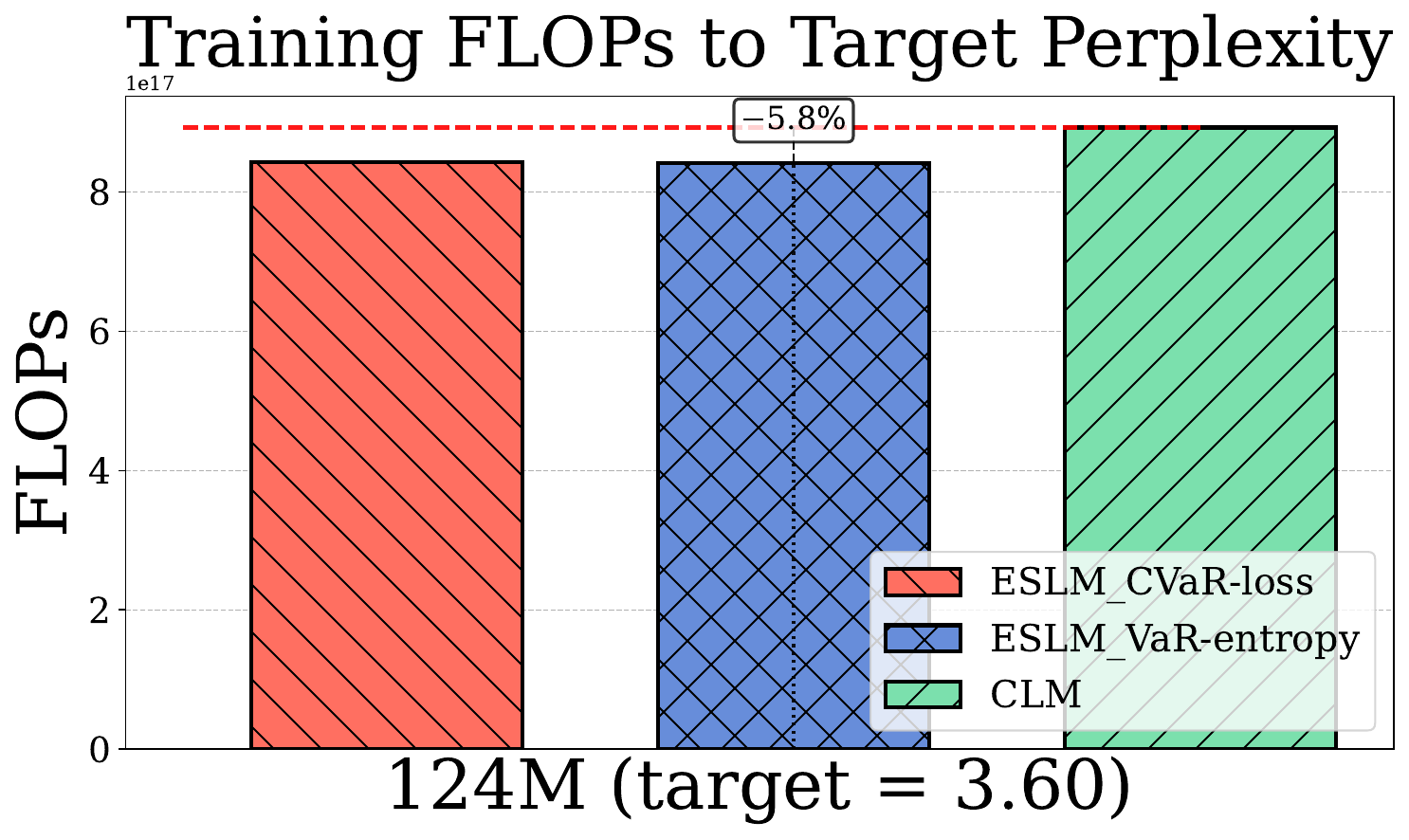}}
    \caption{ 124M (FLOPs).}
    \label{fig:label-noise-results-flops-2}
    \end{subfigure}%
    \setlength{\belowcaptionskip}{-7pt}
    \caption{ \textbf{Pretraining under 10\% label noise.} 
    \textbf{(a):} As shown with the learning trajectory using a running average with a window size of five evaluation points, \textsc{Eslm} variants converge faster to lower validation loss values under label noise compared to standard training.  
    \textbf{(b):} Even under noisy supervision, \textsc{Eslm} achieves the target validation loss with -5.8\% FLOPs savings compared to CLM.
    \looseness -1
    }
\label{fig:label-noise-results-10}
\end{figure}

}

\subsection{Comparison with Simpler Selection Methods}
\label{app:comparison-simpler-selection}

To isolate the source of ESLM's efficiency gains, we compare against a broader set of baselines beyond standard CLM and the closest prior selective-training methods, Rho-1 and GREATS. Specifically, we evaluate: \textit{(i)} random token selection, which discards tokens without using any model-dependent signal; \textit{(ii)} static pruning, which removes the bottom-10\% of tokens by a fixed score without a dynamic, distribution-adaptive threshold; and two ablated variants of ESLM that isolate individual components of its design — \textit{(iii)} CVaR-instance only, which applies risk-aware selection at the sequence level rather than the token level, and \textit{(iv)} CVaR-token loss shaping only, which reweights token losses by risk score without the hierarchical VaR-thresholded selection. Table~\ref{tab:additional-result-simpler-comp} reports the training compute (FLOPs) required for each method to reach a fixed target validation perplexity (log-ppl = 2.60) at the 124M-parameter scale.

\begin{table}[ht]
\centering
\caption{FLOPs required to reach target validation perplexity (log-perplexity~$=2.60$) at the 124M-parameter scale. Lower is better. The relative column shows change vs. the CLM baseline.}
\label{tab:additional-result-simpler-comp}
\begin{tabular}{lcc}
\toprule
\textbf{Method} & \textbf{FLOPs required} & \textbf{Relative to CLM} \\
\midrule
Random selection                          & $6.690 \times 10^{17}$ & $+298\%$ \\
GREATS                                    & $5.900 \times 10^{17}$ & $+251\%$ \\
Rho-1                                     & $1.887 \times 10^{17}$ & $+12\%$  \\
CLM (standard training)                            & $1.679 \times 10^{17}$ & --       \\
Static removing bottom-10\%               & $1.653 \times 10^{17}$ & $-2\%$   \\
\textsc{ESLM} -- token loss shaping only           & $1.647 \times 10^{17}$ & $-2\%$   \\
\textsc{ESLM} -- instance only                & $1.429 \times 10^{17}$ & $-15\%$  \\
\textbf{\textsc{ESLM}-CVaR}           & $\mathbf{1.404 \times 10^{17}}$ & $\mathbf{-16\%}$ \\
\bottomrule
\end{tabular}
\end{table}

Table~\ref{tab:additional-result-simpler-comp} shows that random selection performs worst by a wide margin (nearly 4× the FLOPs of CLM), confirming that ESLM's gains cannot be attributed to simply training on less data — the choice of which tokens to retain is what matters, not the reduction in token count itself. Second, neither static pruning nor loss-shaping alone meaningfully improves on CLM (both land within $~2\%$ of the CLM baseline), which shows that discarding or downweighting low-value tokens by a fixed, non-adaptive criterion is insufficient: the benefit of ESLM does not come from hard-example filtering in isolation, but from combining a risk-aware score with a distribution-adaptive VaR threshold that adjusts to the empirical score distribution during training. Third, instance-level risk-aware selection recovers most, but not all, of the full method's gains ($-15\%$ vs. $-16\%$ FLOPs), indicating that token-level granularity provides an additional, complementary benefit on top of the risk-aware selection mechanism itself, consistent with our motivation that instance-level filtering cannot preserve informative tokens embedded in otherwise low-value sequences.

Taken together, these results indicate that ESLM's efficiency gains stem specifically from the interaction between \textit{(a)} a distribution-aware thresholding mechanism, rather than a fixed cutoff, and \textit{(b)} token-level, rather than instance-level, granularity — and not merely from training on a smaller, hard-example-filtered subset of the corpus, as simpler baselines such as random selection, static pruning, or instance-only filtering fail to reproduce the full effect.

\section{ESLM token selection analysis}
\label{app:token-selection-analysis}
\begin{figure}[ht]
    \centering  
    \begin{subfigure}[t]{0.5\linewidth}\centering{\includegraphics[width=1\linewidth,trim=0 0 0 0,clip]{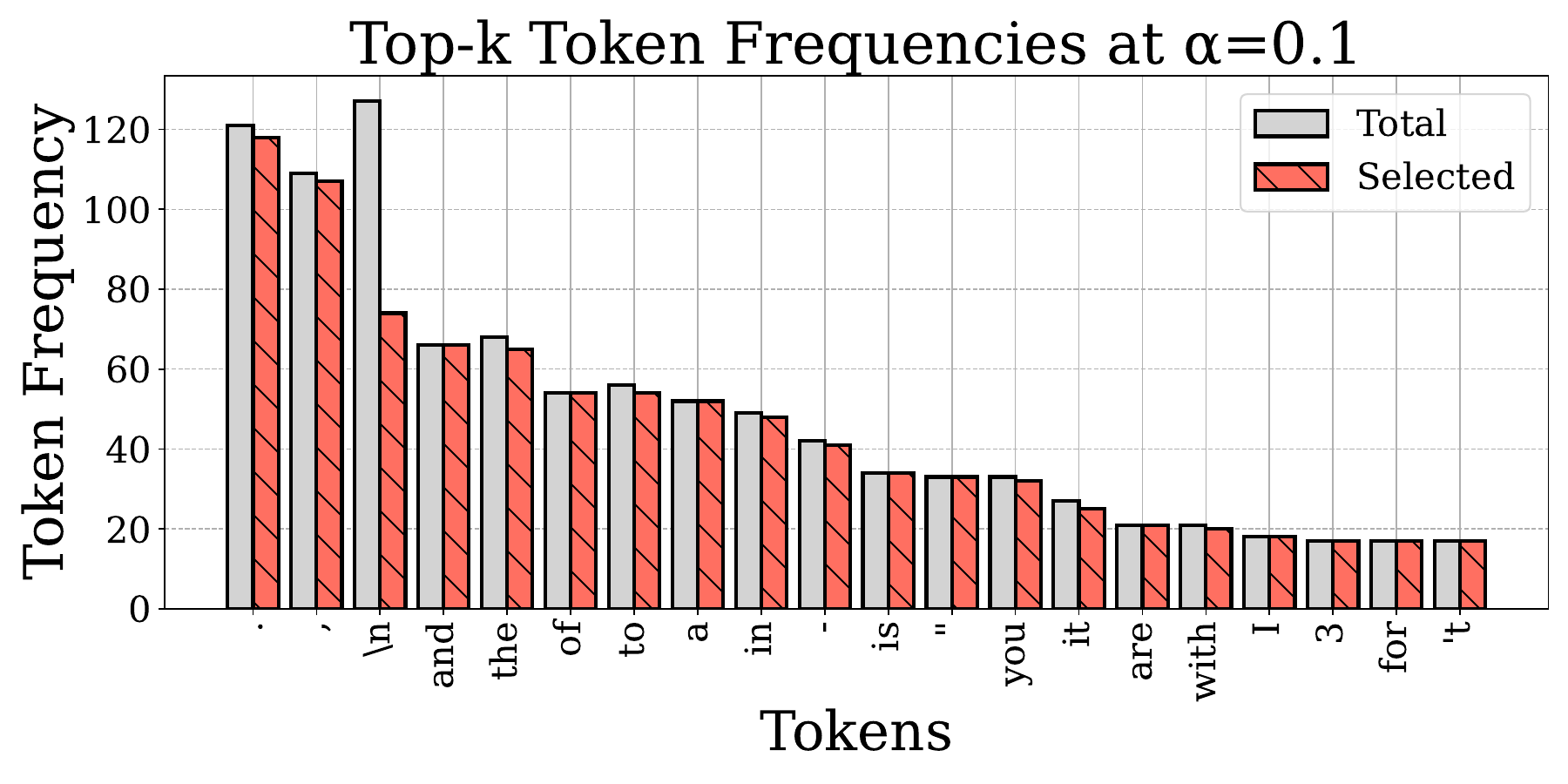}}
    \caption{\textsc{Eslm}-\myinlinecolorbox{salmon!30}{$\operatorname{CVaR}$-loss}, $\alpha=0.1$.}
    \label{fig:top-selected-tokens-eslm-cvar-loss-alpha0.1}
    \end{subfigure}%
    \begin{subfigure}[t]{0.5\linewidth}\centering{\includegraphics[width=1\linewidth,trim=0 0 0 0,clip]{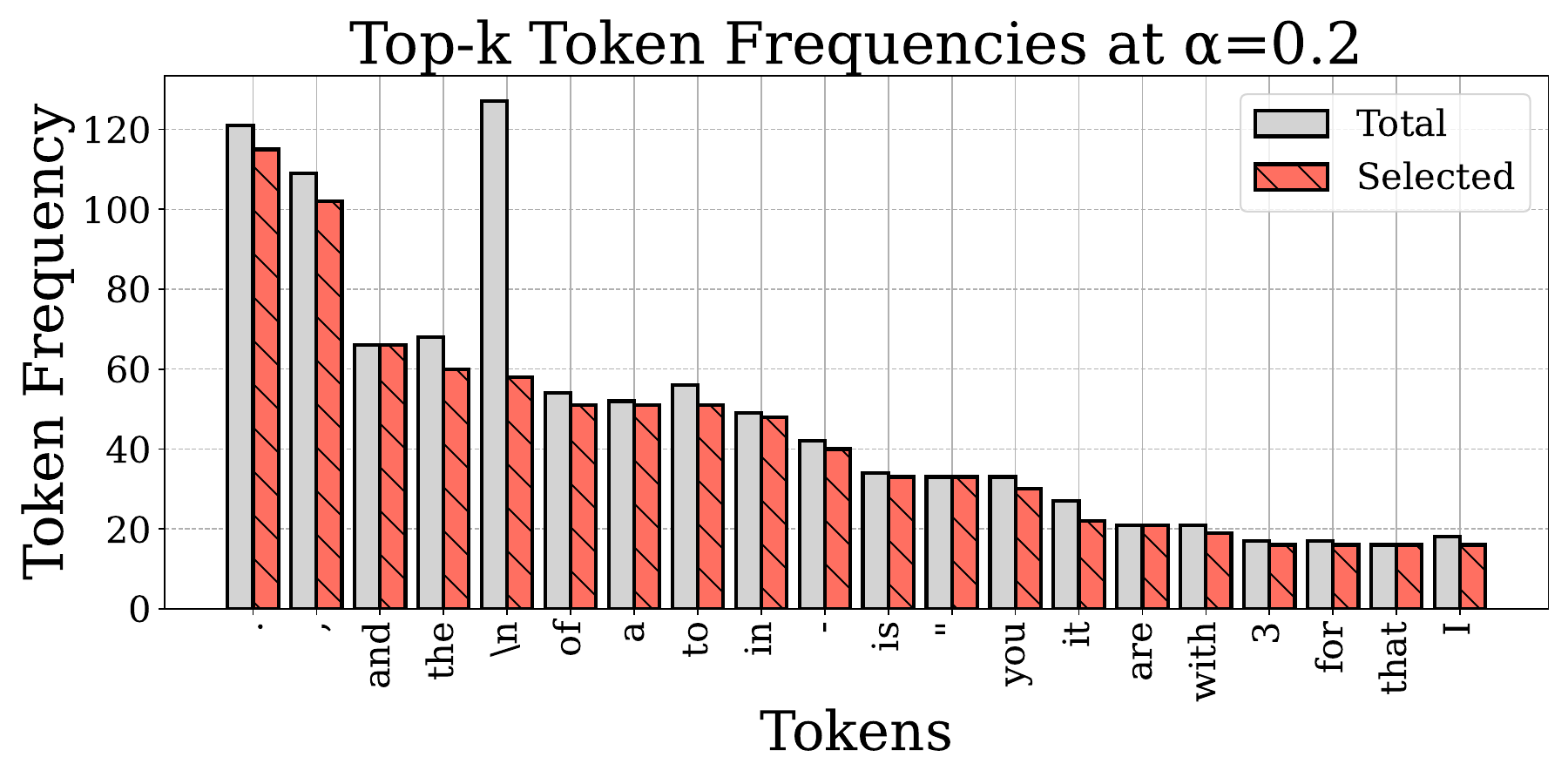}}
    \caption{\textsc{Eslm}-\myinlinecolorbox{salmon!30}{$\operatorname{CVaR}$-loss}, $\alpha=0.2$.}
    \label{fig:top-selected-tokens-eslm-cvar-loss-alpha0.2}
    \end{subfigure}%
    \\
    \begin{subfigure}[t]{0.5\linewidth}\centering{\includegraphics[width=1\linewidth,trim=0 0 0 0,clip]{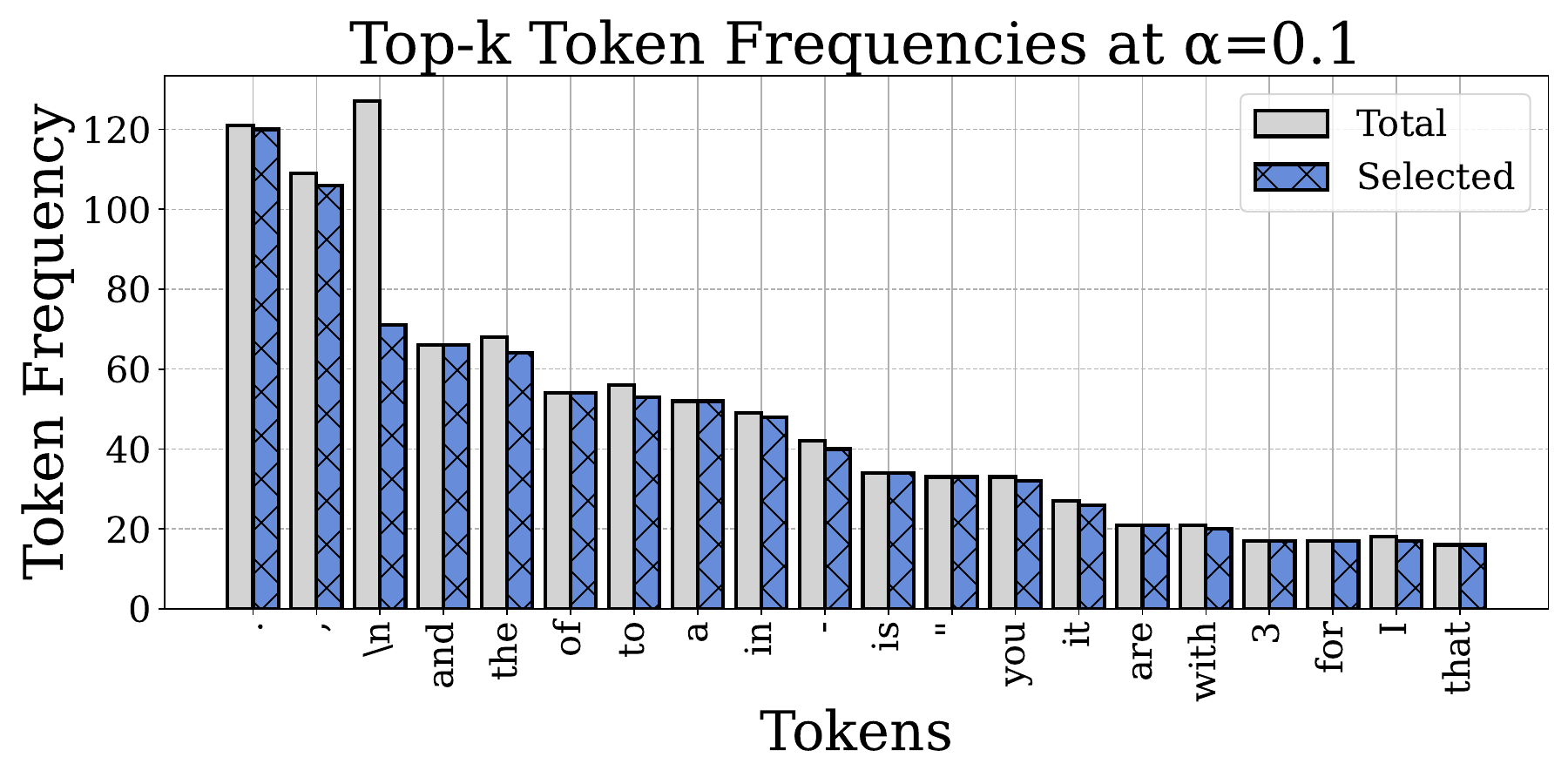}}
    \caption{\textsc{Eslm}-\myinlinecolorbox{electric-blue!15}{$\operatorname{VaR}$-entropy}, $\alpha=0.1$.}
    \label{fig:top-selected-tokens-eslm-var-entropy-alpha0.1}
    \end{subfigure}%
    \begin{subfigure}[t]{0.5\linewidth}\centering{\includegraphics[width=1\linewidth,trim=0 0 0 0,clip]{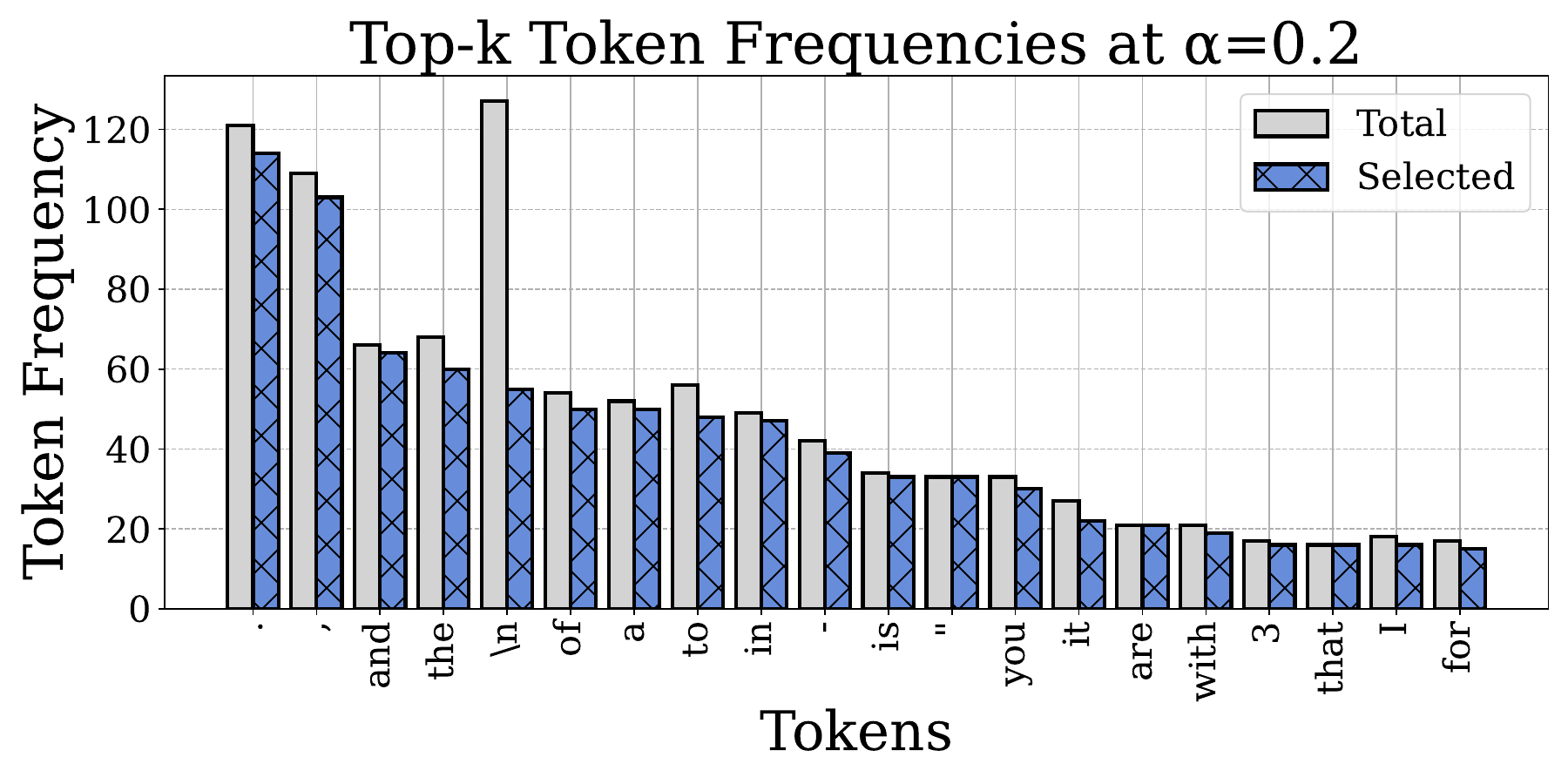}}
    \caption{\textsc{Eslm}-\myinlinecolorbox{electric-blue!15}{$\operatorname{VaR}$-entropy}, $\alpha=0.2$.}
    \label{fig:top-selected-tokens-eslm-var-entropy-alpha0.2}
    \end{subfigure}%
    \caption{\textbf{Frequency of top-20 tokens selected by \textsc{ESLM} (774M) variants from validation sequences shown in Figures~\ref{fig:eslm-example-selected-tokens-eslm-entropy-774M}-\ref{fig:eslm-example-selected-tokens-eslm-loss-774M}.} 
    As $\alpha$ decreases (more tokens are selected from the batch), the \textsc{Eslm}-\myinlinecolorbox{electric-blue!15}{$\operatorname{VaR}$-entropy} emphasizes punctuation tokens more than \textsc{Eslm}-\myinlinecolorbox{salmon!30}{$\operatorname{CVaR}$-loss}, reflecting its sensitivity to positional and contextual ambiguity, even in high-frequency tokens.}
\label{fig:top-selected-tokens}
\end{figure}

\begin{figure}[t]
    \centering
    \includegraphics[width=0.9\linewidth,trim=0 50 0 0,clip]{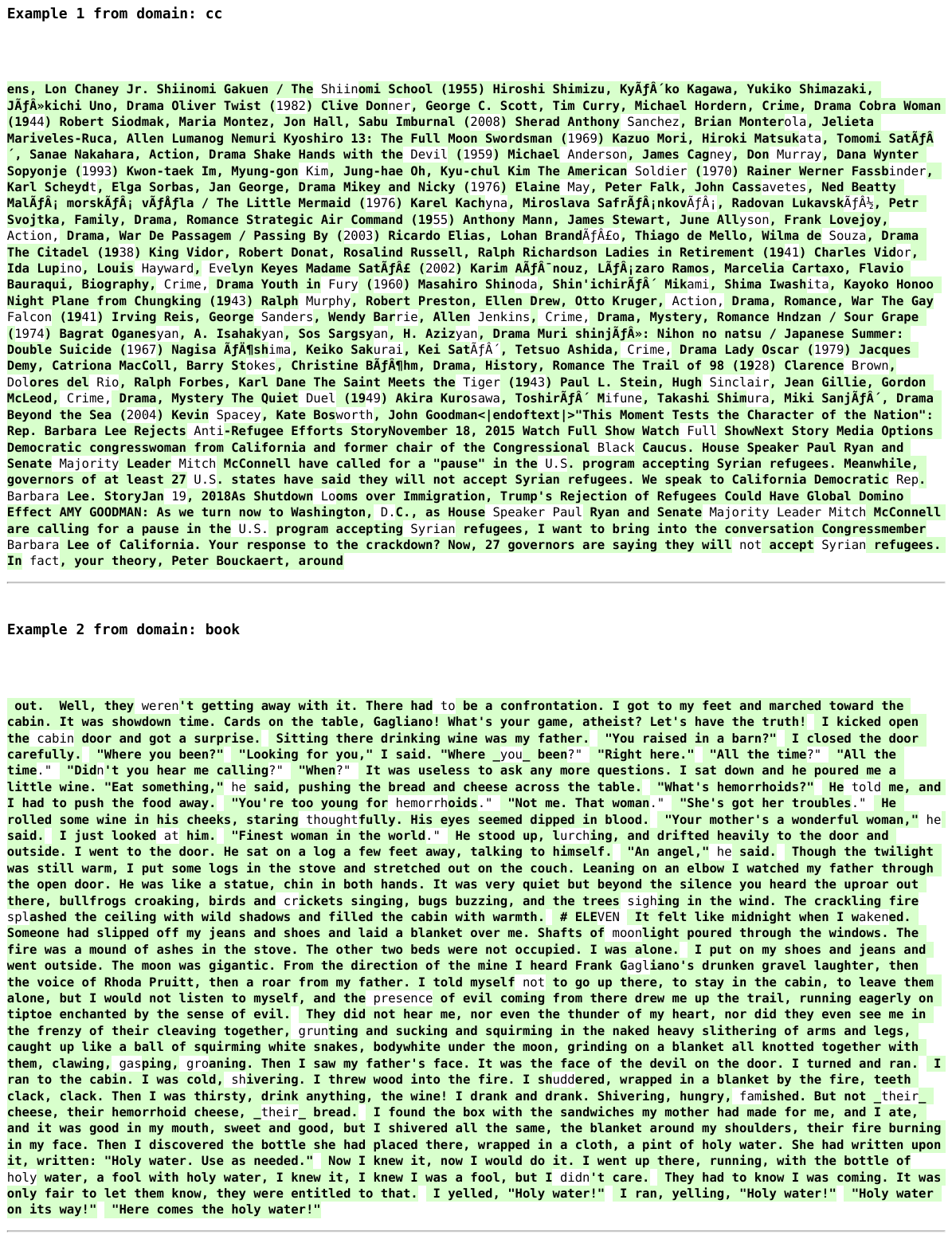}
    \caption{Example inputs from SlimPajama-6B-Unif mixture showing the \myinlinecolorbox{tokengreen}{\textbf{selected tokens}} by \textsc{Eslm}-\myinlinecolorbox{electric-blue!15}{$\operatorname{VaR}$-entropy} (\textbf{124M}) with $\alpha=0.1$. [\textit{Note: These examples are drawn from public datasets \citep{cerebras2023slimpajama} and may contain intense language, political references, or mature content. These excerpts are included solely for the purpose of analyzing model behavior.}] }  
    \label{fig:eslm-example-selected-tokens-eslm-entropy}
\end{figure}

\begin{figure}[t]
    \centering
    \includegraphics[width=0.9\linewidth,trim=0 50 0 0,clip]{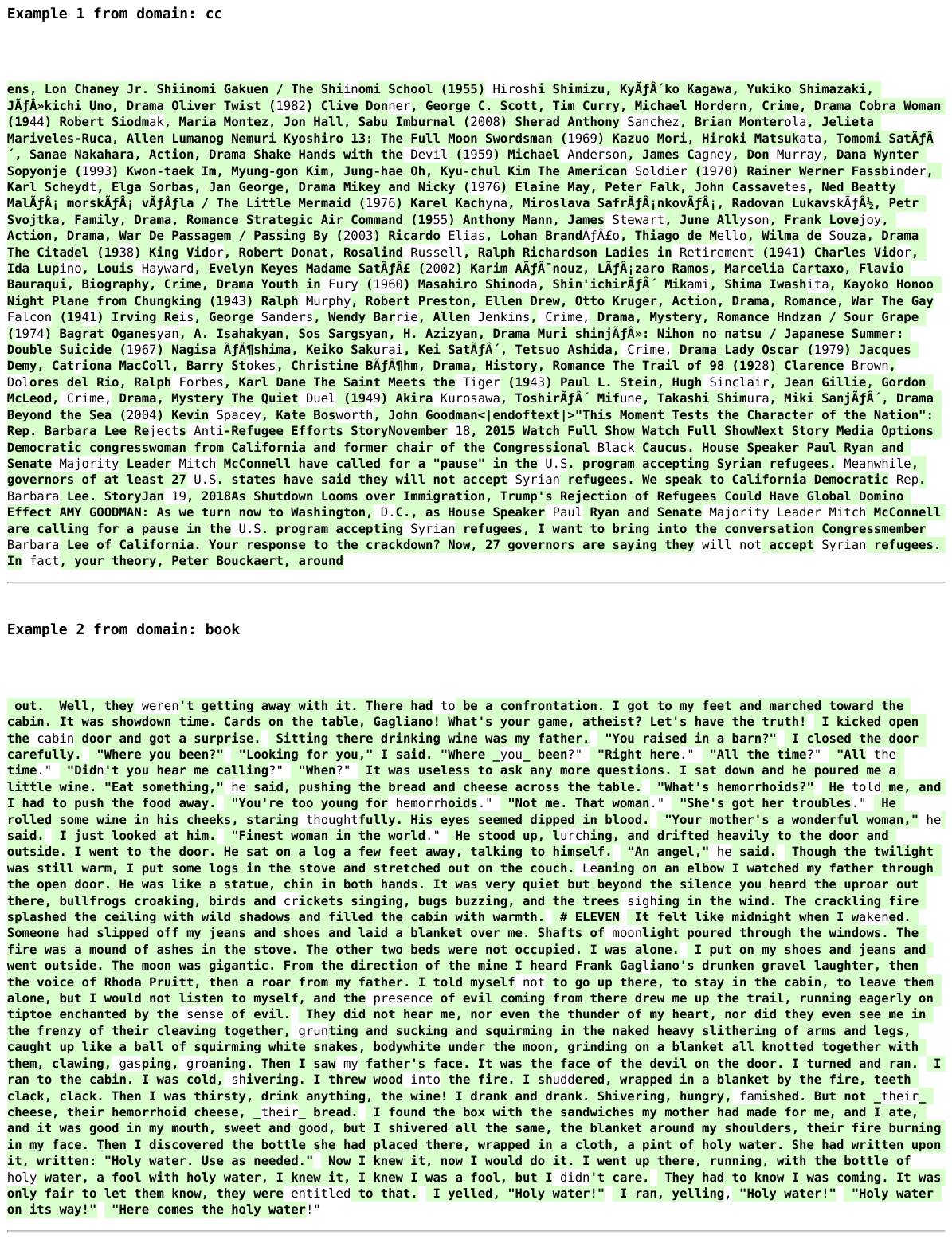}
    \caption{Example inputs from SlimPajama-6B-Unif mixture showing the \myinlinecolorbox{tokengreen}{\textbf{selected tokens}} by \textsc{Eslm}-\myinlinecolorbox{salmon!30}{$\operatorname{CVaR}$-loss} (\textbf{124M}) with $\alpha=0.1$. [\textit{Note: These examples are drawn from public datasets \citep{cerebras2023slimpajama} and may contain intense language, political references, or mature content. These excerpts are included solely for the purpose of analyzing model behavior.}]}  
    \label{fig:eslm-example-selected-tokens-eslm-loss}
\end{figure}

\begin{figure}[t]
    \centering
    \begin{subfigure}[t]{0.6\linewidth}\centering{\includegraphics[width=1\linewidth,trim=0 460 0 25,clip]{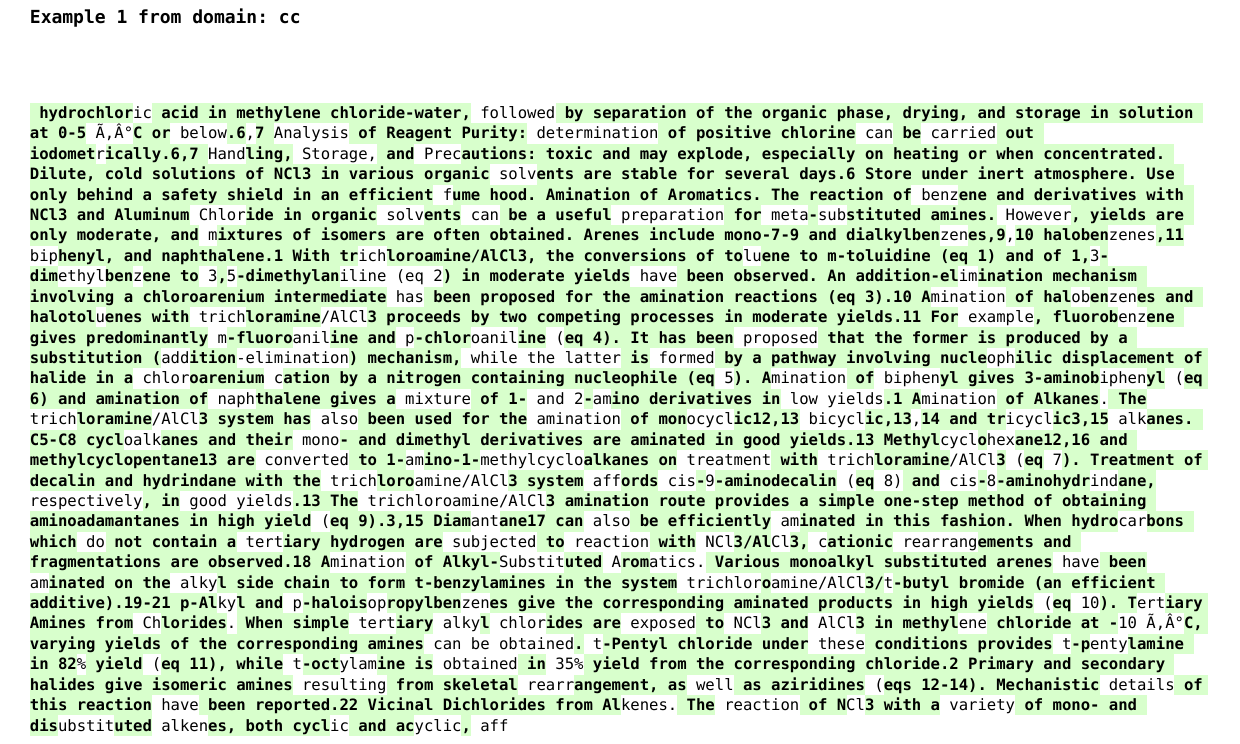}}
    \label{fig:entropy-ex-1}
    \end{subfigure}%
    \\
    \begin{subfigure}[t]{0.6\linewidth}\centering{\includegraphics[width=1\linewidth,trim=0 370 0 25,clip]{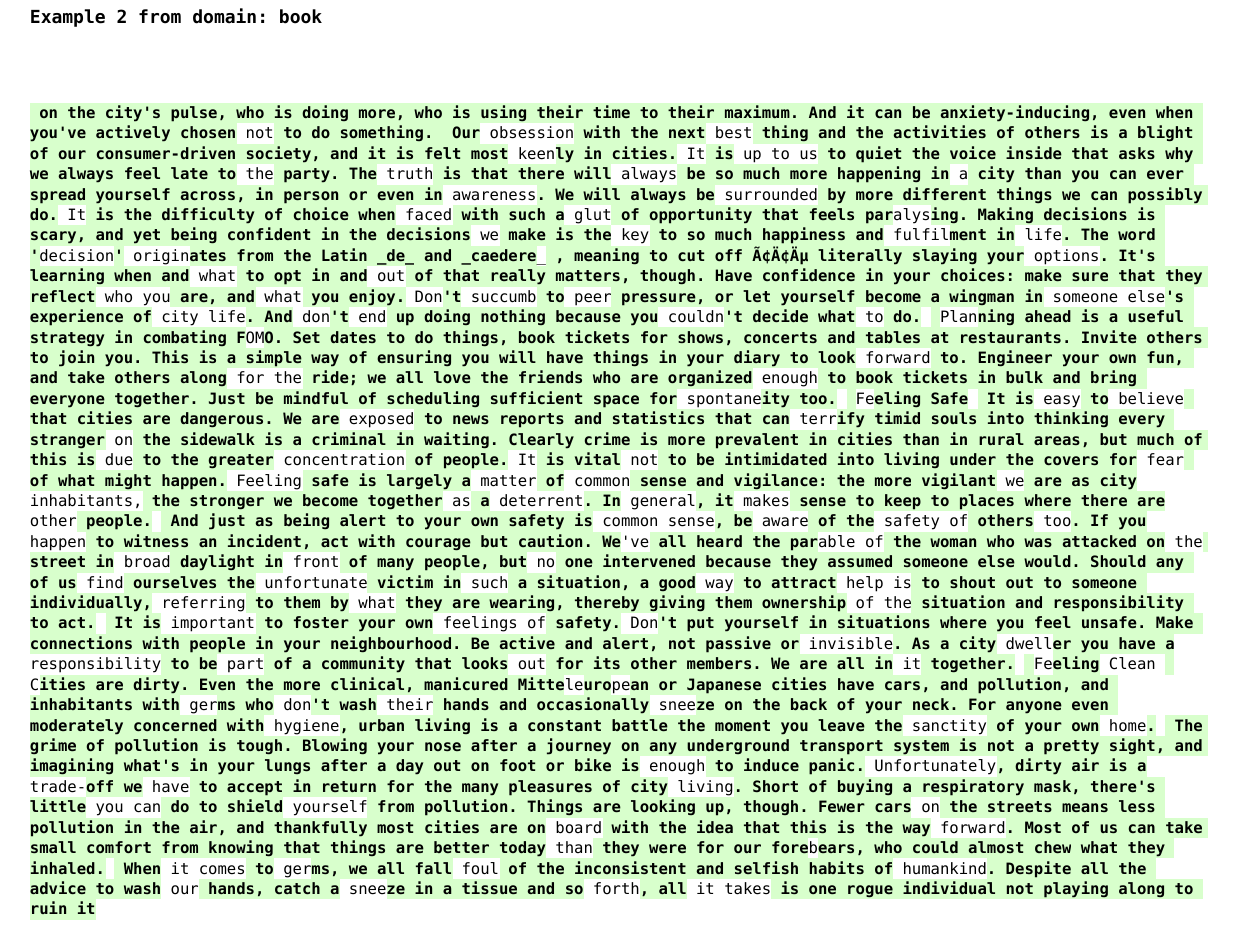}}
    \label{fig:entropy-ex-2}
    \end{subfigure}%
    \\
    \begin{subfigure}[t]{0.6\linewidth}\centering{\includegraphics[width=1\linewidth,trim=0 460 0 25,clip]{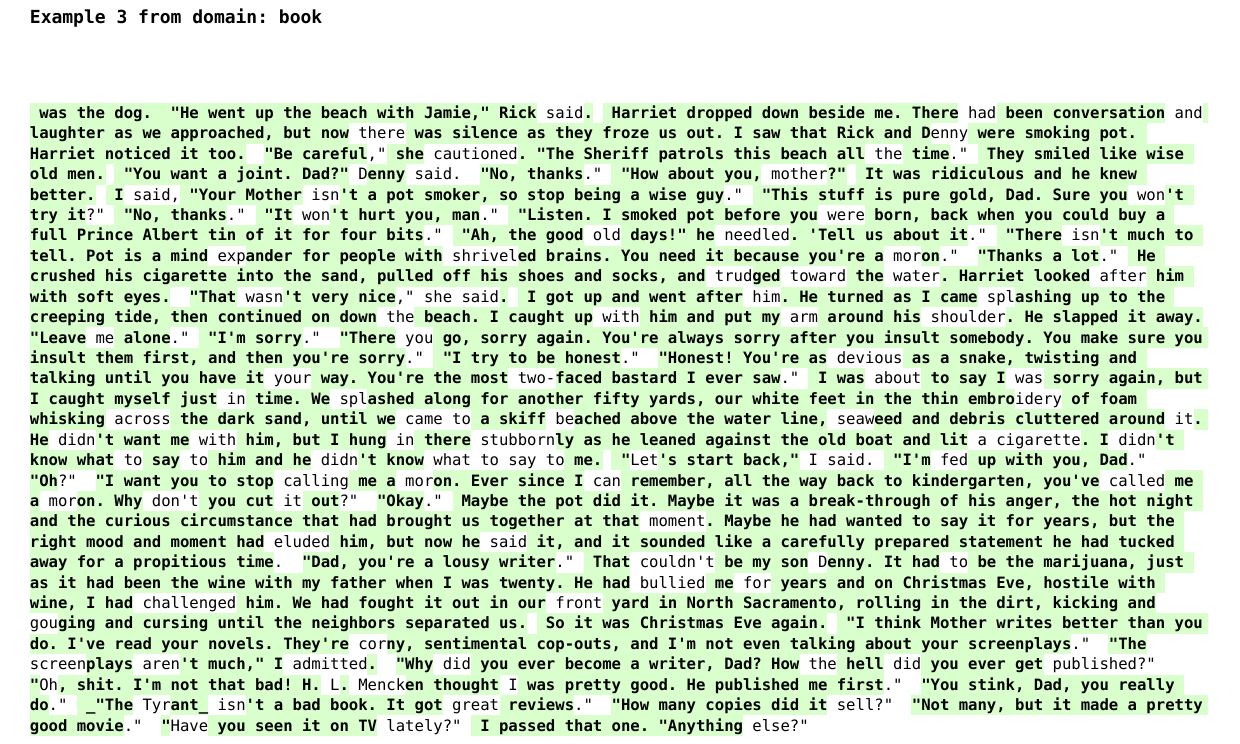}}
    \label{fig:entropy-ex-3}
    \end{subfigure}%
    \caption{Example inputs from SlimPajama-6B-Unif mixture showing the \myinlinecolorbox{tokengreen}{\textbf{selected tokens}} by \textsc{Eslm}-\myinlinecolorbox{electric-blue!15}{$\operatorname{VaR}$-entropy} (\textbf{774M}) with $\alpha=0.1$. [\textit{Note: These examples are drawn from public datasets \citep{cerebras2023slimpajama} and may contain intense language, political references, or mature content. These excerpts are included solely for the purpose of analyzing model behavior.}] }  
    \label{fig:eslm-example-selected-tokens-eslm-entropy-774M}
\end{figure}

\begin{figure}[t]
    \centering
    \begin{subfigure}[t]{0.6\linewidth}\centering{\includegraphics[width=1\linewidth,trim=0 460 0 25,clip]{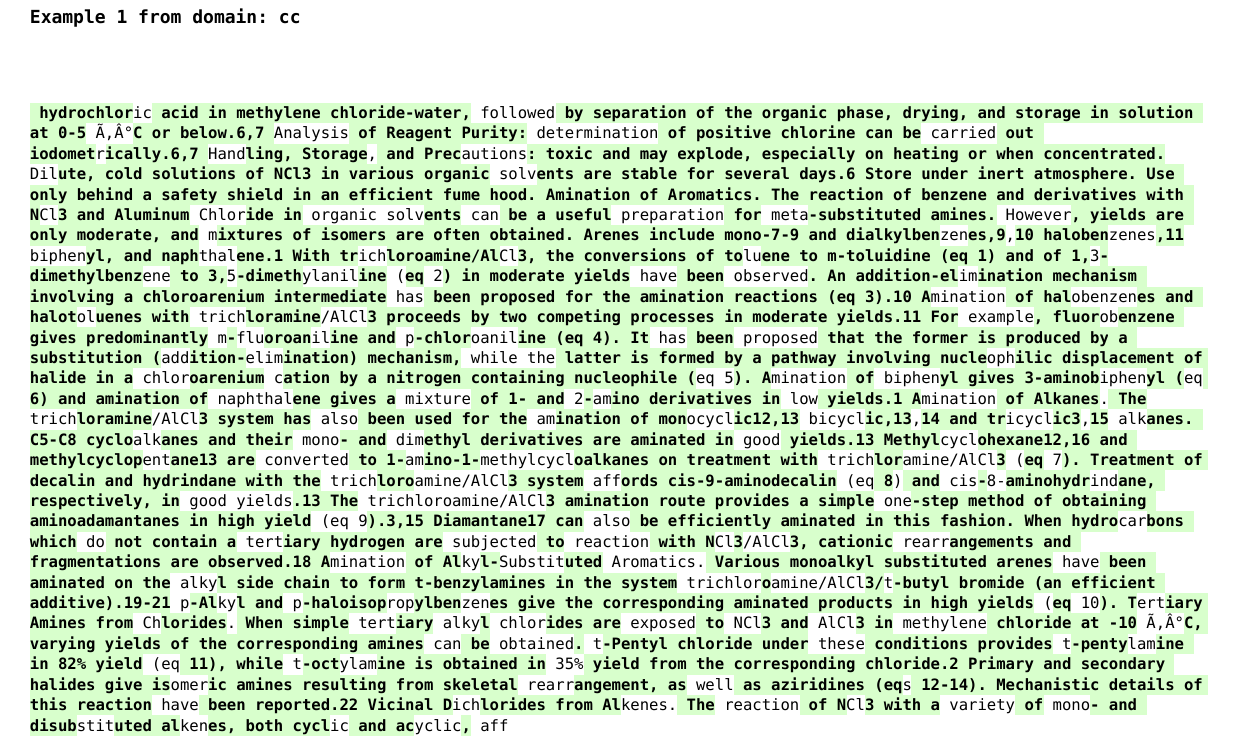}}
    \label{fig:loss-ex-1}
    \end{subfigure}%
    \\
    \begin{subfigure}[t]{0.6\linewidth}\centering{\includegraphics[width=1\linewidth,trim=0 370 0 25,clip]{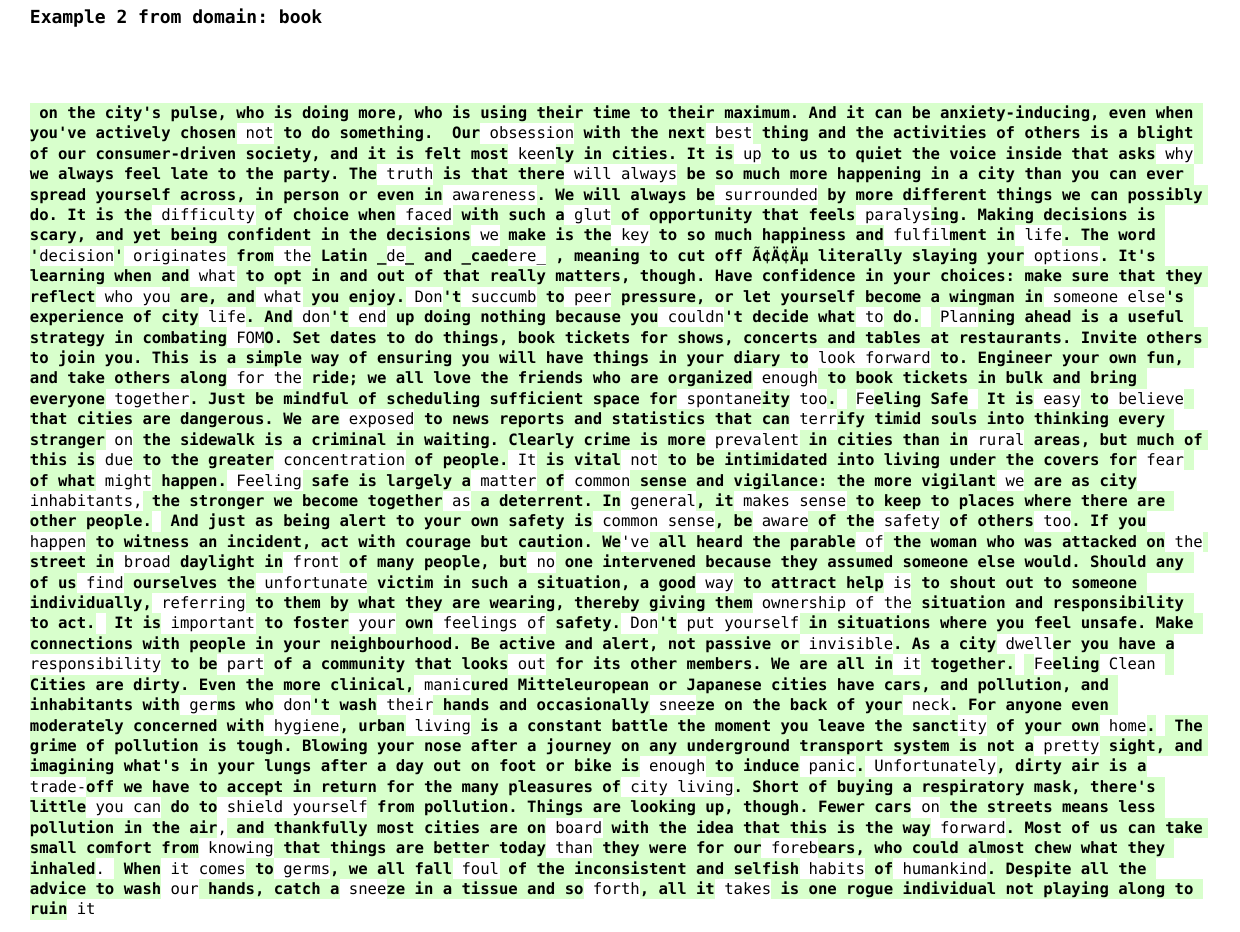}}
    \label{fig:loss-ex-2}
    \end{subfigure}%
    \\
    \begin{subfigure}[t]{0.6\linewidth}\centering{\includegraphics[width=1\linewidth,trim=0 460 0 25,clip]{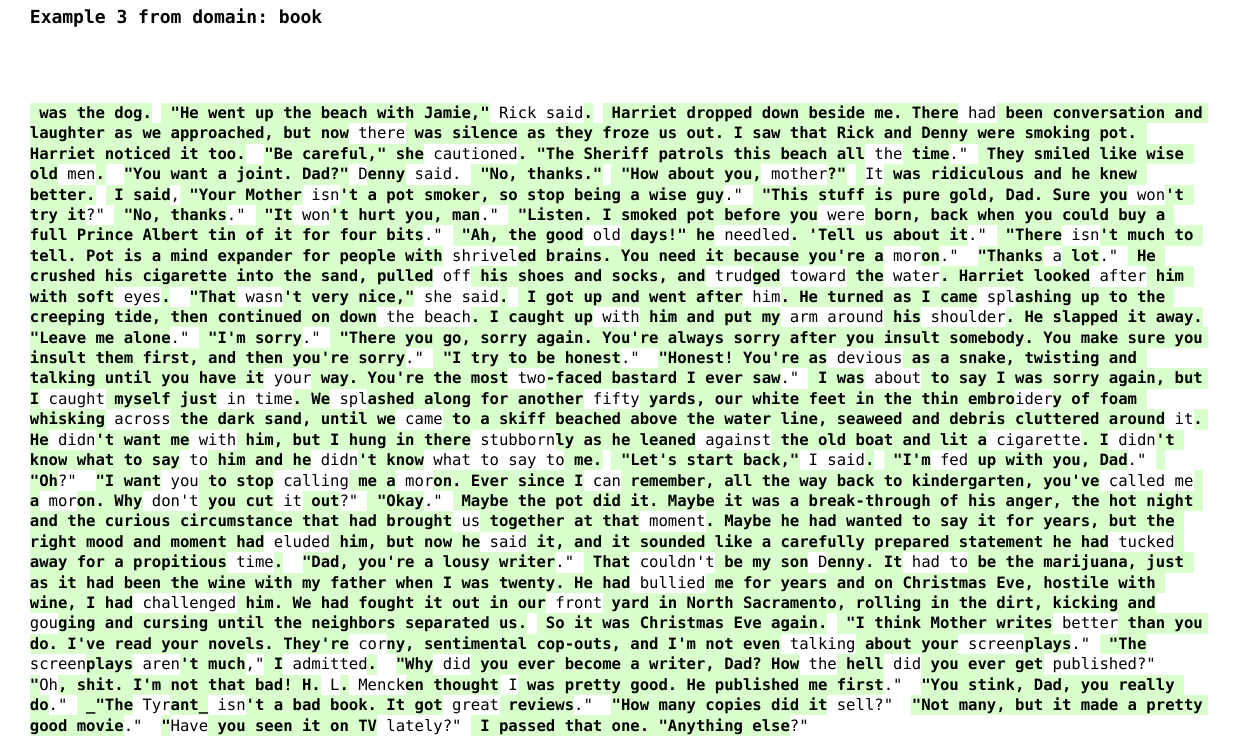}}
    \label{fig:loss-ex-3}
    \end{subfigure}%
    \caption{Example inputs from SlimPajama-6B-Unif mixture showing the \myinlinecolorbox{tokengreen}{\textbf{selected tokens}} by \textsc{Eslm}-\myinlinecolorbox{salmon!30}{$\operatorname{CVaR}$-loss} (\textbf{774M}) with $\alpha=0.1$. [\textit{Note: These examples are drawn from public datasets \citep{cerebras2023slimpajama} and may contain intense language, political references, or mature content. These excerpts are included solely for the purpose of analyzing model behavior.}] }  
    \label{fig:eslm-example-selected-tokens-eslm-loss-774M}
\end{figure}

To better understand the behavior of \textsc{Eslm}, we conduct a qualitative analysis of token selection during pretraining. Specifically, we compare the selection patterns of two \textsc{Eslm} variants on the same SlimPajama-6B validation sequences.
Figures~\ref{fig:eslm-example-selected-tokens-eslm-entropy}-\ref{fig:eslm-example-selected-tokens-eslm-loss} present examples where \myinlinecolorbox{tokengreen}{\textbf{highlighted tokens}} represent those selected for backpropagation
by 124M models using a fixed confidence level $\alpha = 0.1$ (i.e., top 90\% high-risk tokens retained in the objective).
In Figures~\ref{fig:eslm-example-selected-tokens-eslm-entropy-774M}-\ref{fig:eslm-example-selected-tokens-eslm-loss-774M}, we further show selected tokens by 774M models under $\alpha = 0.2$ (i.e., top 80\% high-risk tokens retained).
We observe that both variants prioritize rare or informative tokens—such as named entities, foreign words, and domain-specific phrases—but differ in the nature of the signals they capture:
\vspace{-0.1cm}
\begin{itemize}[left=0cm]
    \item \textsc{Eslm}-\myinlinecolorbox{electric-blue!15}{$\operatorname{VaR}$-entropy} emphasizes tokens associated with high predictive uncertainty, often selecting structurally or semantically transitional words, including common function words (e.g., “the”, “and”, “of”), punctuation, and formatting artifacts when they appear in unpredictable or shifting contexts. For instance, in the second passage (Figure~\ref{fig:eslm-example-selected-tokens-eslm-entropy-774M}), the model selects not only semantically meaningful words such as “anxiety-inducing”, “consumer-driven”, but also emphasizes “the”, “.”, and “with” in contexts where uncertainty over their grammatical role or continuation is high.
    
    \item \textsc{Eslm}-\myinlinecolorbox{salmon!30}{$\operatorname{CVaR}$-loss} instead selects tokens that incur high training loss—typically semantically complex or underfit tokens. 
    In the first passage (Figure~\ref{fig:eslm-example-selected-tokens-eslm-loss-774M}), we observe selection of technical terms such as “alkanes”, “aminated”; but tend to avoid repeating tokens such as "eq". This variant tends to avoid punctuation and common syntactic tokens unless they contribute directly to high loss.
\vspace{-0.1cm}
\end{itemize}
Figure~\ref{fig:top-selected-tokens} further illustrates the frequency of top-20 selected tokens by 774M \textsc{Eslm} models, from the validation examples given in Figures~\ref{fig:eslm-example-selected-tokens-eslm-entropy-774M}-\ref{fig:eslm-example-selected-tokens-eslm-loss-774M}.
The results reveal that as we allow for more tokens to be selected ($\alpha$ decreases), \textsc{Eslm}-\myinlinecolorbox{electric-blue!15}{$\operatorname{VaR}$-entropy} selects syntactically ambiguous tokens such as punctuations more than \textsc{Eslm}-\myinlinecolorbox{salmon!30}{$\operatorname{CVaR}$-loss}, reflecting its sensitivity to positional and contextual ambiguity, even in high-frequency tokens.

Crucially, this overall analysis also highlights the strength of token-level selection: \textsc{Eslm} captures the informativeness within sequences, in contrast to instance-level methods such as GREATS that filter entire examples. As a result, \textsc{Eslm} preserves valuable learning signals that would otherwise be discarded, offering a more fine-grained and efficient form of selective pretraining.

\end{document}

%% file: math_commands.tex
\usepackage{amsmath,amsfonts,bm}

\def\eqref#1{equation~\ref{#1}}

\def\1{\bm{1}}

\DeclareMathAlphabet{\mathsfit}{\encodingdefault}{\sfdefault}{m}{sl}
\SetMathAlphabet{\mathsfit}{bold}{\encodingdefault}{\sfdefault}{bx}{n}

%% file: config/definitions.tex
\newtheorem*{theorem*}{Theorem}
\newtheorem*{proposition*}{Proposition}

\definecolor{darkgreen}{RGB}{104, 196, 84}